\documentclass[11pt,a4paper]{article}

\usepackage[a4paper,left=2.2cm,right=2.2cm, top=2.2cm, bottom=2.7cm]{geometry}

\usepackage[pdftex]{graphicx}
\usepackage{placeins}  
\usepackage{amsmath}                   
\usepackage{amsthm}
\usepackage{amssymb}
\usepackage{latexsym} 

\usepackage{verbatim}

\usepackage[round]{natbib}

\usepackage{tikz}
\usetikzlibrary{decorations.markings}

\numberwithin{equation}{section} 

\newtheorem{my_theorem}{Theorem}[section]
\newtheorem{my_lemma}[my_theorem]{Lemma}

\title{Fast image segmentation and restoration using parametric curve
  evolution with junctions and topology changes}
\author{Heike Benninghoff\footnote{Deutsches Zentrum f\"ur Luft- und
    Raumfahrt (DLR), 82234 We\ss ling, Germany,
    heike.benninghoff@dlr.de} \,\, and 
        Harald Garcke\footnote{Fakult{\"a}t f{\"u}r Mathematik, Universit{\"a}t Regensburg, 93040 Regensburg, Germany, harald.garcke@ur.de}}

\date{}
\begin{document}
\maketitle

\begin{abstract} Curve evolution schemes for image segmentation based
  on a region based contour model allowing for junctions,
  vector-valued images and topology changes are introduced. Together
  with an a posteriori denoising in the segmented homogeneous regions
  this leads to a fast and efficient method for image segmentation and
  restoration. An uneven spread of mesh points is avoided by
  using the tangential degrees of freedom. Several numerical
  simulations on artificial test problems and on real images
  illustrate the performance of the method.
\end{abstract}



\section{Introduction}
\label{sec:intro}
Image segmentation and image smoothing are two fundamental tasks in
image processing. Image segmentation is the problem of partitioning
an image into constituent parts and to identify edges in a given
image. As image smoothing of the whole image typically leads to a
blurring of edges a possible strategy in image processing is to first
find homogeneous regions and in a second step to perform a smoothing
only in these homogeneous regions. 

Many different approaches for image segmentation have been
proposed. Active contours (or `snakes') have been first proposed by
\cite{Kass88} and have been later used by 
\cite{Caselles97} and \cite{Kichenassamy96}. The basic idea is
to evolve a curve in order to find local minima of a suitable energy
functional, and thus localize the object. The energy is chosen such
that small values are attained at edges. The original snake models by
\cite{Kass88} used an energy which was not
independent of the parameterization chosen for a given curve. In addition
it was argued that the inability to change topology was a significant
disadvantage of this original snake model. 

\cite{Caselles97} and \cite{Malladi95} 
 then used a geometric functional instead in
which the energy no longer depends on the curve parameterization. In
addition the level set framework of \cite{OsherSethian88} made
it possible to split and join curves and hence also topological
changes were possible. 

The above mentioned active contour models typically depend on the
gradient of the given image which is used to stop the evolution of the
curve and hence these models are restricted to objects for which edges
are defined by a gradient. However, there are regions whose boundaries
are badly defined through the gradient as e.g. smeared boundaries or
larger objects which are given by grouping smaller ones, see
e.g. \cite{AubertKornprobst06}. In addition, gradient based methods
heavily depend on noise and in particular in the presence of noise the
evolution often get trapped as several local maxima of the image
gradient exist. Later new region based active contour models have been
developed which can detect objects which boundaries are not
necessarily defined by the gradient of the image. These active
contour models are called active contours `without edges' and the main
new ingredient is that the information inside the detected regions is
used and not only information at their boundaries. We refer to
\cite{Chan01}, \cite{Ron} and \cite{Tsai01} for such approaches. The
approach of \cite{Chan01} uses a two-phase, piecewise constant
approximation of the functional of \cite{Mumford89} to partition a
given image into two phases, i.e. object and background. In their
approach a curve evolution in the context of a level set method is
used to solve the partition problem. Topological derivatives and
sensitivities have also been used to solve the Chan-Vese segmentation
problem, see e.g. \cite{HO} and \cite{HL}.  In principle, also
non-constant functions in the regions can be used and we refer to
\cite{Tsai01} who use the full Mumford-Shah functional to define a
region-based descent direction for the curve evolution which is also
solved in the level set context.

In principle the curve evolution problem in region based active
contour models can also be solved with the evolution of parametric
curves, see e.g. \cite{Dogan08}. As mentioned above the evolution of
curves does not allow for an automatic change of topology. However,
different authors proposed methods to handle topology changes, see
\cite{AYIT}, \cite{Dogan08} and the recent paper by \cite{NT}. 

The direct parametric approach often has a typical undesirable
behavior during the evolution, namely during the geometric flow some
points bunch together whereas other points drift apart. Due to this
uneven spread of points quantities like curvature and normal tangent are
often not computed to a high precision and also the segmentation is
not very accurate. This problem has been addressed in the context of
the Mumford-Shah formulation by \cite{CSW} who modified the usual
length constrained on the contour by taking a so-called diffusion
snake. As a result a hybrid model is obtained, which combines the
external energy of the Mumford-Shah functional with the internal
energy of the snake of \cite{Kass88}. However, this approach has the
disadvantage that it is not parameter free and also topology changes
have not been considered in \cite{CSW}. We also refer to \cite{SCRS}
who use the equidistribution strategy for the curve mesh points of
\cite{Mikula06} to stabilize the curve evolution by introducing a
highly nonlinear tangential evolution term for the mesh points. 
This method has been successfully used for edge driven active
contour models and for tracking problems, see \cite{SCRS}. 
Another important problem in image analysis is the segmentation of
vector-valued images. To the knowledge of the authors this problem has
not been tackled within the parametric approach. 

In this paper we introduce a novel parametric method for
segmentation and denoising of images which is based on a recent
approach of \cite{BGN07a,BGN07b} which was developed for the numerical
treatment of curvature driven geometric evolutions of curves and leads
to asymptotically equidistributed mesh points. The nearly
equidistributed spacing of mesh points is essential for a stable
contour evolution and for a robust and efficient detection of topology
changes.  

Let us state the {\it main features of the proposed approach}.

\begin{itemize}
\item {\it Image segmentation} and {\it denoising} of images can be
  dealt with by first using a Chan-Vese type approach, see
  \cite{Chan01,Chan00,Vese02}, to segment the image and a posteriori an
  image smoothing by minimizing the Mumford-Shah functional for a
  fixed contour set is applied.
\item The mesh properties of the discrete parametric curves stay good
  due to the tangential redistribution method of \cite{BGN07a,BGN07b}
  which keeps mesh points nearly {\it equidistributed}.
\item The possibility of {\it junctions} is included into the method. 
\item The approach can deal with 
{\it vector valued images}, i.e. e.g. colored images can be considered.
\item {\it Topological changes}, including topological 
changes involving junctions, are
  computed {\it efficiently} with an effort of $\mathcal{O}(N)$ where $N$ is
  the number of mesh points in the curve network. This is possible due to the
  equidistribution property together with a generalization of the
  approach of \cite{MU}. 
\item The approach has the advantage that for segmentation only
  one-dimensional geometric PDEs have to be solved which discretized
  versions can be solved very fast with a direct numerical algebra
  solver. 
\end{itemize}


\section{A model for image segmentation and denoising}
\label{sec:models}
Let $\Omega \subset \mathbb{R}^2$ be a rectangular image domain and
$u_0: \Omega \rightarrow [0,1]^d$ a given image function. In case of a
scalar, gray-scaled image $d=1$, $u_0$ describes the gray value, where
$0$ corresponds to black and $1$ corresponds to white. In case of a
color image $d=3$, the components of $u_0$ describe for example the
intensities of the red, green and blue part of the color.  We first
consider the scalar case $d=1$ and later state how color images can be
handled.

The objective of image segmentation is to identify objects and edges
shown in the image. The model of \cite{Mumford89} for optimal
approximation of images aims at finding a set of curves $\Gamma
= \Gamma_1 \cup ... \cup \Gamma_{N_C}$ and a piecewise smooth function
$u: \Omega \rightarrow \mathbb{R}$ approximating $u_0$ with possible
discontinuities across $\Gamma$. The energy to be minimized is
\begin{equation}
E(\Gamma,u) = \sigma |\Gamma | + \int_{\Omega \setminus \Gamma} \|\nabla u\|^2 \, \mathrm{d}x + \lambda \int_\Omega (u_0-u)^2 \,\mathrm{d}x,
\label{eq:mumford_shah}
\end{equation}
where $\sigma, \lambda > 0$ are weighting parameters, $|\Gamma|$
denotes the total length of the curves in $\Gamma$ and $\|\,.\,\|$ is
the Euclidean norm. The first term penalizes the length of the curves,
the second term does not allow $u$ to change too much in $\Omega \setminus
\Gamma$ and the third term requests that $u$ is a good approximation
of $u_0$. The general Mumford-Shah problem is difficult to solve
without restrictions, for example on the class of approximating
functions $u$. Similar to \cite{Chan01}, we consider for the
segmentation of the image a reduced
problem. We assume that the curves in $\Gamma$ partition the set
$\Omega$ in connected components and let $\Omega_1, \ldots, \Omega_{N_R}$
be the connected components of $\Omega \setminus \Gamma$. Further, let
each curve $\Gamma_i$ be the interface between two regions
$\Omega_{k^+(i)}$ and $\Omega_{k^-(i)}$ with $k^+(i), k^-(i) \in \{1,
\ldots, N_R\}$. Thus, we have the following decomposition of the
domain $\Omega$:
\begin{equation}
\Omega = \Omega_1 \cup \ldots \cup \Omega_{N_R} \cup \Gamma_1 \cup \ldots \cup \Gamma_{N_C}.
\label{eq:decomposition_omega}
\end{equation}

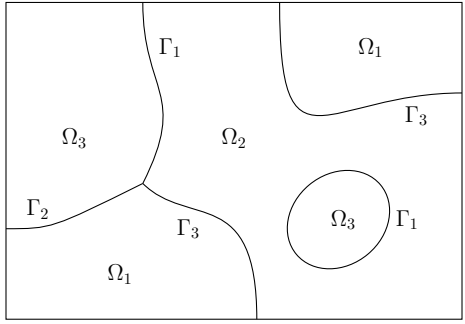
\begin{figure}[t]
\begin{center}
\begin{tikzpicture}[scale=0.6,transform shape]
\draw (0,0) rectangle (10,7);
\draw[rotate=40] (7,-3) ellipse (1.2 and 1);
\draw (3,3) .. controls (4,5) and (3,5) .. (3,7);
\draw (3,3) .. controls (1,2) and (1,2)  .. (0,2);
\draw (3,3) .. controls (4,2) and (5.5,3) .. (5.5,0);
\draw (6,7) .. controls (6,3) and (7,5) .. (10,5);

\draw (2.5,1) node{\Large $\Omega_1$};
\draw (1.5,4) node{\Large $\Omega_3$};
\draw (5,4) node{\Large $\Omega_2$};
\draw (7.4,2.2) node{\Large $\Omega_3$};
\draw (8,6) node{\Large $\Omega_1$};
\draw (3.6,6) node{\Large $\Gamma_1$};
\draw (0.7,2.43) node{\Large $\Gamma_2$};
\draw (4,2) node{\Large $\Gamma_3$};
\draw (8.8,2.2) node{\Large $\Gamma_1$};
\draw (9,4.5) node{\Large $\Gamma_3$};
\end{tikzpicture}
\end{center}
\caption{Example of a decomposition in regions $\Omega_k$ and interfaces $\Gamma_i$, $k=1,\ldots N_R=3$, $i=1, \ldots, N_C = 3$.}
\label{fig:decomposition}
\end{figure}

\noindent
Figure \ref{fig:decomposition} gives an example of a decomposition of a rectangular domain. The regions and interfaces can have more than one connected component. The interfaces can be closed, can meet at a triple junction and boundary intersections with $\partial \Omega$ can occur. 

In a first step, we restrict ourselves to piecewise constant functions $u$ of the form 
\begin{equation}
u = \sum_{k=1}^{N_R} c_k \chi_{\Omega_k},
\label{eq:piecewise_const_approx}
\end{equation}
where $\chi_{\Omega_k}$ denotes the characteristic function of $\Omega_k$. As $u$ is piecewise constant, we have $\nabla u = 0$ on $\Omega\setminus \Gamma$ and the functional \eqref{eq:mumford_shah} reduces to 
\begin{equation}
E(\Gamma,c_1,\ldots,c_{N_R}) = \sigma |\Gamma| + \lambda \sum_{k=1}^{N_R} \int_{\Omega_k} (u_0-c_k)^2\,\mathrm{d}x.
\label{eq:Mumford_Shah_const}
\end{equation}
The length term $\sigma |\Gamma|$ ensures that the curves $\Gamma_1, \ldots, \Gamma_N$ have finite length. The second term is a bulk energy, designed such that $\Omega_k$ best approximates the several objects in the image. The interfaces $\Gamma_i$ are attracted to the edges of objects in the image. 

In so-called region based active contours methods like \eqref{eq:mumford_shah} and \eqref{eq:Mumford_Shah_const}, only the raw image intensity function $u_0$ is involved in contrast to edge based active contours models which make use of the gradient of $u_0$ to detect edges of objects, cf. \cite{Kass88,Malladi95,Caselles97}. In particular, no prior smoothing of $u_0$ needs to be done when applying region based active contours models. Since the model does not depend on the gradient, noisy images and images with weak edges \citep{Chan01}, i.e. smooth transitions between gray values with small $\|\nabla u\|$, can be handled.

\subsection{Multi-phase image segmentation}
Fixing $\Gamma$ in \eqref{eq:Mumford_Shah_const} and considering a variation in $c_k$, $k \in \{1, \ldots, N_R\}$, leads to 
\begin{equation}
c_k = \frac{\int_{\Omega_k} u_0\,\mathrm{d}x}{\int_{\Omega_k} 1\,\mathrm{d}x},
\label{eq:c_k}
\end{equation}
i.e. $c_k \in \mathbb{R}$ is set to the mean of $u_0$ in $\Omega_k$. 

For a variation of the edges $\Gamma$, we consider time-dependent
curves and regions: Let $\Gamma_1(t), \ldots, $ $\Gamma_{N_C}(t)
\subset \Omega$, $t \in [0,T]$, be smooth, evolving curves. Let an
approximation $u(t)$ of $u_0$ be given by a piecewise constant
function with $u(t)_{|\Omega_k(t)} = c_k(t) \in \mathbb R$,
$k=1\ldots, N_R$, where $c_k(t)$ is set to the mean of $u_0$ in
$\Omega_k(t)$. The curve $\Gamma_i(t)$ should evolve in time such that
the energy \eqref{eq:Mumford_Shah_const} decreases most quickly.

We now introduce a representation of the curves by smooth
parameterizations. Let $\vec x_i: I_i \times [0,T] \rightarrow
\mathbb{R}^2$ be a smooth function, such that $\vec x_i(.,t)$ is a
smooth parameterization of $\Gamma_i(t)$. The set $I_i$ is a
one-dimensional reference manifold, e.g. $I_i = [0,1]$ for open
curves, i.e. curves with $\partial \Gamma_i(t) \neq \emptyset$, and
$I_i = S^1 \cong \mathbb{R} / \mathbb{Z}$ for closed curves,
i.e. curves with $\partial \Gamma_i(t) = \emptyset$. Further, we
define a normal vector field $\vec \nu_i(.,t)$ on $\Gamma_i(t)$ by $\vec
\nu_i : I_i \times [0,T] \rightarrow \mathbb{R}^2$ such that $\vec
\nu_i(\rho,t)$ is a normal on $\Gamma_i(t)$ at $\vec x_i(\rho,t)$. In
detail, we set $\vec \nu_i(\rho,t) =
\left(\begin{array}{rr}0&-1\\1&0\end{array}\right) (\vec
x_i)_s(\rho,t)$, where $s$ is the arc-length and $(\vec x_i)_s = (\vec
x_i)_\rho / \|(\vec x_i)_\rho\|$ denotes the derivative of $\vec x_i$
with respect to arc-length. We choose the parameterization such that
the normal vector field defines an orientation of $\Gamma_i(t)$ with
$\vec \nu_i(.,t)$ pointing from phase $\Omega_{k^-(i)}(t)$ to
$\Omega_{k^+(i)}(t)$.

Using methods from the calculus of variations one obtains the
following evolution law for the curves $\Gamma_i(t)$, see
e.g. \cite{Chan01,DeckelnickDziukElliott05,AubertKornprobst06}, 
\begin{equation}\label{eq:general_evolution_eq}
(V_n)_i=\sigma\kappa_i+F_i,
\end{equation}
where $(V_n)_i$ denotes the normal velocity and $\kappa_i$ the
curvature of the curve $\Gamma_i(t)$ and $F_i$ is an external forcing
term defined by 
\begin{equation}\label{}
F_i(t)=\lambda[(c_{k+(i)}(t)-u_0)^2-(c_{k-(i)}(t)-u_0)^2]\,,
\end{equation}
where $k^+(i)$ and $k^-(i)$ denote the indices of the two phases
separated by $\Gamma_i(t)$. 
 
The curvature term in \eqref{eq:general_evolution_eq} can be derived
by a variation of the length functional weighted with a constant
$\sigma > 0$. Using the parametric description the evolution equation
\eqref{eq:general_evolution_eq} can be rewritten as
\begin{subequations}
\label{eq:multiphase_scheme}
\begin{equation}
(\vec x_i)_t \,.\, \vec \nu_i = \sigma \kappa_i + F_i
\label{eq:scheme1}
\end{equation}
and the curvature $\kappa_i: I_i \times [0,T] \rightarrow \mathbb R$ is related to $\vec x_i$ by 
\begin{equation}
\kappa_i \vec \nu_i = (\vec x_i)_{ss}, \quad i=1,\ldots, N_C.
\label{eq:scheme2}
\end{equation} 
\end{subequations}
Without the external energy term, the evolution would
reduce to the well-known curvature flow. The term $F_i$ can be
derived by considering a variation of $\Gamma$ and using a transport
theorem for the external energy in \eqref{eq:Mumford_Shah_const}.

\subsection{Triple junctions and boundary intersection}
\label{subsec:tj_bi}
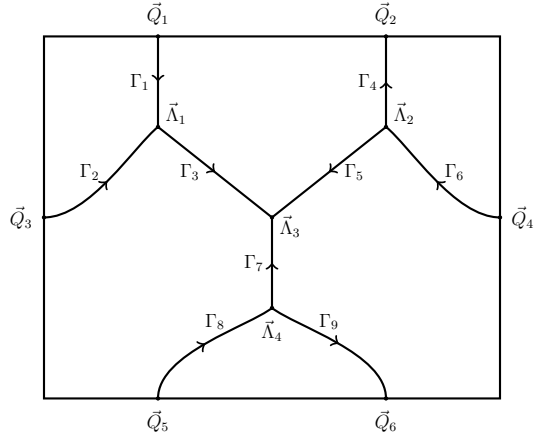
\begin{figure}[t]
\begin{center}
\begin{tikzpicture}[scale=0.6,transform shape]
\draw[thick] (0,0) rectangle (10,8);
\begin{scope}[thick,decoration={markings,mark=at position 0.5 with {\arrow{>}}}] 
\draw[postaction={decorate}] (2.5,6)--(5,4);   
\draw[postaction={decorate}](7.5,6)--(5,4);   
\draw[postaction={decorate}] (5,2)--(5,4);     
\draw[postaction={decorate}](2.5,8)--(2.5,6); 
\draw[postaction={decorate}](7.5,6) -- (7.5,8); 
\draw[postaction={decorate}](0,4) .. controls (1,4) and (2,5.6) .. (2.5,6);       
\draw[postaction={decorate}](10,4) .. controls (9,4) and (8,5.6) .. (7.5,6);      
\draw[postaction={decorate}](2.5,0) .. controls (2.5,1) and (4.5,1.6) .. (5,2);   
\draw[postaction={decorate}](5,2) .. controls (5.5,1.6) and (7.5,1) ..(7.5,0) ;   
\end{scope}
\filldraw (2.5,6) circle (1pt); 
\filldraw (5,4) circle (1pt);   
\filldraw (7.5,6) circle (1pt); 
\filldraw (5,2) circle (1pt);   
\filldraw (2.5,8) circle (1pt); 
\filldraw (7.5,8) circle (1pt); 
\filldraw (0,4) circle (1pt);   
\filldraw (10,4) circle (1pt) ; 
\filldraw (2.5,0) circle (1pt); 
\filldraw (7.5,0) circle (1pt); 

\draw (2.9,6.3) node{\large $\vec\Lambda_1$};
\draw (7.9,6.3) node{\large $\vec\Lambda_2$};
\draw (5.4,3.8) node{\large $\vec\Lambda_3$};
\draw (5,1.5) node{\large $\vec\Lambda_4$};
\draw (2.5,8.5) node{\large $\vec Q_1$};
\draw (7.5,8.5) node{\large $\vec Q_2$};
\draw (-0.5,4) node{\large $\vec Q_3$};
\draw (10.5,4) node{\large $\vec Q_4$};
\draw (2.5,-0.5) node{\large $\vec Q_5$};
\draw (7.5,-0.5) node{\large $\vec Q_6$};

\draw (2.1,7) node{\large $\Gamma_1$};
\draw (1,5) node{\large $\Gamma_2$};
\draw (3.2,5) node {\large $\Gamma_3$};
\draw (7.1,7) node{\large $\Gamma_4$};
\draw (6.8,5) node{\large $\Gamma_5$};
\draw (9,5) node {\large $\Gamma_6$};
\draw (4.6,3) node{\large $\Gamma_7$};
\draw (3.75,1.7) node{\large $\Gamma_8$};
\draw (6.25,1.7) node {\large $\Gamma_9$};
\end{tikzpicture}
\end{center}
\caption{Example of a curve network with triple junctions and boundary intersection points}
\label{fig:tp_bi}
\end{figure}

The methods presented above can be generalized to complex structures
of curves which involve triple junctions and boundary intersection
points. At these points, additional conditions need to be stated,
cf. \citep{BGN07a}. Figure \ref{fig:tp_bi} shows an example of a curve
network with triple and boundary intersection points.

Let $\Gamma_i$ be a non-closed curve with a smooth parameterization
$\vec x_i : I_i \rightarrow \mathbb{R}^2$ and $I_i = [0,1]$. Let $\vec
\Lambda_k \in \Omega$, $k=1, \ldots, N_T$, denote the triple
junctions. For each $k \in \left\{1, \ldots, N_T\right\}$ let
$i_{k,1}, i_{k,2}, i_{k,3} \in \{1, \ldots, N_C\}$ denote the indices
of curves $\Gamma_{i_{k,l}}$, $l=1,2,3$, $i_{k,1}\neq i_{k,2}\neq
i_{k,3} \neq i_{k,1}$, such that
\begin{equation*}
\vec{x}_{i_{k,1}}(\rho_{k,1}) = \vec{x}_{i_{k,2}}(\rho_{k,2})= \vec{x}_{i_{k,3}}(\rho_{k,3}) =\vec \Lambda_k,
\end{equation*}
where $\rho_{k,l} \in \{0,1\}$ corresponds to the start or end point of the curve $i_{k,l}$, $l=1,2,3$.

At the triple junctions $\vec \Lambda_k$, $k=1, \ldots, N_T$,  an attachment condition and Young's law need to hold: 
\begin{subequations}
\label{eq:triple_junction_cond}
\begin{align}
& \text{the triple junction $\vec \Lambda_k$ does not pull apart}, \label{eq:attachment_cond_tj}\\
& \sum_{l=1}^3 (-1)^{\rho_{k,l}} \,\vec\tau_{i_{k,l}} = 0, \label{eq:youngs_law}
\end{align}
\end{subequations}
where $\vec\tau_{i_{k,l}} := (\vec x_{i_{k,l}})_s$ is a tangent vector
field at $\Gamma_{i_{k,l}}$, $l=1,2,3$. The condition
(\ref{eq:youngs_law}) is equivalent to a $120^\circ$ angle condition
at triple junctions, see \cite{BGN07a}. 

Let $ \vec Q_k \in \partial\Omega$, $k=1, \ldots, N_I$, be the set of boundary intersection points. For $k=1, \ldots, N_I$ let $i_{I,k}$ denote the curve index and $\rho_{I,k} \in \{0,1\}$ such that $\vec x_{i_{I,k}}(\rho_{I,k}) =  \vec Q_k$. Let $\vec\tau_{i_{I,k}} := (\vec x_{i_{I,k}})_s$ be the corresponding tangent vector field at $\Gamma_{i_{I,k}}$. 
The following conditions need to hold at the boundary intersection points $\vec Q_k$, $k=1, \ldots, N_I$:
\begin{subequations}
\label{eq:boundary_intersections_cond} 
\begin{align}
& \text{the curve endpoint $\vec Q_k$ remains attached to } \partial \Omega, \label{eq:attachment_cond_bi}\\
& \vec\tau_{i_{I,k}}(\rho_{I,k})\,.\, \vec n_{\partial \Omega}(\vec Q_k)^\perp = 0, \label{eq:angle_cond}
\end{align}
\end{subequations}
where $\vec n_{\partial \Omega}$ is a normal vector field at the
boundary of the rectangular image domain $\Omega$. The first equation
is an attachment condition and the second equation enforces a $90$
degree angle condition at the boundary intersection point.

\subsection{Image smoothing with edge enhancement}
\label{subsec:image_smoothing}
The presented image segmentation method automatically provides a piecewise constant approximation of a possibly noisy image given by \eqref{eq:piecewise_const_approx}. For a variety of real images, a piecewise constant approximation poses a too large simplification. A piecewise smooth approximation can be found by reconsidering the Mumford-Shah functional \eqref{eq:mumford_shah}, with approximations $u_{|\Omega_k}=u_k$, $k=1,\ldots, N_R$, where $u_k \in C^1(\Omega_k, \mathbb{R})$. Fixing curves $\Gamma_1, \ldots, \Gamma_{N_C}$, the energy functional reduces to 
\begin{equation}
E^{MS,2}(u_1,\ldots, u_{N_R}) =  \sum_{k=1}^{N_R} \left(\lambda_k \int_{\Omega_k} (u_0-u_k)^2 \,\mathrm{d}x 
 + \int_{\Omega_k} \|\nabla u_k\|^2 \,\mathrm{d}x\right).
\label{eq:image_diffusion_functional}
\end{equation} 
In contrast to \eqref{eq:mumford_shah}, we also allow region-dependent parameters $\lambda_k > 0$. 
By considering variations of the form $u_k + \epsilon \eta$, $\epsilon \in (-\epsilon_0,\epsilon_0)$, $\eta \in C^1(\overline{\Omega_k}, \mathbb{R})$ in \eqref{eq:image_diffusion_functional}, we obtain using the classical methods from the theory of calculus of variations the following boundary value problem:

Find $u: \Omega \rightarrow \mathbb{R}$ such that 
\begin{subequations}
\label{eq:image_diffusion_scheme}
\begin{align}
- \frac{1}{\lambda_k} \Delta u + u &= u_0 && \text{ in } \Omega_k, \,\,k=1, \ldots, N_R, \\
\nabla u \,.\, \vec \nu_i &= 0 && \text{ on } \Gamma_i,\,\,i=1, \ldots, N_C, \\
\nabla u \,.\, \vec n_{\partial \Omega} &= 0 && \text{ on } \partial \Omega.
\end{align}
\end{subequations}
We separately solve the partial differential equations with Neumann boundary conditions on the regions $\Omega_1,$ $\ldots, \Omega_{N_R}$. The solution $u$ poses an approximation of the original image $u_0$. The parameter $\frac{1}{\lambda_k} > 0$ controls the smoothing effect of the Laplace operator. The bigger $\lambda_k$, the closer is the approximation to the original image. The smaller $\lambda_k$, the smoother is $u$. As we solve the equation separately in each region, the interfaces $\Gamma_i$ are not smoothed out. Consequently the edges of objects in the image remain sharp if the interfaces match with the edges. 

The system \eqref{eq:image_diffusion_scheme} is close to the scheme
proposed by \cite{Tsai01}, where the authors derived a level set method for two
phases from the Mumford-Shah functional by considering the
approximation $u$ as the optimal estimate of a stochastic process.

\subsection{Color images}
A color image is given as a vector-valued image function $\vec u_0=
(u_{0,1},u_{0,2},u_{0,3}): \Omega \rightarrow [0,1]^3$ where the
components of $\vec u_0$ for example 
denote the red, green and blue color channel (RGB) of the image. 
Using the RGB image data, we consider the following energy, cf. \cite{Chan00}:
\begin{equation}
E(\Gamma,\vec u) = \sigma |\Gamma| + \sum_{j=1}^3 \lambda_j \int_{\Omega} (u_{0,j}-u_{j})^2 \,\mathrm{d}x,
\end{equation}
where $\vec u=(u_1,u_2,u_3):\Omega \rightarrow \mathbb{R}^3$ is
piecewise constant, i.e. $\vec u_{|\Omega_k} =: \vec c_k = (c_{k,1},
c_{k,2}, c_{k,3})$, $k=1,\ldots,N_R$. Considering variations in the $c_{k,j}$
leads to
\begin{equation}
c_{k,j} =  \frac{\int_{\Omega_k} u_{0,j} \,\mathrm{d}x}{\int_{\Omega_k} 1 \,\mathrm{d}x}, \quad j=1,2,3, \,\, k=1,\ldots,N_R.
\end{equation}
Thus, each component of $\vec c_k$ is set to the mean of $u_{0,j}$ in $\Omega_k$. Considering a variation in $\Gamma$ leads to the evolution equation
\begin{subequations}
\begin{align}
(V_n)_i &= \sigma \kappa_i + F_i, \quad i=1, \ldots, N_C, \\
F_i &= \sum_{j=1}^3 \lambda_j \left[ (u_{0,j}-c_{k^+(i),j})^2 - (u_{0,j}-c_{k^-(i),j})^2 \right].
\end{align}
\end{subequations}

By modifying the weighting parameters $\lambda_j$, $j=1,2,3$, we can control the segmentation with respect to the red, green and blue part of the image. However for some images, one may make use of other color spaces which allow to consider a color's chromaticity, brightness and saturation component separately. Two possible spaces are the CB space (chromaticity-brightness) and the HSV color space (hue, saturation and value). The color spaces CB and HSV can be used for a variety of image processing tasks like segmentation, denoising or enhancement of color images, see \cite{Chan01_2}, \cite{Tang02}, \cite{Aujol06}. 

For an RGB image function $\vec u_0$, the chromaticity function $\vec v_0: \Omega \rightarrow S^2$ and the brightness function $b_0: \Omega \rightarrow \mathbb{R}$ are defined by 
\begin{equation}
\vec v_0 = \frac{\vec u_0}{\|\vec u_0\|}, \,\, b_0= \|\vec u_0\|.
\end{equation}
The energy to be minimized is
\begin{equation}
E(\Gamma, \vec v, b) = \sigma |\Gamma | +  \lambda_C \int_{\Omega} \| \vec v_0 - \vec v \|^2 \,\mathrm{d}x + \lambda_B \int_\Omega (b_0-b)^2 \,\mathrm{d}x,
\end{equation}
where $\vec v:\Omega \rightarrow \mathbb{R}^3$, $b:\Omega \rightarrow \mathbb{R}$ are piecewise constant, i.e. $\vec v_{|\Omega_k} = \vec v_k \in \mathbb{R}^3$, $b_{|\Omega_k} = b_k \in \mathbb R$, $k=1, \ldots, N_R$. The parameters $\sigma, \lambda_C, \lambda_B > 0$ weight the length, chromaticity and brightness term. Further, the constraint $\|\vec v\|=1$ needs to be satisfied. Considering variations in $b_k$ leads to 
\begin{equation}
b_k = \frac{\int_{\Omega_k} b_0 \,\mathrm{d}x}{\int_{\Omega_k} 1 \,\mathrm{d}x},
\end{equation}
i.e. $b_k\in\mathbb{R}$ is set to the mean of the brightness $b_0 = \|\vec u_0\|$ in $\Omega_k$. We now consider a variation of $\vec v$. This leads to a constrained minimization problem where the constraint $\|v_k\| = 1$ has to be enforced. This leads to the following computation of $\vec v_k$:
\begin{equation}
\vec V_k = \int_{\Omega_k} \vec v_0 \,\mathrm{d}x, \quad \vec v_k = \frac{\vec V_k}{\| \vec V_k\|}.
\end{equation}
Finally, fixing $\vec v$ and $\vec b$ and considering a variation of $\Gamma$ leads to the evolution equation
\begin{subequations}
\begin{align}
(V_n)_i &= \sigma \kappa_i + F_i, \quad i=1, \ldots, N_C, \\
F_i &= \lambda_C \left[ \|\vec v_0-\vec v_{k^+(i)}\|^2 - \|\vec v_0-\vec v_{k^-(i)}\|^2 \right] + \lambda_B \left[(b_0-b_{k^+(i)})^2 - (b_0-b_{k^-(i)})^2 \right].
\end{align}
\end{subequations}

In the HSV space, a color is given by three components: The periodical hue component describes the chromaticity ranging from red to yellow, green, cyan, blue, magenta and back to red. The saturation ranges from $0$ to $1$. A gray color (all three components of the corresponding RGB color are equal) has a saturation value of $0$. A saturation of $1$ is a maximum saturated color, i.e. it has one RGB value equal to zero and is therefore a mixture of only two RGB basic colors. The value component describes the luminosity of the color and is set to the maximum of the red, green or blue value. 

Let $\vec h_0: \Omega \rightarrow S^1 \subset \mathbb{R}^2$, $s_0: \Omega \rightarrow[0,1]$ and $v_0: \Omega \rightarrow [0,1]$ be the HSV components corresponding to an RGB image function $\vec u_0$. The energy to be minimized is 
\begin{equation}
E(\Gamma,\vec h, s, v) = \sigma |\Gamma| + \lambda_H \int_\Omega \|\vec h_0 - \vec h\|^2 \mathrm{d}x + \lambda_S \int_\Omega (s_0-s)^2 \mathrm{d}x + \lambda_V \int_\Omega (v_0-v)^2 \mathrm{d}x,
\end{equation}
where $\vec h:\Omega \rightarrow \mathbb{R}^2$, $s:\Omega \rightarrow [0,1]$, $v:\Omega \rightarrow [0,1]$ are piecewise constant, i.e. $\vec h_{|\Omega_k} =: \vec h_k \in \mathbb{R}^2$, $s_{|\Omega_k} =:  s_k \in \mathbb{R}$, $v_{|\Omega_k} =:  v_k \in \mathbb{R}$. The constraint $\|\vec h\| = 1$ needs to be satisfied.  The parameters $\sigma, \lambda_H, \lambda_S, \lambda_V > 0$ weight the length, hue, saturation and value term. Similar as above, considering variations of the single components $\vec h_k$, $s_k$, $v_k$ leads to 
\begin{equation}
\vec H_k = \int_{\Omega_k} \vec h_0\,\mathrm{d}x, \quad\vec h_k = \frac{ \vec H_k}{\| \vec H_k\|}, \quad s_k = \frac{\int_{\Omega_k} s_0\,\mathrm{d}x}{\int_{\Omega_k} 1\,\mathrm{d}x}, \quad v_k = \frac{\int_{\Omega_k} v_0\,\mathrm{d}x}{\int_{\Omega_k} 1\,\mathrm{d}x},
\end{equation}
where a constrained minimization problem has to be solved for $\vec h_k$. 

Fixing $\vec h$, $s$ and $v$ and considering a variation of $\Gamma$ leads to the evolution equation
\begin{subequations}
\begin{align}
(V_n)_i &= \sigma \kappa_i + F_i, \quad i=1, \ldots, N_C, \\
F_i &= \lambda_H \left[ \|\vec h_0-\vec h_{k^+(i)}\|^2 - \|\vec h_0-\vec h_{k^-(i)}\|^2 \right] + \lambda_S \left[(s_0-s_{k^+(i)})^2 - (s_0-s_{k^-(i)})^2 \right] \nonumber \\
&\quad +  \lambda_V\left[(v_0-v_{k^+(i)})^2 - (v_0-v_{k^-(i)})^2 \right].
\end{align}
\end{subequations}
To sum up, only the external forcing term $F_i$ needs to be adapted for color images. The parametric scheme \eqref{eq:multiphase_scheme} can be used for both scalar and vector-valued images.

\section{Numerical Approximation}
\label{sec:numerics}
\subsection{Finite difference approximation}
\label{subsec:fd_appr}
We introduce a finite difference approximation for the equations
\eqref{eq:multiphase_scheme} with \eqref{eq:triple_junction_cond} and
\eqref{eq:boundary_intersections_cond} in case of triple junctions and
boundary intersections. The approach follows ideas of
\cite{BGN07a,BGN07b} who consider curvature flows and formulate the
discretization in a finite element context. 

For $i=1, \ldots, N_C$, let $0=q_0^i < q_1^i < \ldots < q_{N_i}^i = 1$
be a decomposition of the interval $I_i$. If $\Gamma_i$ is a closed
curve, we make use of the periodicity $N_i=0$, $N_i+1=1$, $-1=N_i-1$,
etc. Let $0 = t_0 < t_1 < \ldots < t_M = T$ be a partitioning of the
time interval $[0,T]$ into possibly variable time steps $\tau_m :=
t_{m+1} - t_m$, $m=0, \ldots, M-1$.

Let $\kappa^m = (\kappa_1^m, \ldots \kappa_{N_C}^m) \in
C(I_1,\mathbb{R}) \times \ldots \times C(I_{N_C},\mathbb{R})$ be an
approximation of $\kappa(.,t_m)=(\kappa_1(.,t_m), \ldots,
\kappa_{N_C}(.,t_m))$ and $\vec X^m = (\vec X_1^m, \ldots, \vec
X_{N_C}^m) \in C(I_1,\mathbb{R}^2)\times \ldots \times
C(I_{N_C},\mathbb{R}^2)$ an approximation of $\vec x(.,t_m) = (\vec
x_1(.,t_m), \ldots, \vec x_{N_C}(.,t_m))$ such that $\kappa_i^m$ and
$\vec X_i^m$ are piecewise linear on $[q_{j-1}^i,q_j^i]$, $i=1,
\ldots, N_C$, $j=1, \ldots, N_i$. Further, we make use of the short
hand notations
\begin{equation*}
  \kappa_{i,j}^m := \kappa_i^m(q_j^i), \quad\vec X_{i,j}^m := \vec X_i^m(q_j^i), \quad h_{i,j-\frac12}^m := \| \vec X_{i,j}^m - \vec X_{i,j-1}^m\|.
\end{equation*}
Let $h^m := \max_{i=1,\ldots,N_C, \,j=1,\ldots, N_i}
h_{i,j-\frac12}^m$ be the maximal distance between two neighboring
nodes of the polygonal curves. Let $\vec \nu^m := (\vec \nu_1^m,
\ldots, \vec \nu_{N_C}^m)$ such that $\vec \nu_i^m$, given by
\begin{equation*}
\vec\nu_i^m|_{[q_{j-1}^i, q_j^i]} := \vec\nu_{i,j-\frac12}^m :=\frac{\left(\vec X_{i,j}^m - \vec X_{i,j-1}^m\right)^\perp}{h_{i,j-\frac12}^m},
\end{equation*}
is a discrete normal vector field on $\Gamma_i^m$, $i=1, \ldots,
N_C$. We define the following weighted approximating normal vector at
$\vec X_{i,j}^m$ by
\begin{equation}
\vec\omega_{i,j}^m := \frac{h_{i,j-\frac12}^m \vec\nu_{i,j-\frac12}^m + h_{i,j+\frac12}^m \vec\nu_{i,j+\frac12}^m}{h_{i,j-\frac12}^m  + h_{i,j+\frac12}^m}
= \frac{\left(\vec X_{i,j+1}^m - \vec X_{i,j-1}^m\right)^\perp}{h_{i,j-\frac12}^m  + h_{i,j+\frac12}^m}, 
\end{equation}
for $j=1,\ldots,N_i$ if $\partial \Gamma_i^m = \emptyset$ and for $j=1, \ldots, N_i-1$ if $\partial\Gamma_i^m \neq \emptyset$. 
In the latter case, we set 
\begin{equation}
\vec\omega_{i,0}^m := \vec\nu_{i,\frac12}^m = \frac{(\vec X_{i,1}^m-\vec X_{i,0}^m)^\perp}{h_{i,\frac12}^m},\quad\quad 
\vec\omega_{i,N_i}^m := \vec\nu_{i,N_i-\frac12}^m = \frac{(\vec X_{i,N_i}^m-\vec X_{i,N_i-1}^m)^\perp}{h_{i,N_i-\frac12}^m}. 
\end{equation}

As $\kappa_i^{m+1}$ and $\vec X_i^{m+1}$ are piecewise linear, they are uniquely defined by their values at the nodes $q_j^i$. Therefore, we consider $\kappa^{m+1}$ and $\vec X^{m+1}$ as elements in $\mathbb R^N$ and $(\mathbb{R}^2)^{N}$ with $N=\sum_{i=1}^{N_C} N_i^*$, where $N_i^*=N_i$ for closed curves and $N_i^* = N_i+1$ for open curves. Similarly, we consider $\vec X^m \in (\mathbb R^2)^N$ and set $\delta\vec X^{m+1}:= \vec X^{m+1} - \vec X^m \in (\mathbb R^2)^N$. 

An approximation for \eqref{eq:scheme1} is given by 
\begin{equation}
\frac{1}{\tau_m} \,\left(\delta\vec X_{i,j}^{m+1}\right) \,.\, \vec\omega_{i,j}^m = \sigma \kappa_{i,j}^{m+1} + F_{i,j}^m
\label{eq:discrete_scheme1}
\end{equation}
for $i=1,\ldots,N_C$, $j=1, \ldots, N_i$ if $\partial \Gamma_i^m = \emptyset$ and $j=0,1,\ldots, N_i$, if $\partial \Gamma_i^m \neq \emptyset$. Here, $F_{i,j}^m := F_i(\vec X_{i,j}^m)$ denotes the external forcing term evaluated at the node $\vec X_{i,j}^m$. 

In order to propose a finite difference approximation of \eqref{eq:scheme2}, we need to define an approximation of 
 $\vec x_{ss}(q_j^i, t_{m+1})$. For $i=1, \ldots, N_C$ and $j=1, \ldots, N_i$, if $\Gamma_i^m$ is closed, and $j=1, \ldots, N_i-1$, if $\Gamma_i^m$ is not closed, we set
\begin{equation}
\Delta_2^{h,m} \vec X_{i,j}^{m+1} :=  \frac{2}{h_{i,j-\frac12}^m  + h_{i,j+\frac12}^m} \left(  \frac{\vec X_{i,j+1}^{m+1} - \vec X_{i,j}^{m+1}}{h_{i,j+\frac12}^m}-\frac{\vec X_{i,j}^{m+1} - \vec X_{i,j-1}^{m+1}}{h_{i,j-\frac12}^m}\right).
\end{equation}
In case of equal spatial step sizes $h_{i,j-\frac12}^m =
h_{i,j+\frac12}^m =: h_i^m$, the term reduces to $(\vec X_{i,j-1}^m -
2 \vec X_{i,j}^m + \vec X_{i,j+1}^m)/((h_i^m)^2)$.

An approximation of \eqref{eq:scheme2} is  given by 
\begin{equation}
\kappa_{i,j}^{m+1} \vec\omega_{i,j}^m = \Delta_2^{h,m} \vec X_{i,j}^{m+1},
\label{eq:discrete_scheme2}
\end{equation}
for $i=1,\ldots,N_C$, $j=1, \ldots, N_i$ in case of closed curves and $j=1, \ldots, N_i-1$ in case of open curves.

In case of $\partial \Gamma_i^m \neq \emptyset$, the boundary points
belong either to a triple junction $\vec\Lambda_k$, $k\in\{1, \ldots,
N_T\}$, or to a boundary intersection point $\vec Q_k \in \partial
\Omega$, $k\in\{1, \ldots, N_I\}$.  Let $i_1, i_2, i_3$ denote the
indices of three curves meeting at a triple point $\vec\Lambda \in
\{\vec\Lambda_1, \ldots, \vec \Lambda_{N_T}\}$. For the ease of
illustration, assume that the initial points of the curves meet at the
junction. The attachment condition \eqref{eq:attachment_cond_tj}
enforces
\begin{equation}
\vec X_{i_1,0}^{m+1} = \vec X_{i_2,0}^{m+1} = \vec X_{i_3,0}^{m+1}.
\label{eq:attachment_triple}
\end{equation}

Let $\vec Q_k \in \partial\Omega$ be a boundary intersection point and
$i$ a curve index such that $\vec Q_k = \vec X_{i,0}^m$. The
attachment condition \eqref{eq:attachment_cond_bi} is approximated by
\begin{equation}
\left(\vec X_{i,0}^{m+1} - \vec X_{i,0}^m\right)  \,.\, \vec n_{\partial \Omega}(\vec X_{i,0}^m) = 0.
\label{eq:attachment_boundary}
\end{equation}
If the equation above is satisfied, the curve will only move in direction tangential to $\partial\Omega$. As the image domain $\Omega$ is rectangular, $\partial \Omega$ is locally flat and $\vec Q_k$  keeps attached to $\partial \Omega$.

The considerations above motivate to define an operator
\begin{align*}
\vec P: (\mathbb R^2)^{N} \rightarrow \mathbb X :=& \left\{ (\vec z_1, \ldots, \vec z_{N_C}) \in (\mathbb R^2)^{N} \,:\, [\vec z_{i_{k,1}}]_{j_{k,1}} = [\vec z_{i_{k,2}}]_{j_{k,2}} = [\vec z_{i_{k,3}}]_{j_{k,3}}, \,k=1, \ldots N_T,\right. \\
&\left.\text{ and }\,  [\vec z_{i_{I,k}}]_{j_{I,k}} \,.\,  \vec n_{\partial \Omega}(\vec X_{i_{I,k},j_{I,k}}^m) = 0,\,k=1,\ldots, N_I\right\},
\end{align*}
which is the orthogonal projection onto $\mathbb X$. In the definition of $\mathbb X$, $\vec z_i \in (\mathbb R^2)^{N_i^*}$ and $[\vec z_i]_j \in \mathbb{R}^2$ is the $j$-th component of the vector $\vec z_i$. We use an index notation according to the notation introduced in Section \ref{subsec:tj_bi}. 

In order to state a matrix formulation for the discrete system
\eqref{eq:discrete_scheme1} and \eqref{eq:discrete_scheme2}, we
introduce the following matrices
\begin{equation*}
M:= \left(
\begin{array}{ccc}
M^1 & \cdots & 0 \\
\vdots &  \ddots & \vdots \\
0 & \ldots & M^{N_C} 
\end{array}
\right),
\vec N:= \left(
\begin{array}{ccc}
\vec N^1  & \cdots & 0 \\
\vdots  & \ddots & \vdots \\
0 &  \ldots & \vec N^{N_C} 
\end{array}
\right),
\vec A:= \left(
\begin{array}{ccc}
\vec A^1  & \cdots & 0 \\
\vdots &  \ddots & \vdots \\
0 &  \ldots & \vec A^{N_C} 
\end{array}
\right),
\end{equation*}
where $M^i \in \mathbb{R}^{N_i^* \times N_i^*}$, $\vec N^i \in (\mathbb{R}^2)^{N_i^* \times N_i^*}$, $\vec A^i \in (\mathbb{R}^{2\times 2})^{N_i^* \times N_i^*}$, $i=1, \ldots, N_C$, are defined by 
\begin{align*}
M_{j,l}^i &= \frac12 (h_{i,j-\frac12}^m +h_{i,j+\frac12}^m) \,\delta_{jl},  &
\vec N_{j,l}^i &= \frac12 (h_{i,j-\frac12}^m +h_{i,j+\frac12}^m) \vec\omega_{i,j}^m \,\delta_{jl}, \\
\vec A_{j,j}^i &= (\frac{1}{h_{i,j-\frac12}^m} +\frac{1}{h_{i,j+\frac12}^m}) \,\vec{\mathrm{Id}}_{2 \times 2}, &
\vec A_{j,j-1}^i &= -\frac{1}{h_{i,j-\frac12}^m} \,\vec{\mathrm{Id}}_{2 \times 2},  \\
\vec A_{j,j+1}^i &= -\frac{1}{h_{i,j+\frac12}^m} \,\vec{\mathrm{Id}}_{2 \times 2},  &
\vec A_{j,l}^i &= \vec 0, \, \text{ for } l\not \in \{j-1,j,j+1\},
\end{align*}
if $j \not\in \{0, N_i\}$ for open curves. For open curves, we set in addition
\begin{align*}
M_{0,l}^i &= \frac12 h_{i,\frac12}^m \,\delta_{0l},  &
\vec N_{0,l}^i &= \frac12 h_{i,\frac12}^m  \vec\omega_{i,0}^m \,\delta_{0l}, \\
\vec A_{0,0}^i &= \frac{1}{h_{i,\frac12}^m} \,\vec{\mathrm{Id}}_{2 \times 2}, &
\vec A_{0,1}^i &= -  \frac{1}{h_{i,\frac12}^m} \,\vec{\mathrm{Id}}_{2 \times 2},  \\
M_{N_i,l}^i &= \frac12 h_{i,N_i-\frac12}^m \,\delta_{N_i\,l},  &
\vec N_{N_i,l}^i &= \frac12 h_{i,N_i-\frac12}^m  \vec\omega_{i,N_i}^m \,\delta_{N_i\,l}, \\
\vec A_{N_i,N_i}^i &=  \frac{1}{h_{i,N_i-\frac12}^m} \,\vec{\mathrm{Id}}_{2 \times 2}, &
\vec A_{N_i,N_i-1}^i &=   -\frac{1}{h_{i,N_i-\frac12}^m} \,\vec{\mathrm{Id}}_{2 \times 2},  \\
\vec A_{0,l}^i &= \vec 0, \, \text{ for } l\not \in \{0,1\}, &
\vec A_{N_i,l}^i &= \vec 0, \, \text{ for } l\not \in \{N_i,N_i-1\}.
\end{align*}
In the terms above, $\vec{\mathrm{Id}}_{2 \times 2} \in \mathbb R^{2 \times 2}$ denotes the identity matrix, $\vec 0\in \mathbb R^{2 \times 2}$ a matrix with all entries equal to zero and $\delta_{jl}$ denotes the Kronecker symbol. 

Further, we define $b^m = (b_1^m, \ldots, b_{N_C}^m) \in \mathbb{R}^{N}$ by 
\begin{equation}
b_{i,j}^m =  \frac12 (h_{i,j-\frac12}^m +h_{i,j+\frac12}^m) \,F_{i,j}^m, \,\,\text{ resp. }\,\,b_{i,j}^m =  \frac12 h_{i,j\pm\frac12}^m \, F_{i,j}^m,
\label{eq:righthandside}
\end{equation}
where the latter term holds in case of boundary nodes. 

We propose the following linear system which includes the discrete scheme \eqref{eq:discrete_scheme1} and \eqref{eq:discrete_scheme2}, the attachment conditions \eqref{eq:attachment_triple} and \eqref{eq:attachment_boundary} and approximations of Young's law at triple junctions and of the  90 degrees angle condition at boundary intersection points: Find $\kappa^{m+1} \in \mathbb{R}^{N}$ and $\delta X^{m+1} \in \mathbb X$ such that
\begin{equation}
\left(
\begin{array}{cc}
-\sigma \tau_m M & \vec N^T \vec P \\
\vec P \vec N & \vec P \vec A \vec P 
\end{array}
\right) \left(
\begin{array}{c}
\kappa^{m+1} \\
\delta \vec X^{m+1} 
\end{array}
\right) = \left( 
\begin{array}{c}
\tau_m b^m \\
-\vec P \vec A \vec X^m
\end{array} \right).
\label{eq:linear_system}
\end{equation}
The attachment conditions at triple junctions and boundary intersection points are satisfied by $\delta \vec X^{m+1} \in \mathbb{X}$ on assuming that $\vec X^0$ fulfills the attachment condition at triple junctions. 
Recall, that $\vec P^T = \vec P$ and $\vec P$ is the identity on $\mathbb{X}$. Approximations of Young's law and the angle condition at boundary intersection points can be derived from the second equation in \eqref{eq:linear_system}. 
We now consider a triple junction $\vec\Lambda \in \{\vec \Lambda_1, \ldots, \vec\Lambda_{N_T}\}$ with $\vec X_{i_1,0}^{m+1}=\vec X_{i_2,0}^{m+1}=\vec X_{i_3,0}^{m+1}= \vec\Lambda$. The second equation of the system \eqref{eq:linear_system} provides
\begin{equation}
\sum_{l=1}^3 \frac12 h_{i_l,\frac12}^m \,\vec\omega_{i_l,0}^m \,\kappa_{i_l,0}^{m+1} + \sum_{l=1}^3 \frac{1}{h_{i_l,\frac12}^m}\left(\vec X^{m+1}_{i_l,0}- \vec X^{m+1}_{i_l,1}\right)= 0.
\label{eq:youngs_law_discr}
\end{equation}
For $h^m \rightarrow 0$, the first sum approaches zero, whereas the second sum approaches $-(\vec\tau_{i_1}(\vec \Lambda) +  \vec\tau_{i_2}(\vec \Lambda) +\vec\tau_{i_3}(\vec \Lambda))$. In the limit, we therefore can derive Young's law from \eqref{eq:youngs_law_discr}
\begin{equation}
\vec\tau_{i_1}(\vec \Lambda) +  \vec\tau_{i_2}(\vec \Lambda) +\vec\tau_{i_3}(\vec \Lambda) = 0.
\end{equation}
Consider a boundary intersection point $\vec Q \in \{\vec Q_1, \ldots, \vec Q_{N_I}\}$.  Let $i$ be a curve index such that $\vec X_{i,0}^{m}=\vec Q$. Then, the second equation in \eqref{eq:linear_system} and the definition of $\vec P$ provides
\begin{equation}
(\vec{\mathrm{Id}}_{2 \times 2} - \vec n_{\partial\Omega}(\vec Q) \otimes \vec n_{\partial\Omega}(\vec Q)) \left(\frac12 h_{i,\frac12}^m \,\vec\omega_{i,0}^m \,\kappa_{i,0}^{m+1} + \frac{1}{h_{i,\frac12}^m}\left(\vec X^{m+1}_{i,0}- \vec X^{m+1}_{i,1}\right)\right)= 0.
\end{equation}
The limit of these terms, as $h^m$ approaches $0$, is
\begin{equation}
\vec\tau_i(\vec Q) - \left(\vec \tau_i(\vec Q) \,.\, \vec n_{\partial\Omega}(\vec Q)\right) \, \vec n_{\partial\Omega}(\vec Q) = 0.
\end{equation}
Thus, $\vec \tau_i(\vec Q)$ is parallel to $\vec n_{\partial\Omega}(\vec Q)$. Consequently, the curve meets the external boundary with a 90 degrees angle at $\vec Q$. 

As $M$ is non-singular, \eqref{eq:linear_system} can be reformulated to 
\begin{subequations}
\begin{align}
\kappa^{m+1} & = \frac{1}{\sigma \tau_m} M^{-1} \left( \vec N^T \vec P \,\delta \vec X^{m+1} - \tau_m b^m\right), \\
\left( \vec P \vec A \vec P + \frac{1}{\sigma \tau_m} \vec P \vec N M^{-1} \vec N^T \vec P \right) \delta \vec X^{m+1} &= \frac{1}{\sigma } \vec P \vec N M^{-1} b^m - \vec P \vec A \vec X^m,\label{eq:schur}
\end{align}
\end{subequations}
by applying a Schur complement approach. The linear equation
\eqref{eq:schur} can be solved with an iterative solver, for example
with the method of conjugate gradients with possible preconditioning,
or with a direct solver for sparse matrices. 

Under very mild assumptions one can show that the system matrix in \eqref{eq:schur} is symmetric, positive definite and hence a unique solution of \eqref{eq:schur} exists. A detailed proof of existence and uniqueness is given in \cite*{BGN07a}.

\subsection{Semi-discrete scheme}
We now consider a scheme which is discrete in space and continuous in
time. Therefore let $\vec X = (\vec X_1, \ldots, \vec X_{N_C}) $,
$\kappa = (\kappa_1, \ldots, \kappa_{N_C})$, such that $\vec X_i : I_i
\times [0,T] \rightarrow \mathbb{R}^2$, and $\kappa_i: I_i \times
[0,T] \rightarrow \mathbb{R}$, $i=1, \ldots, N_C$, are piecewise
linear on $[q_{j-1}^i, q_j^i]$, $j=1, \ldots, N_i$. We make use of a
similar notation as in the fully discrete case by just omitting the
superscripts $m$ and $m+1$. All quantities are time-dependent in this
section. We now state a result which demonstrates that the
semi-discrete scheme leads to equidistributed meshes.

\begin{my_lemma}
The semi-discrete scheme
\begin{subequations}
\label{eq:semi-discrete_scheme}
\begin{align}
(\vec X_{i,j})_t \,.\, \vec\omega_{i,j} &= \sigma \kappa_{i,j} + F_{i,j}, \\
\kappa_{i,j} \,\vec\omega_{i,j} &= \Delta_2^{h} \vec X_{i,j}, \label{eq:semi2}
\end{align}
\end{subequations}
for $i=1, \ldots, N_C$, $j=1, \ldots, N_i$, if $\Gamma_i$ is open and $j=1, \ldots, N_i-1$, if $\Gamma_i$ is closed, provides an equidistribution of the mesh points along those parts of the curves $\Gamma_i$ which are not locally flat. 
\end{my_lemma}
The proof is given in \cite{BGN07b} (Remark 2.3). 
For the fully discrete scheme \eqref{eq:discrete_scheme1} and \eqref{eq:discrete_scheme2} we cannot prove the equidistribution of mesh points. However practical experiments show that the nodes are good distributed and therefore no \textit{local} refinement or coarsening strategy needs to be developed. As a curve can grow and shrink during the evolution, a \textit{global} refinement or coarsening is recommended. For that, the ratio of curve length and number of node points is considered.  

\subsection{Topological changes}
\label{subsec:topchange}
During the evolution of curves the following topological changes can
occur: (1) Splitting of a curve into two curves, (2) merging of two
curves to one single curve, (3) emergence of a triple junction and (4)
emergence of a boundary intersection point. If a triple junction is
detected, a new curve will be created. Similarly, if a curve length
shrinks, a curve has to be deleted. If the curve connects two triple
junctions, the triple junctions will also be deleted and other curves
which previously ended at the junctions have to be adapted. Therefore,
the number of curves $N_C^m$, the number of nodes $N_i^m$, $i=1,\ldots
N_C^m$, the number of triple junctions $N_T^m$ and of boundary
intersection points $N_I^m$ are time dependent.

Using a parametric approach, topological changes are not automatically handled in contrast to level-set methods. Rather, topological changes can be interpreted as singularities. We propose a method how to continue after such singularities occur. The detection of the topological changes used in this paper is close to the method proposed by \cite{Balazovjech12} who consider the case without triple junctions and boundary intersections. We generalize their approach such that also junctions and boundary points can be dealt with. 
 
\begin{figure}
\centering
\includegraphics[viewport = 120 480 430 740, width = 0.48\textwidth]{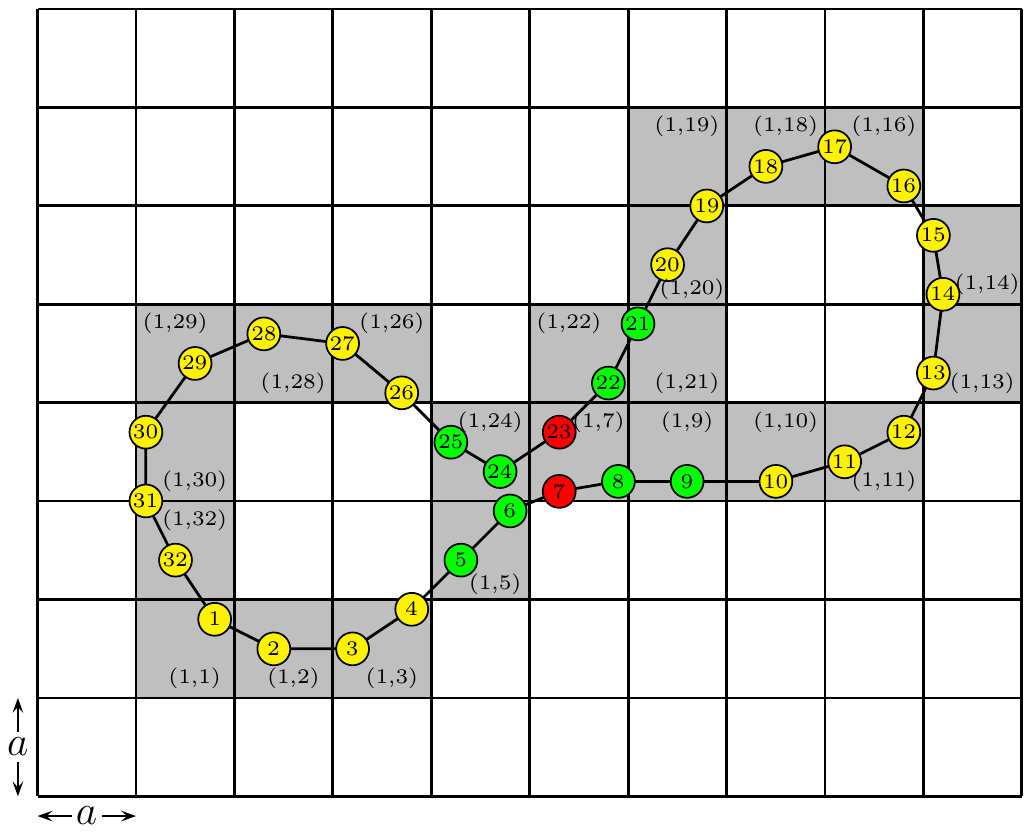}
\includegraphics[viewport = 120 480 430 740, width = 0.48\textwidth]{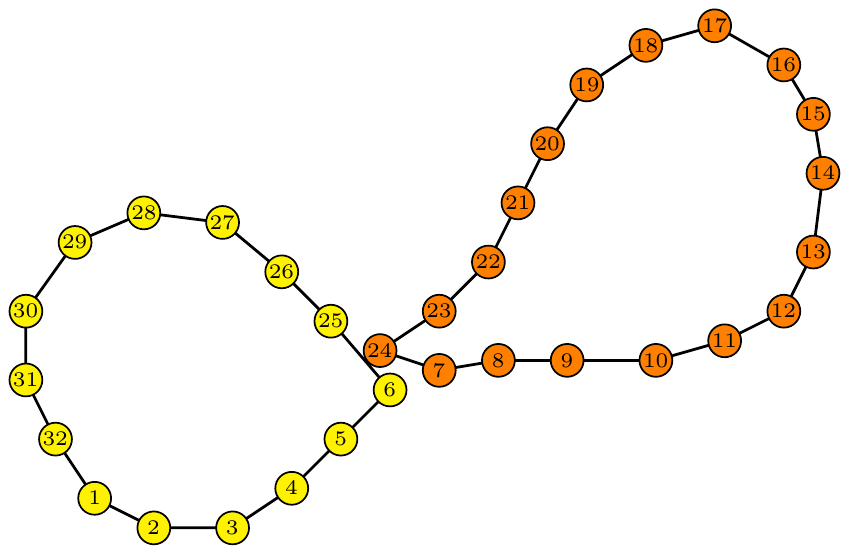}
\caption{Left: Detection of splitting of a curve $\Gamma_1^m$ near the nodes 7 and 23, Right: Splitting of the curve in two curves}
\label{fig:topchange}
\end{figure}

To detect a change in topology, we construct a uniform background grid which covers the rectangular image domain $\Omega$. The grid size $a>0$ can be adaptively chosen, for example it can be set according to the minimum or average distance between two neighboring node points of the polygonal curves. The grid can be stored as a sparse matrix or two dimensional array where the elements correspond to squares of the grid of size $a \times a$. Using a background grid, topological changes can efficiently be detected in two steps: 
\begin{enumerate}
\item For $i=1, \ldots, N_C^m$ and $j=0, \ldots, N_i^m$, we initialize the square in which $\vec X_{i,j}^m$ lies with $(-1,-1)$. If the square is close to the boundary of $\partial \Omega$, the node is marked as a boundary intersection point. Therefore it is stored in a list for boundary intersection points such that we can handle several topological changes at once. 
\item In a second loop, we again consider for $i=1, \ldots, N_C^m$ and $j=0, \ldots, N_i^m$ the corresponding square of the grid. If the square is marked with $(-1,-1)$, the square marking is overwritten with $(i,j)$. If the square has already been marked with $(i_1,j_1)$, a topological change is likely to occur close to the nodes. If $i=i_1$ and if the two nodes are direct neighbor points or have a common neighbor, no topological change is detected. Otherwise, a set of a limited number $n$ of neighbor nodes around each $\vec X_{i,j}^{m}$ and $\vec X_{i_1,j_1}^{m}$ is considered. As the two sets contain only $n$ nodes, the pair $(i,j^*)$ and $(i_1,j_1^*)$ with the smallest distance can be quickly found. In practice, $n=5$ is a good choice. If $i=i_1$, a splitting of the curve $\Gamma_i^m$ occurs and the pair $(i,j^*)$ and $(i_1,j_1^*)$ is stored in a list for splitting points. If $i\neq i_1$, we consider $k^+(i),k^-(i),k^+(i_1)$ and $k^-(i_1)$, i.e. the indices of the regions $\Omega_k$, $k\in \{1, \ldots, N_R\}$, separated by $\Gamma_i^m$ and $\Gamma_{i_1}^m$, respectively. Recall, that the normal $\vec\nu_i^m$ at the curve $\Gamma_i^m$ points from $k^-(i)$ to $k^+(i)$. If $k^+(i)=k^+(i_1)$ and $k^-(i)=k^-(i_1)$, a merging of two curves occurs. The pair is stored in a list for merging points. We exclude the case  $k^+(i)=k^-(i_1)$ and $k^-(i)=k^+(i_1)$, as this situation can never occur when taking care of the orientation of the initial curves. In case of $k^+(i)\neq k^+(i_1)$ or $k^-(i)\neq k^-i_1)$, triple junctions occur and the pair of nodes is stored in a list for triple junctions. 
\end{enumerate}

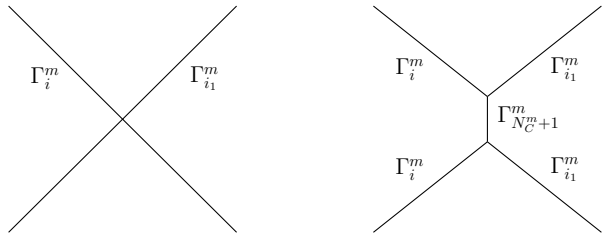
\begin{figure}[b]
\begin{center}
\begin{tikzpicture}[scale=0.6,transform shape]
\draw (0,0) -- (2.5, 2.5) -- (0,5); 
\draw (5,0) -- (2.5, 2.5) -- (5,5); 
\draw (0.8,3.4) node{\Large $\Gamma_i^m$};
\draw (4.3,3.4) node{\Large $\Gamma_{i_1}^m$};

\draw (8,0) -- (10.5, 2); 
\draw (10.5,3)-- (8,5); 
\draw (10.5,2) -- (10.5,3);
\draw (13,0) -- (10.5, 2); 
\draw (10.5,3) -- (13,5); 
\draw (8.8,3.6) node{\Large $\Gamma_i^m$};
\draw (12.2,3.6) node{\Large $\Gamma_{i_1}^m$};
\draw (8.8,1.4) node{\Large $\Gamma_i^m$};
\draw (12.2,1.4) node{\Large $\Gamma_{i_1}^m$};
\draw (11.4,2.5) node {\Large $\Gamma_{N_C^m+1}^m$};
\end{tikzpicture}
\end{center}
\caption{Conversion of a quadruple junction to a network with two triple junctions}
\label{fig:quadruple}
\end{figure}

Topological changes can thus be detected efficiently performing only
two loops over all mesh points. Consequently, the computational effort
for detecting topological changes remains small, more precisely it is
$\mathcal{O}(N)$. 
The algorithm can handle several topological changes at once. The
lists of split, merge, triple and boundary points can be sequentially
considered. In case of splitting and merging of curves, the neighbor
relation is modified by connecting node $j^*$ with $j_1^*+1$ and
$j_1^*$ with $j^*+1$.

Figure \ref{fig:topchange} presents an example where a curve splits up
in two single curves: The node $(1,23)$ lies in a square which has
been previously marked with $(1,7)$. Among the mesh points in a
neighborhood of $(1,7)$ and in a neighborhood of $(1,23)$, the pair
$(1,6)$ and $(1,24)$ is identified as the pair of nodes with smallest
distance. The neighbor relationships at $(1,6)$ and $(1,24)$ need to
be modified as shown in the right sub-figure.  To prevent another 
splitting and merging near the same point in $\Omega$, a
square can be blocked for new topological changes for some time-steps
after a topological change has taken place.

If a boundary point occurs, the point is first projected orthogonal to the boundary $\partial \Omega$ and is then duplicated. If the previous curve is closed, the curve becomes an open curve. If the curve has already been an open curve, it is separated in two single curves. The two boundary nodes, the detected node and its duplication, are the end point of one curve and the first point of the second curve. They are allowed to move away from each other along $\partial \Omega$ in the next time steps. Similarly, the complementary situation can be handled. An open curve can become a closed curve, if the two boundary points are too close to each other, i.e. lie in the same square of the background grid. Two curves can merge if their boundary points are too close to each other. Again, the square where a boundary intersection has taken place can be shortly blocked. 

In case of triple junctions, the topological change is handled as follows: First, as an intermediate step, a quadruple junction emerges at the point where $\Gamma_i^m$ and $\Gamma_{i_1}^m$ touch. The quadruple junction is replaced by two triple junctions connected with a small curve, see Figure \ref{fig:quadruple}. The small curve consists of only two initial points, namely $0.5(\vec X_{i,j^*}^m + \vec X_{i_1,j_1^*}^m)$ and $0.5(\vec X_{i,j^*\pm 1}^m + \vec X_{i_1,j_1^*\pm 1}^m)$, where the sign $\pm$ is chosen according to the orientation of the curves. Within the next time steps, the new curve can grow and new nodes are inserted by some global refinements.

\begin{figure}[t]
\begin{center}

\begin{tikzpicture}[scale=0.4,transform shape]
\begin{scope}
\draw (0,0) -- (-3, 3);
\draw (0,0) -- (-3,-3);
\draw (0,0) -- ( 3,-3);
\draw (0,0) -- ( 3, 3);

\draw (0,  2) node{\Huge $k_1$};
\draw (-2, 0) node{\Huge $k_2$};
\draw ( 0,-2) node{\Huge $k_3$};
\draw ( 2, 0) node{\Huge $k_4$};

\draw (0,-5) node[text width=6cm,text centered]{\Huge Quadruple Junction};
\end{scope}  

\begin{scope}[xshift=8cm]
\draw (-3, 3) .. controls (-0.25, 0.25) and (-0.25,-0.25) .. (-3,-3);
\draw ( 3, 3) .. controls ( 0.25, 0.25) and ( 0.25,-0.25) .. ( 3,-3);

\draw (0,  2) node{\Huge $k_1$};
\draw (-2, 0) node{\Huge $k_2$};
\draw ( 0,-2) node{\Huge $k_3$};
\draw ( 2, 0) node{\Huge $k_4$};
\draw ( 0, 0) node{\Huge $=$};

\draw (0,-5) node[text width=6cm,text centered]{\Huge Transform in two curves};
\end{scope}
  
\begin{scope}[xshift=16cm]
\draw (-3, 3) .. controls (-0.25, 0.25) and ( 0.25, 0.25) .. ( 3, 3);
\draw (-3,-3) .. controls (-0.25,-0.25) and ( 0.25,-0.25) .. ( 3,-3);

\draw (0,  2) node{\Huge $k_1$};
\draw (-2, 0) node{\Huge $k_2$};
\draw ( 0,-2) node{\Huge $k_3$};
\draw ( 2, 0) node{\Huge $k_4$};
\draw ( 0, 0) node{\Huge $=$};

\draw (0,-5) node[text width=6cm,text centered]{\Huge Transform in two curves};
\end{scope}

\begin{scope}[xshift=24cm]
\draw (-3, 3) -- (-0.5,0) -- (-3,-3);
\draw ( 3, 3) -- ( 0.5,0) -- ( 3,-3);
\draw ( 0.5,0) -- (-0.5,0);

\draw (0,  2) node{\Huge $k_1$};
\draw (-2, 0) node{\Huge $k_2$};
\draw ( 0,-2) node{\Huge $k_3$};
\draw ( 2, 0) node{\Huge $k_4$};

\draw (0,-5) node[text width=6cm,text centered]{\Huge Transform in a network with two triple junctions};
\end{scope} 

\begin{scope}[xshift=32cm]
\draw (-3, 3) -- (0, 0.5) -- ( 3, 3);
\draw (-3,-3) -- (0,-0.5) -- ( 3,-3);
\draw (0, 0.5) -- (0,-0.5);

\draw (0,  2) node{\Huge $k_1$};
\draw (-2, 0) node{\Huge $k_2$};
\draw ( 0,-2) node{\Huge $k_3$};
\draw ( 2, 0) node{\Huge $k_4$};

\draw (0,-5) node[text width=6cm,text centered]{\Huge Transform in a network with two triple junctions};
\end{scope}

\end{tikzpicture}
\end{center}

\caption{Quadruple junction and several possibilities for continuation}
\label{fig:quadruple2}
\end{figure}
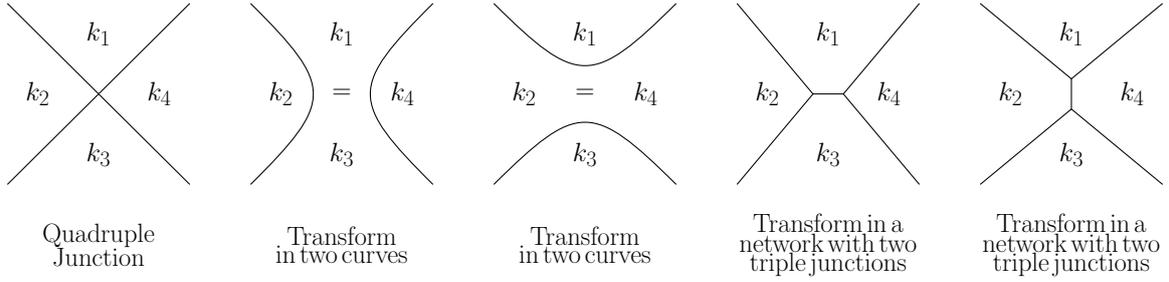

Another group of topological changes occur when a curve's length falls below a minimum length. If the curve is closed or if its two boundary nodes lie both on $\partial \Omega$, it is directly deleted. If both end nodes belong to triple junctions, the curve is deleted and an intermediate quadruple junction occurs. Dependent on the phase indices of the adjacent regions $\Omega_k$, $k\in\{1,\ldots,N_R\}$, and of the previous curve network, there are different ways how to continue, see Figure \ref{fig:quadruple2}. 
If a curve is marked for deletion with one boundary belonging to a triple junction and one boundary located at $\partial \Omega$, the curve and the triple junction are deleted. Further, the boundary points of the remaining two curves which have met at the triple junction are projected to the boundary $\partial \Omega$. In the next time steps, they are allowed to move away from each other.

\subsection{Computations of regions and coefficients}
\label{subsec:comp_regions_coeffs}
As the curves evolve in time, an approximation $\Omega_k^m$ of the regions $\Omega_k(t_m)$ and an approximation $c_k^m$ of the coefficients $c_k(t_m)$ (cf. \eqref{eq:c_k}), $k=1, \ldots, N_R$, has to be computed. As the image function $u_0:\Omega \rightarrow \mathbb{R}$ is locally constant at the pixels, we assign each pixel to a phase $\Omega_k^m$. If a pixel is truncated by a curve, it is assigned to the phase to which the largest part belongs. Let $S_k^m$ be the set of $n_k^m$ pixels belonging to $\Omega_k^m$. Then the approximation $c_k^m$ is set to 
\begin{equation}
c_k^m := \frac{C_k^m}{n_k^m}, \quad\quad C_k^m := \sum_{pix \in S_k^m} u_0|_{pix}.
\end{equation}
The entire image domain needs to be considered only for $m=0$. As the curves move only slightly from one time step to the next, the regions $\Omega_k^m$ and the coefficients $c_k^m$ need to be updated only in a small environment of the curves. This results in a very efficient computation of the regions and the coefficients. As the normal $\vec\nu_i^m$ points from $\Omega_{k^-(i)}^m$ to $\Omega_{k^+(i)}^m$, the pixels close to the curve $\Gamma_i^m$ are assigned to the phase $k^+(i)$ or $k^-(i)$, respectively. 

As initialization we set $n_k^m = n_k^{m-1}$ and $C_k^m = C_k^{m-1}$ for $k=1,\ldots,N_R$. For $i=1,\ldots,N_C$, all pixels in a environment of $\Gamma_i^m$ are subsequently considered. Let a pixel $pix$ be assigned to phase $k \in \{k^+(i), k^-(i)\}$ and let $l\neq k$ be the former phase index of the pixel. We set
\begin{equation}
n_k^m = n_k^m + 1, \quad n_l^m = n_l^m -1, \quad C_k^m = C_k^m + u_0|_{pix}, \quad C_l^m = C_l^m - u_0|_{pix}.
\end{equation}
After having considered all pixels close to the curves, the coefficients are set to $c_k^m = C_k^m / n_k^m$ for $k=1, \ldots, N_R$. 

Similarly, the coefficients in case of colored images can be computed. The coefficients are all means or normalized means of the corresponding component of the image function. In case of CB or HSV color space, the value $u_0|_{pix}$ needs to be first transformed to the CB and HSV color. If image components lie on $S^1$ or $S^2$, the computation needs to be slightly modified: For example, in case of the HSV space, the hue coefficients are computed by $c_k^m = C_k^m / \|C_k^m\|$  with $C_k^m = \sum_{pix \in S_k^m} h_0|_{pix}$, where $h_0 : \Omega \rightarrow S^2$ is the hue component of the image.

\subsection{Solution of the Image Denoising Problem}
\label{subsec:image_smoothing_numerics}
Besides the segmentation of the image, the algorithm provides a piecewise constant approximation $u=\sum_{k=1}^{N_R} c_k^m \chi_{\Omega_k^m}$ of $u_0$. A piecewise smooth approximation can be obtained by solving a discretization of the boundary value problem \eqref{eq:image_diffusion_scheme}. For the bulk equations, a finite difference approximation can be used. 

Therefore, we consider a spatial discretization $\Omega^h := \left\{(ih,jh)\,:\, i=0, \ldots, N_x, \,j=0\ldots,N_y\right\}$, $N_x, N_y \in \mathbb{N}$, of the rectangular image domain $\Omega$. 
As we have to solve an equation on each phase separately, we define the discrete set $\Omega_k^h := \Omega^h \cap \overline{\Omega_k^m}$. In the following, we consider $k$ as fixed. 
We aim at finding a function $u^h$ minimizing a discrete analogue of the energy \eqref{eq:image_diffusion_functional}. 

We define for $i=1,\ldots, N_x$, $j=1,\ldots, N_y$
\begin{align*}
A_x(i,j) &= area\left(\left([(i-1)h,ih]\times[(j-\frac12)h,(j+\frac12)h]\right)\cap \Omega_k^m\right), \\
A_y(i,j) &= area\left(\left([(i-\frac12)h,(i+\frac12)h]\times[(j-1)h,jh]\right)\cap \Omega_k^m\right), \\
\end{align*}
and for $i=0,1,\ldots, N_x$, $j=0,1,\ldots, N_y$
\begin{equation*}
A(i,j) = area\left(([(i-\frac12)h,(i+\frac12)h]\times[(j-\frac12)h,(j+\frac12)h])\cap \Omega_k^m\right),
\end{equation*}
where $area$ denotes the two-dimensional Lebesgue measure. 

We define the following discrete energy
\begin{align}
E_{\mathrm{discr},k}(u^h) 
& = \sum_{i=1}^{N_x} \sum_{j=1}^{N_y} \left( A_x(i,j) \left(\frac{u_{i,j}^h - u_{i-1,j}^h}{h}\right)^2 + A_y(i,j) \left(\frac{u_{i,j}^h - u_{i,j-1}^h}{h}\right)^2 \right)\nonumber\\
& + \sum_{i=0}^{N_x} \sum_{j=0}^{N_y} \lambda_k A(i,j) \left( u_{i,j}^h - u_0(ih,jh)\right)^2 
\label{eq:2d_discrete_energy}
\end{align}
which is a discrete analogue of $\int_{\Omega_k} \left((u_x^2 + u_y^2) + \lambda_k (u-u_0)^2 \right) \,\mathrm{d}x$. 
\begin{figure}
\begin{center}
\begin{tikzpicture}[scale=0.8,transform shape]
\draw(0,0)--(5,0);
\draw(0,1)--(5,1);
\draw(0,2)--(5,2);
\draw(0,3)--(2,3);
\draw(0,4)--(2,4);
\draw(0,5)--(2,5);

\draw(0,0)--(0,5);
\draw(1,0)--(1,5);
\draw(2,0)--(2,5);
\draw(3,0)--(3,2);
\draw(4,0)--(4,2);
\draw(5,0)--(5,2);

\draw (3.0,3.0) node{\large $(i_3,j_3)$};
\draw[->] (2.7,2.7)--(2.1,2.1);

\draw (-1.2,2.0) node{\large $(i_1,j_1)$};
\draw[->] (-0.6,2.0)--(-0.1,2);

\draw (-1.2,0.0) node{\large $(i_2,j_2)$};
\draw[->] (-0.6,0.0)--(-0.1,0);

\end{tikzpicture}
\end{center}
\caption{A possible region $\Omega_k^m$}
\label{fig:region}
\end{figure}
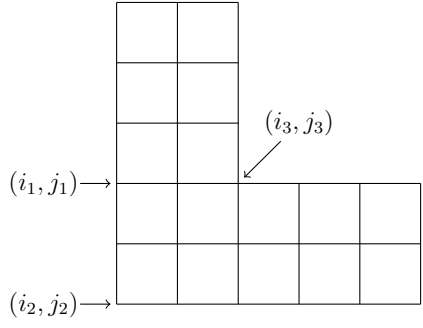
For each $(ih,jh)\in \Omega_k^h$ we take the derivative with respect to
$u_{i,j}^h$ and set the resulting term to zero. This leads to a linear
system. Recall that $\Omega_k^m$ need not be rectangular. Geometries
like the one shown in Figure \ref{fig:region} can occur. 

At interior nodes $(ih,jh)$, we get the following equation considering the derivative with respect to $u_{i,j}^h$: 
\begin{equation*}
\left(4+\lambda_k h^2\right) u_{i,j}^h - u_{i-1,j}^h - u_{i+1,j}^h - u_{i,j-1}^h - u_{i,j+1}^h = \lambda_k h^2 u_0(ih,jh).
\end{equation*}

At boundary points like $(i_1,j_1), (i_2,j_2), (i_3,j_3)$, we get the
following equations from \eqref{eq:2d_discrete_energy}:
\begin{align}
\left(2+\frac12 \lambda_k h^2\right) u_{i_1,j_1}^h - u_{i_1+1,j_1}^h - \frac12 u_{i_1,j_1-1}^h - \frac12 u_{i_1,j_1+1}^h  &= \frac12 \lambda_k h^2 u_0(i_1 h,j_1 h),
\label{eq:equation_i1j1} \\
\left(1+\frac14 \lambda_k h^2\right) u_{i_2,j_2}^h - \frac12 u_{i_2+1,j_2}^h - \frac12 u_{i_2,j_2+1}^h                   &= \frac14 \lambda_k h^2 u_0(i_2 h,j_2 h), \nonumber \\
\left(3+\frac34 \lambda_k h^2\right) u_{i_3,j_3}^h - u_{i_3-1,j_3}^h - \frac12 u_{i_3+1,j_3}^h - u_{i_3,j_3-1}^h - \frac12 u_{i_3,j_3+1}^h &= \frac34 \lambda_k h^2 u_0(i_3 h, j_3 h). \nonumber
\end{align}

At boundary points we can derive the Neumann boundary conditions from
the linear equations considering the limit $h \rightarrow 0$.  For example,
rewriting \eqref{eq:equation_i1j1} as follows
\begin{equation*}
-\frac{u_{i_1+1,j_1}^h - u_{i_1,j_1}^h}{h} + \frac{u_{i_1,j_1}^h - u_{i_1,j_1-1}^h}{2h} -  \frac{u_{i_1,j_1+1}^h - u_{i_1,j_1}^h}{2h} 
  = \frac12 \lambda_k h \left(u_0(i_1 h,j_1 h)-u_{i_1,j_1}^h\right).
\end{equation*}
and considering $h \rightarrow 0$ yields the Neumann boundary condition $u_x=0$. 
In summary, we have a linear system whose system matrix is sparse and strictly diagonally dominant as $\lambda_k > 0$. Thus, the linear system has a unique solution. 

The pixel grid ($h=1$) serves as natural grid for a spatial discretization.  The numerical solution $u^h$ is determined for all corner points of pixels. Consider a pixel in $\Omega_k^m$ with corner points $(i,j)$, $(i-1,j)$, $(i-1,j-1)$ and $(i,j-1)$. Let $p_{i,j}$ be the center of the pixel. Then a denoised image function is defined by 
\begin{equation}
u^h(p_{i,j}) ) = \frac 14 \left(u_{i,j}^h + u_{i-1,j}^h + u_{i-1,j-1}^h + u_{i,j-1}^h \right).
\end{equation}

Color images are denoised component-wise, i.e. each channel is treated like a scalar image. 

It is sufficient to perform the image smoothing for $m=M$, i.e. as a
post-processing step, when the region interfaces $\Gamma_i^M$ match
approximately with the edges of the objects in the image.

\subsection{Summary of the algorithm}
In summing up, we propose the following algorithm for image segmentation. Given a set of polygonal curves $\Gamma^0 = (\Gamma_1^0, \ldots, \Gamma_{N_C}^0)$ and $\vec X^0 = (\vec X_1^0, \ldots, \vec X_{N_C}^0)$ with $\vec X_i^0(I_i)=\Gamma_i^0$, perform the following steps for $m=0, 1, \ldots, M-1$: 
\begin{enumerate}
\item \label{step1} Compute the regions $\Omega_k^m$ and the coefficients $c_k^m$, $k=1,\ldots,N_R$, as described in Section \ref{subsec:comp_regions_coeffs}. 
\item \label{step2} Compute $b^m$ as defined in \eqref{eq:righthandside} by using the coefficients $c_k^m$ of step \ref{step1}. Compute $\vec X^{m+1} = \vec X^m + \delta \vec X^{m+1}$ by solving the linear equation \eqref{eq:schur}, see Section \ref{subsec:fd_appr}.
\item Check if topological changes occur, see Section \ref{subsec:topchange}. In case of a topological change, repeat the steps \ref{step1} and \ref{step2} $n_\mathrm{sub}$-times with a step size of $\tau_m / n_\mathrm{sub}$ and execute the topological change when it occurs in a sub-step. 
\item Compute the length of the polygonal curves $\Gamma_i^m$. If $|\Gamma_i^m|\,/N_i^m < l_{\mathrm{min}}$, perform a global coarsening, and if $|\Gamma_i^m|\,/N_i^m > l_{\mathrm{max}}$, perform a global refinement of the curve discretization. 
\end{enumerate}

We can use a constant parameter $\sigma$ which weights the curvature
term in \eqref{eq:scheme1}. Alternatively, the parameter can be set
adaptively: For that, the internal energy $\sigma |\Gamma|$ in the
energy functional \eqref{eq:Mumford_Shah_const} is set to a certain
percentage of the external energy $E(\Gamma,c_1,\ldots,c_{N_R}) -
\sigma |\Gamma|$. The percentage is shortly called
$\sigma$-factor. The energies and the new value for $\sigma$ need not
be updated at each time step. In the results presented in the next
section, the factor $\sigma$ is computed either every 10th or every
50th time step. 

Having found a final segmentation, a smooth approximation of the image is computed as described in Section \ref{subsec:image_smoothing_numerics}.

\section{Results}
\begin{figure}
\centering
\includegraphics[viewport = 150 280 450 560, width = 0.24\textwidth]{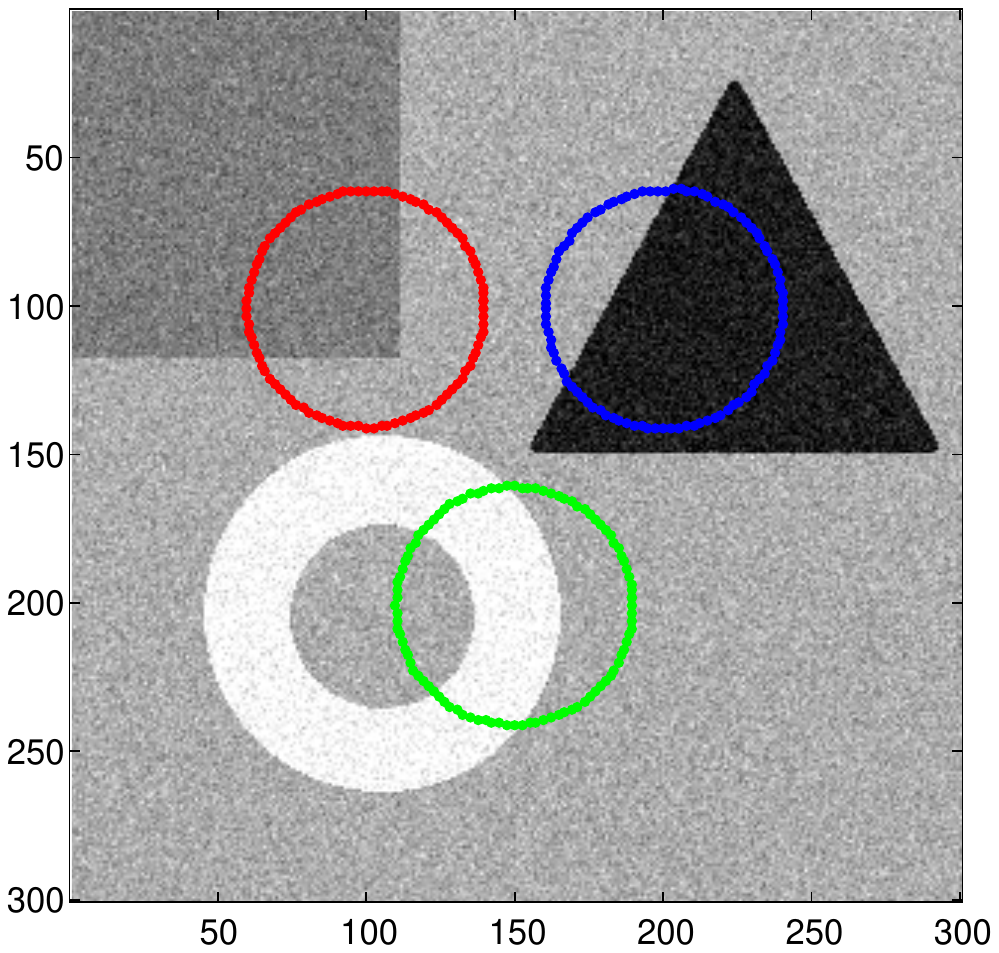}
\includegraphics[viewport = 150 280 450 560, width = 0.24\textwidth]{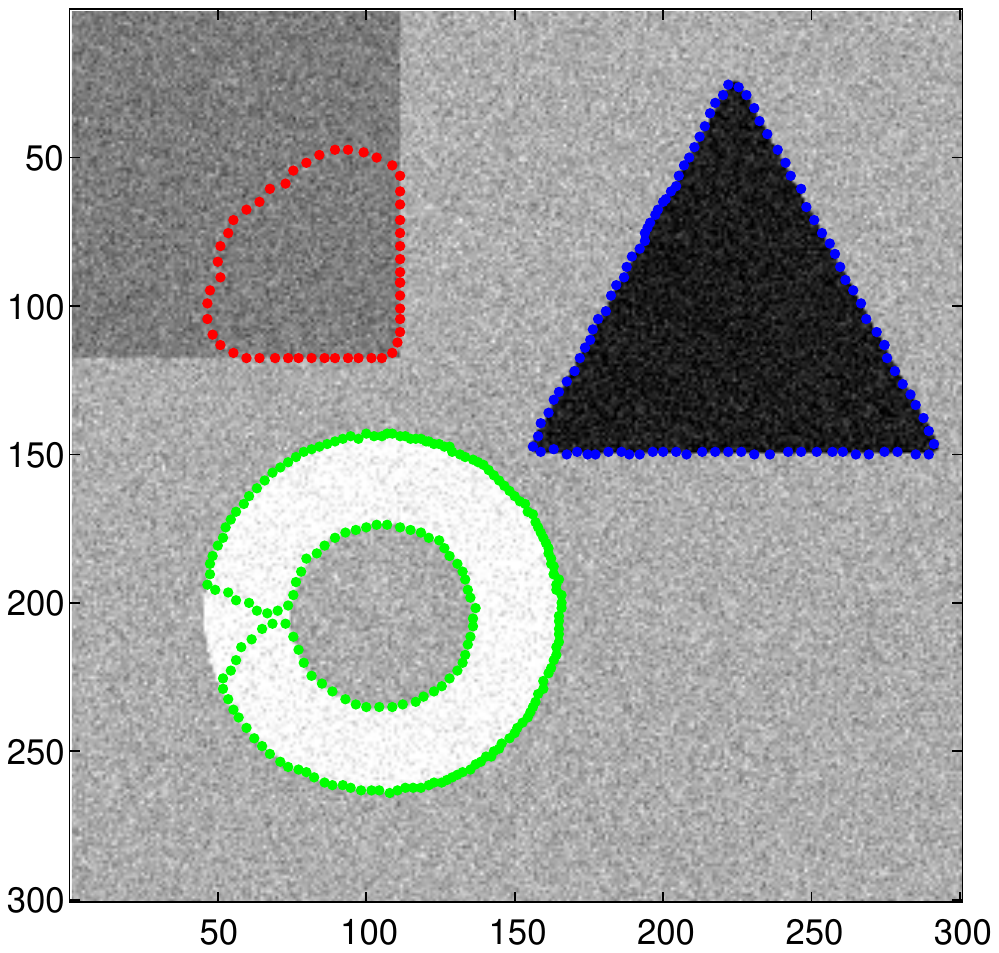}
\includegraphics[viewport = 150 280 450 560, width = 0.24\textwidth]{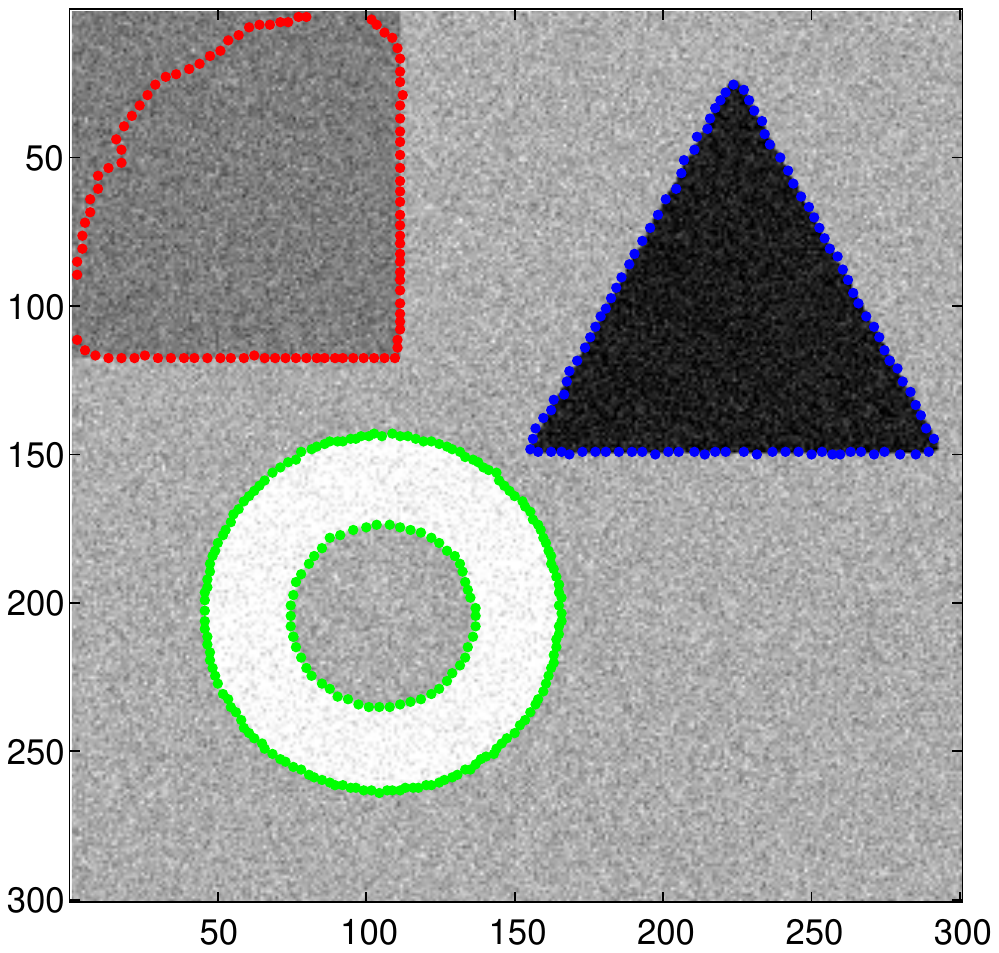}
\includegraphics[viewport = 150 280 450 560, width = 0.24\textwidth]{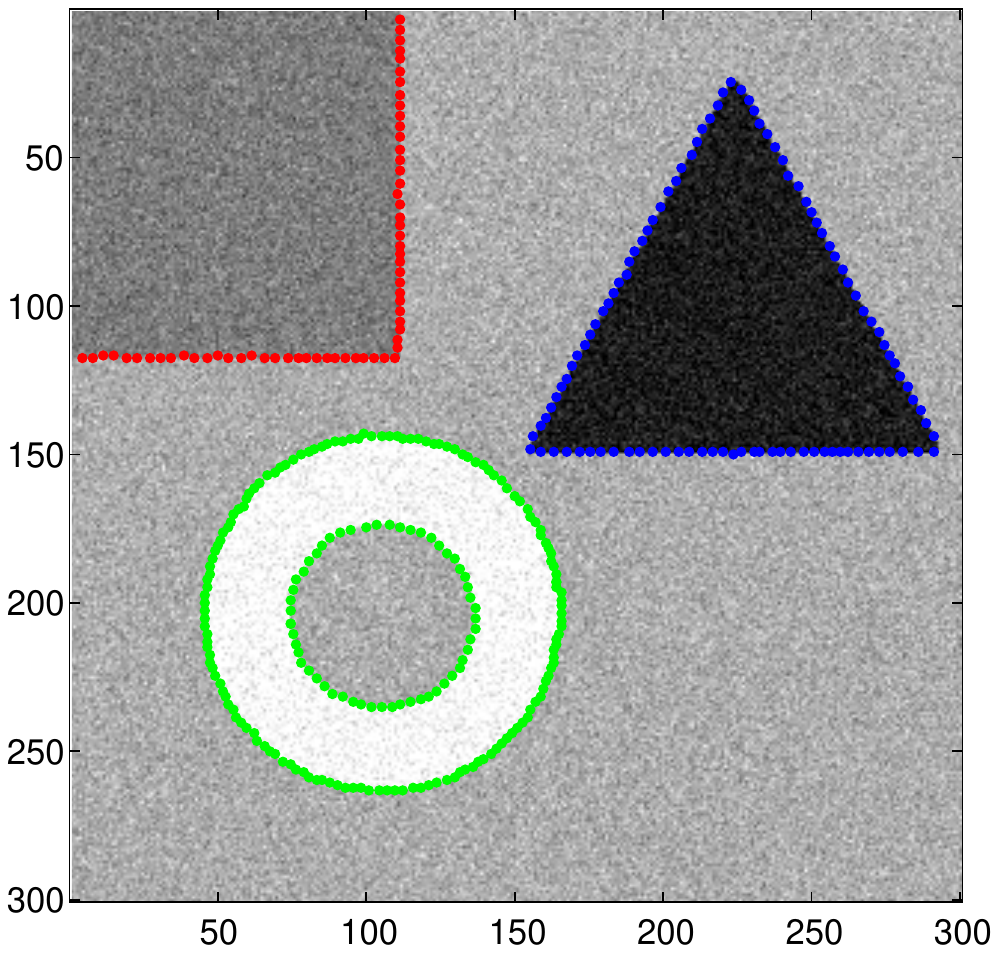}\\
\includegraphics[viewport = 150 280 450 560, width = 0.24\textwidth]{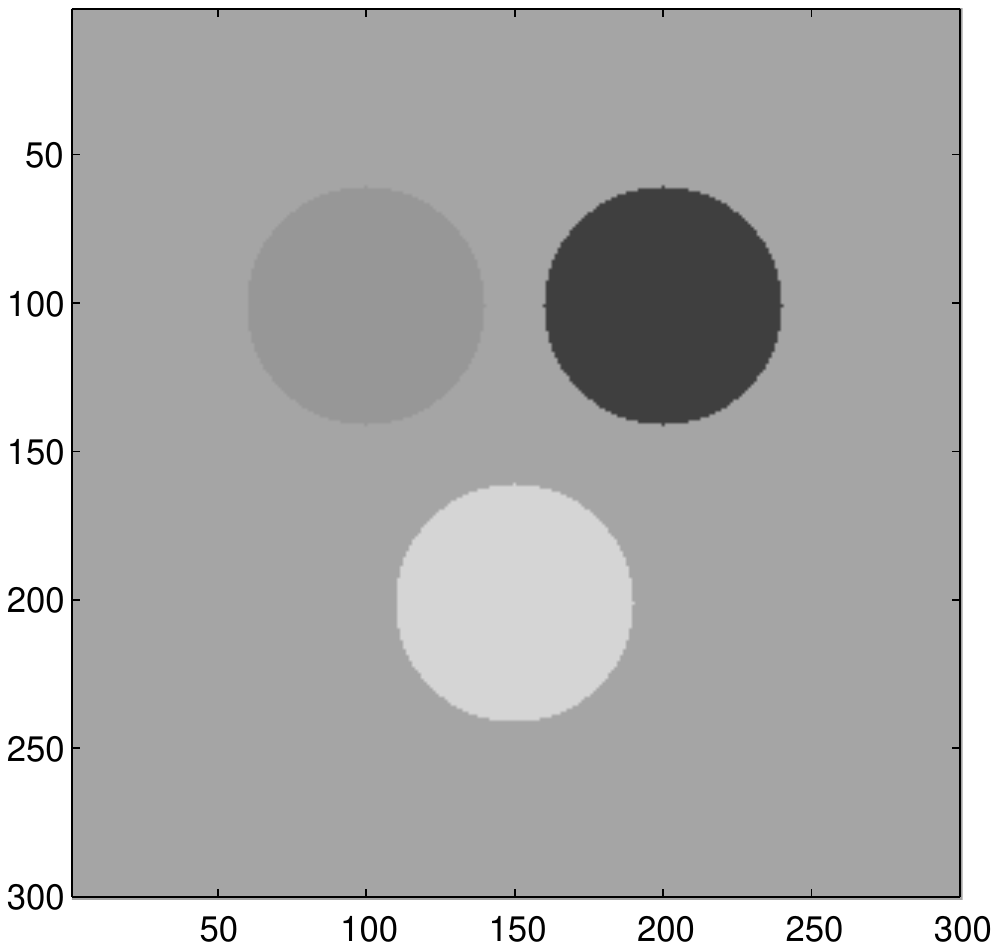}
\includegraphics[viewport = 150 280 450 560, width = 0.24\textwidth]{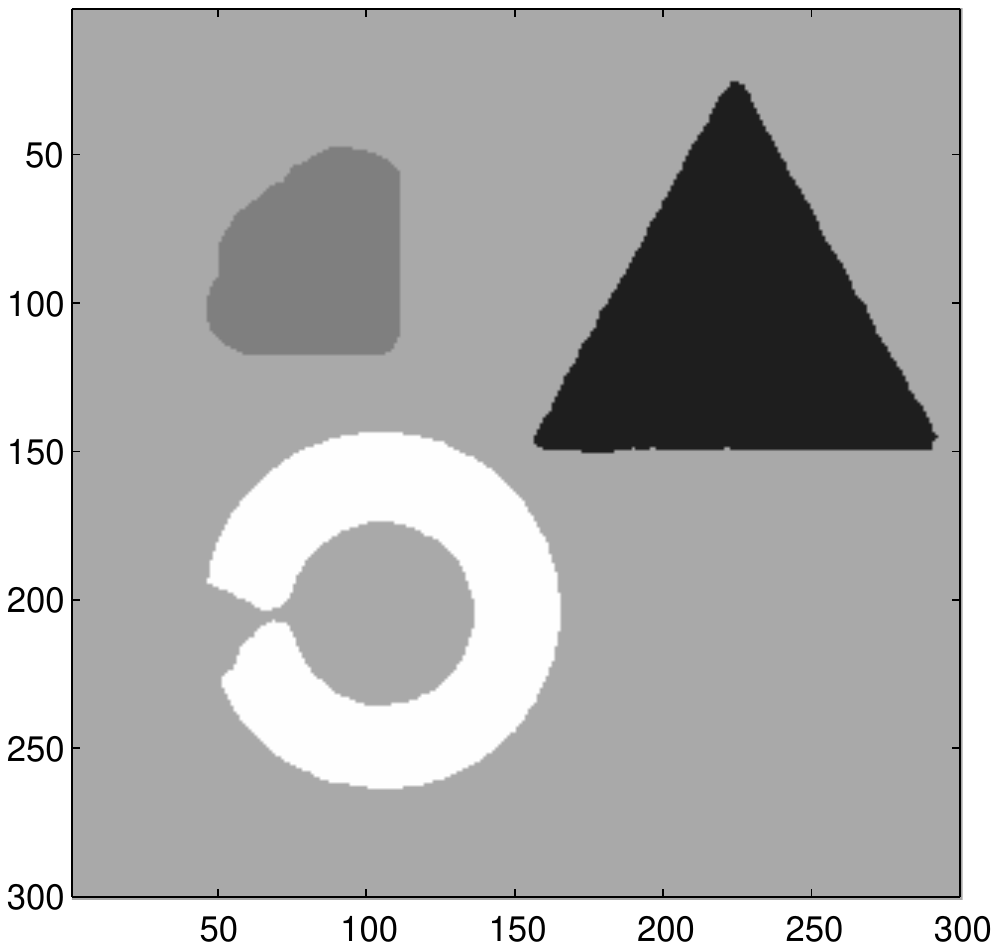}
\includegraphics[viewport = 150 280 450 560, width = 0.24\textwidth]{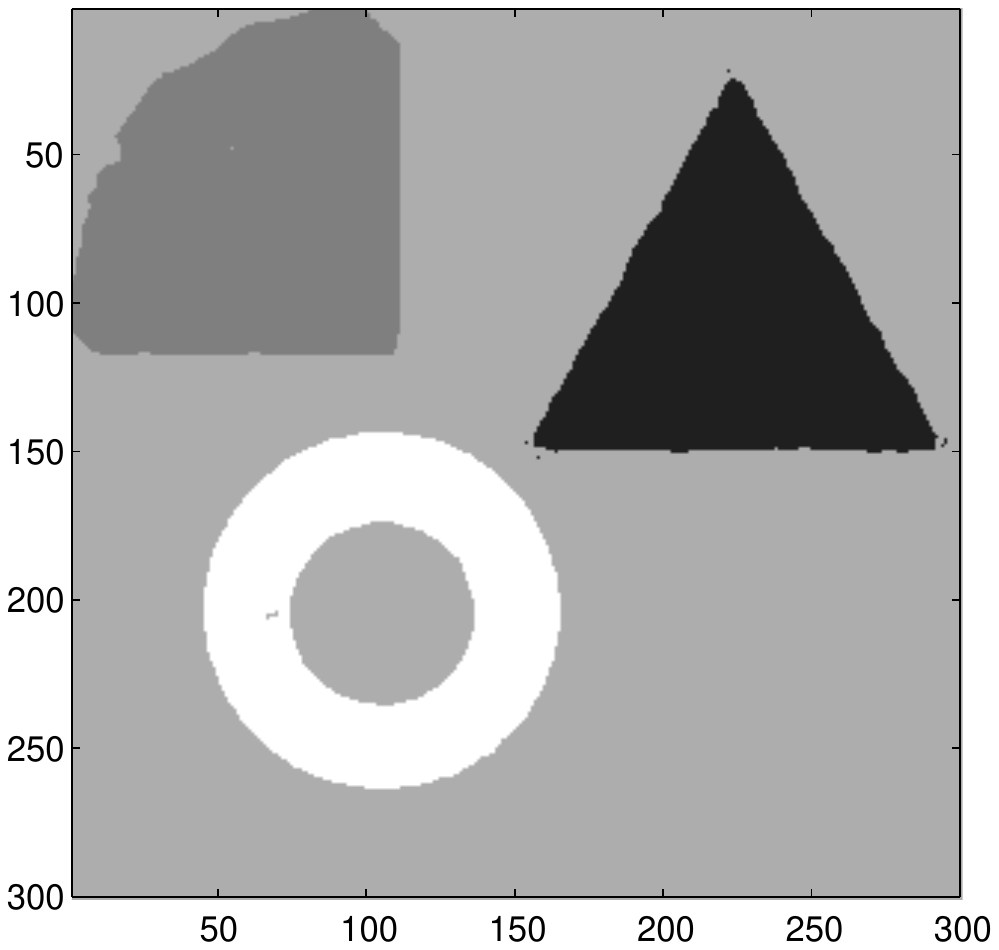}
\includegraphics[viewport = 150 280 450 560, width = 0.24\textwidth]{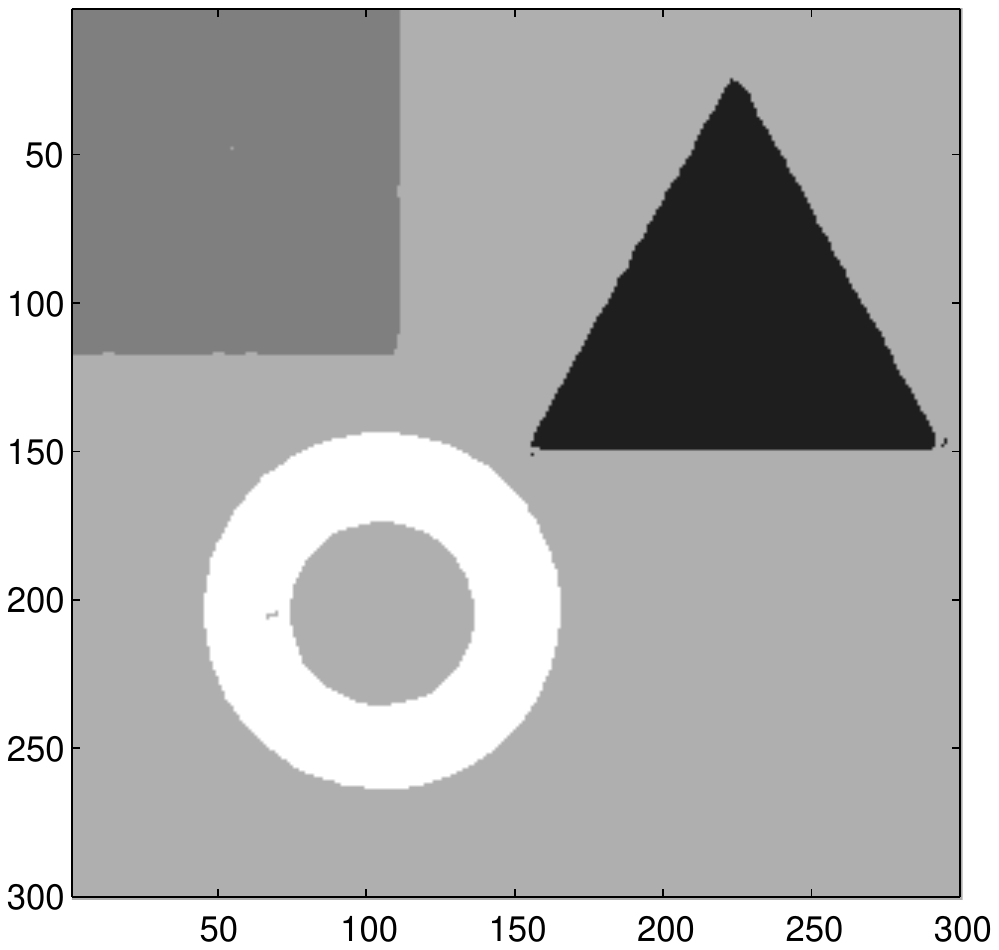}
\caption{Multi-phase image segmentation of a gray-scaled image with multiple objects to be detected, $m=1,560,2150,3000$, $\Delta t=0.1$, $\lambda = 20$, $\sigma$-factor $20\%$}
\label{fig:results_3objects}
\end{figure}

We apply the image segmentation method of Section \ref{sec:models} and
\ref{sec:numerics} on artificial and real, gray-scaled and color
images. The UMFPACK algorithm as MATLAB built-in routine is used to
solve the linear equation (\ref{eq:schur}), see also \cite{Davis}. 

In a first experiment, a gray-scaled image with three objects is
segmented, see Figure \ref{fig:results_3objects}. Typically, the
initial contours are closed curves which evolve with time according to
the evolution equation \eqref{eq:scheme1}. The first row in Figure
\ref{fig:results_3objects} shows the given image $u_0$ and the
contours for four different iteration steps. The second row shows the
piecewise constant, approximating image given by
\eqref{eq:piecewise_const_approx}. The gray value in the region
$\Omega_k$, $k=1, \ldots, N_R$ (here $N_R=4$), is the mean of $u_0$ in
$\Omega_k$, recall \eqref{eq:c_k}. Having detected the objects, the
contours match with the edges of the objects and the approximating
image is a smooth, piecewise constant version of $u_0$.

This sample image also demonstrates two kinds of topological changes: splitting of a curve and creation of boundary intersection points. The green curve splits up into two single sub-curves. At time step $m=560$, two parts of the curve nearly touch. The splitting is detected as described in Section \ref{subsec:topchange}. The red contour in Figure \ref{fig:results_3objects} intersects the image boundary $\partial \Omega$ at two positions. Two open curves with each two boundary intersection points exist at time step $m=2150$. The length of the smaller sub-curve decreases continuously in the following time steps and the curve is deleted close to the upper left corner of the image. The remaining curve is attracted to the boundary of the upper left gray square. 

\begin{figure}
\centering
\includegraphics[viewport = 150 280 450 560, width = 0.24\textwidth]{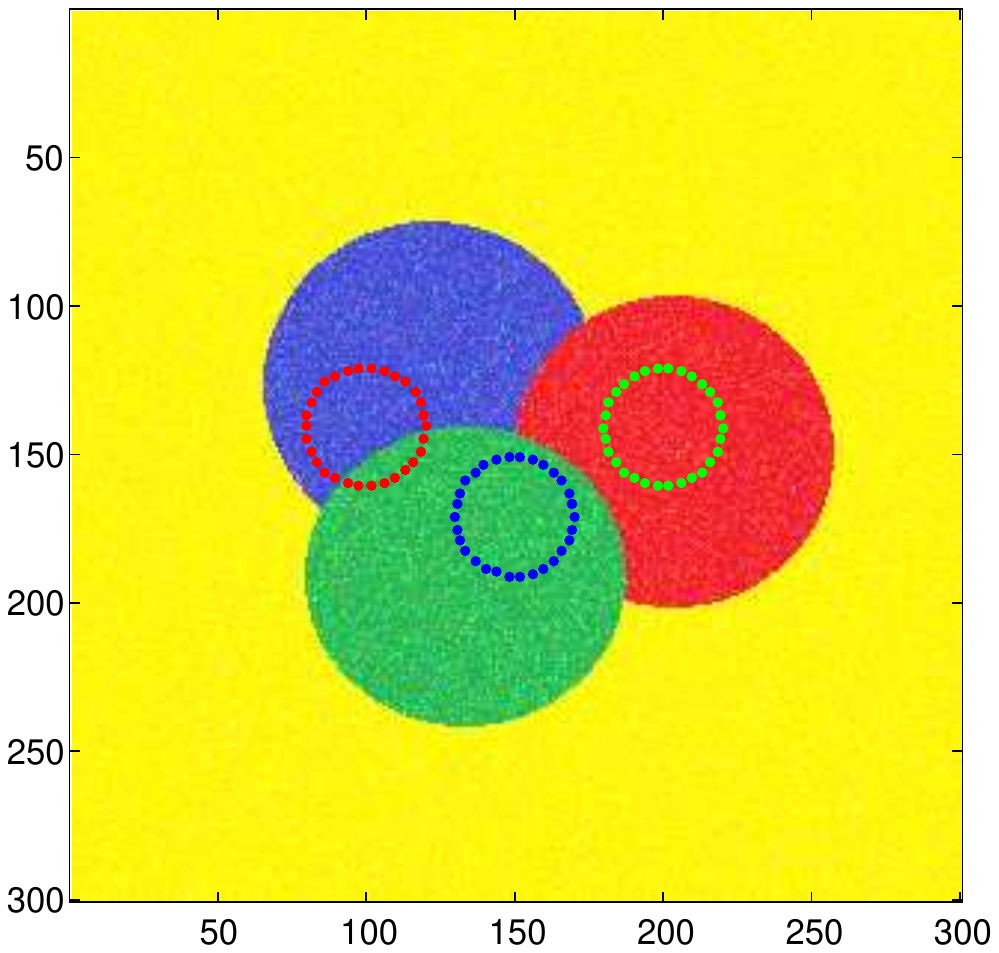}
\includegraphics[viewport = 150 280 450 560, width = 0.24\textwidth]{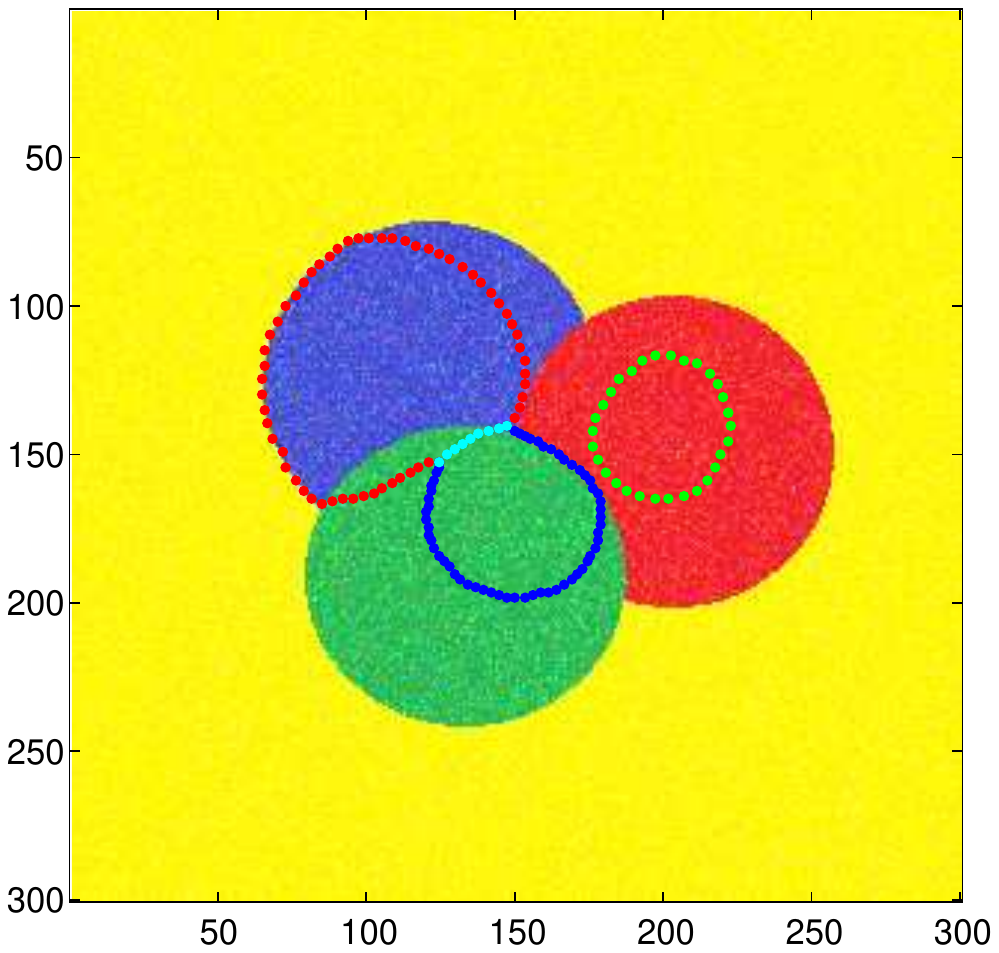}
\includegraphics[viewport = 150 280 450 560, width = 0.24\textwidth]{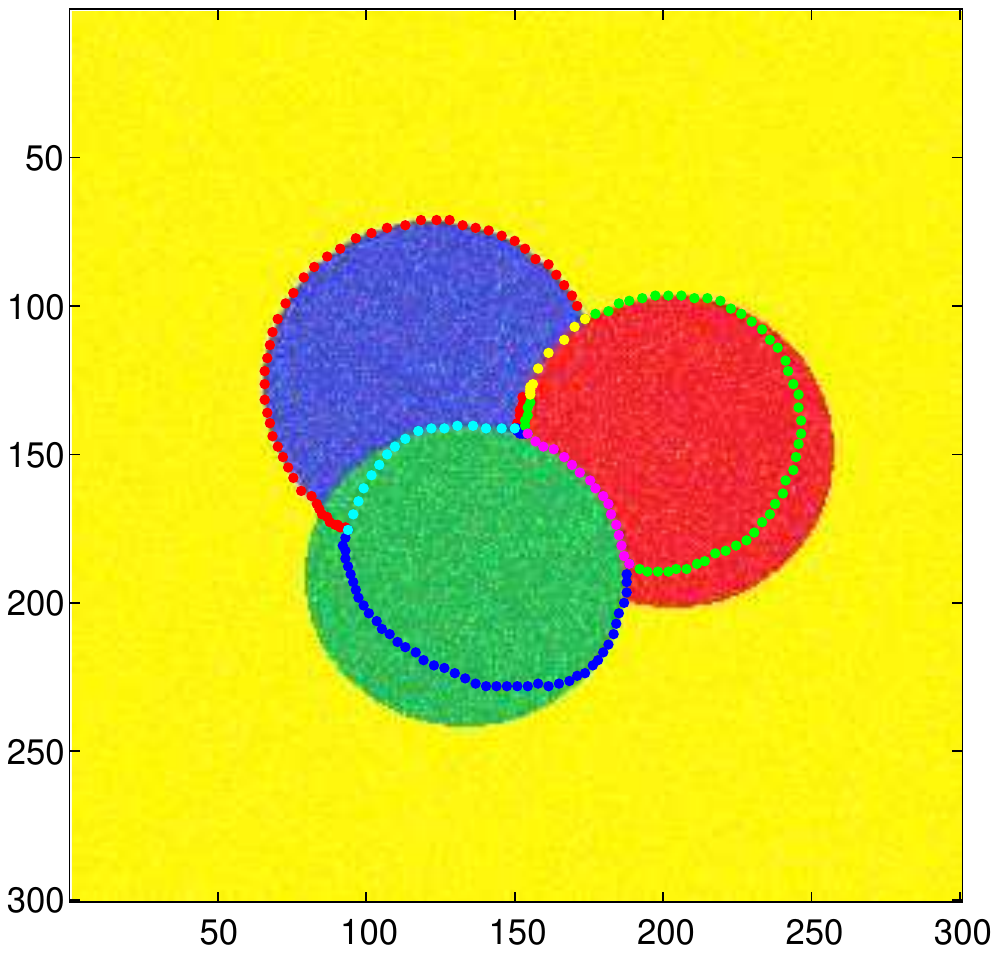}
\includegraphics[viewport = 150 280 450 560, width = 0.24\textwidth]{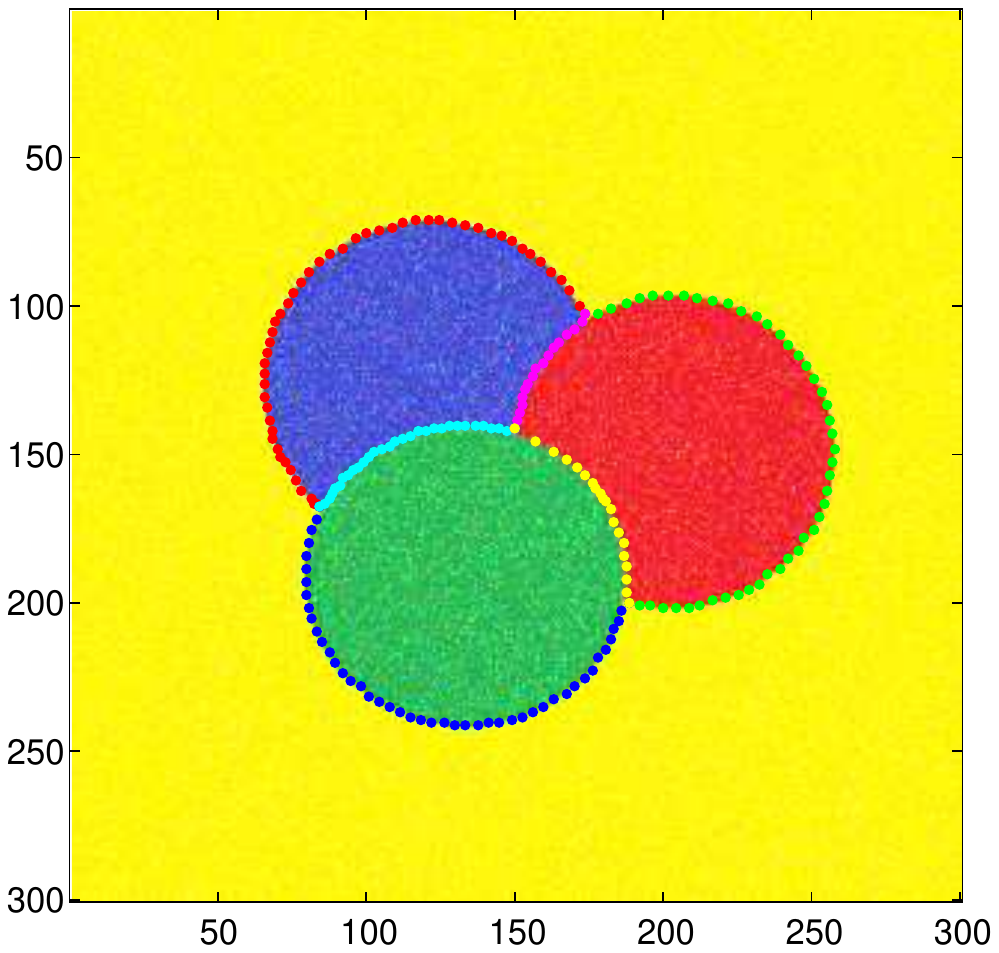}\\
\includegraphics[viewport = 150 280 450 560, width = 0.24\textwidth]{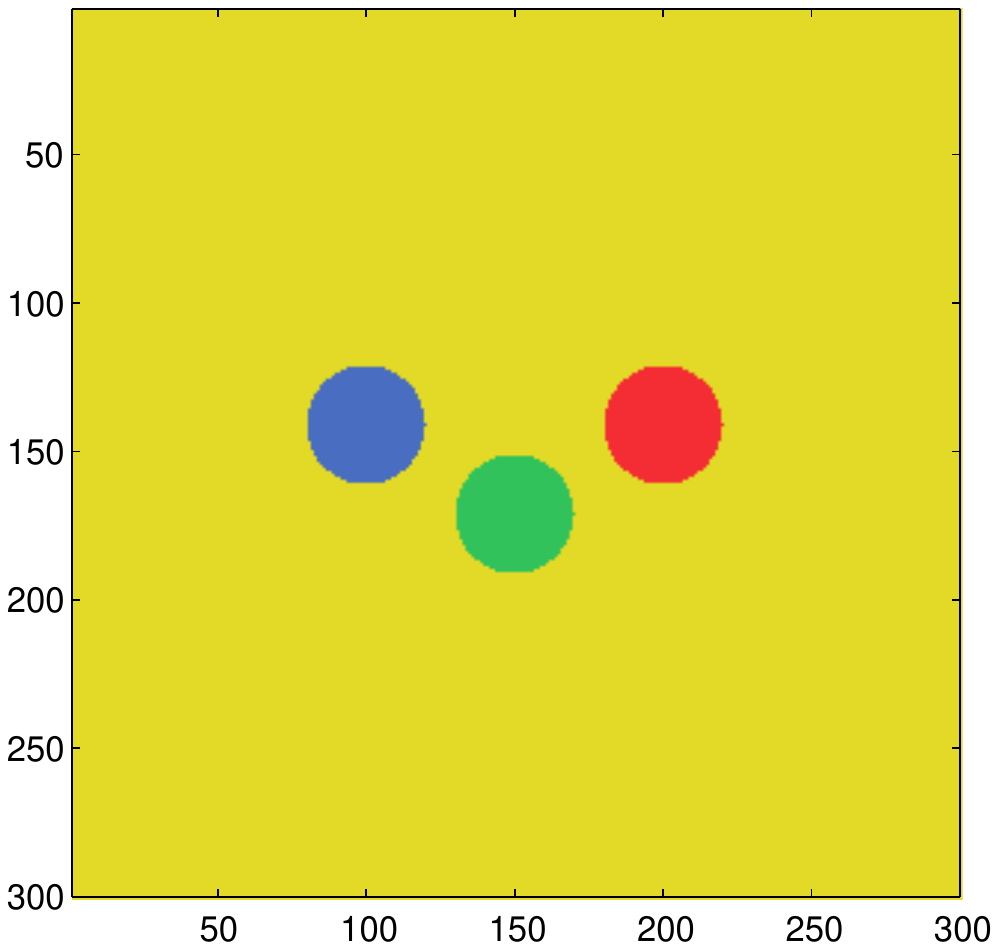}
\includegraphics[viewport = 150 280 450 560, width = 0.24\textwidth]{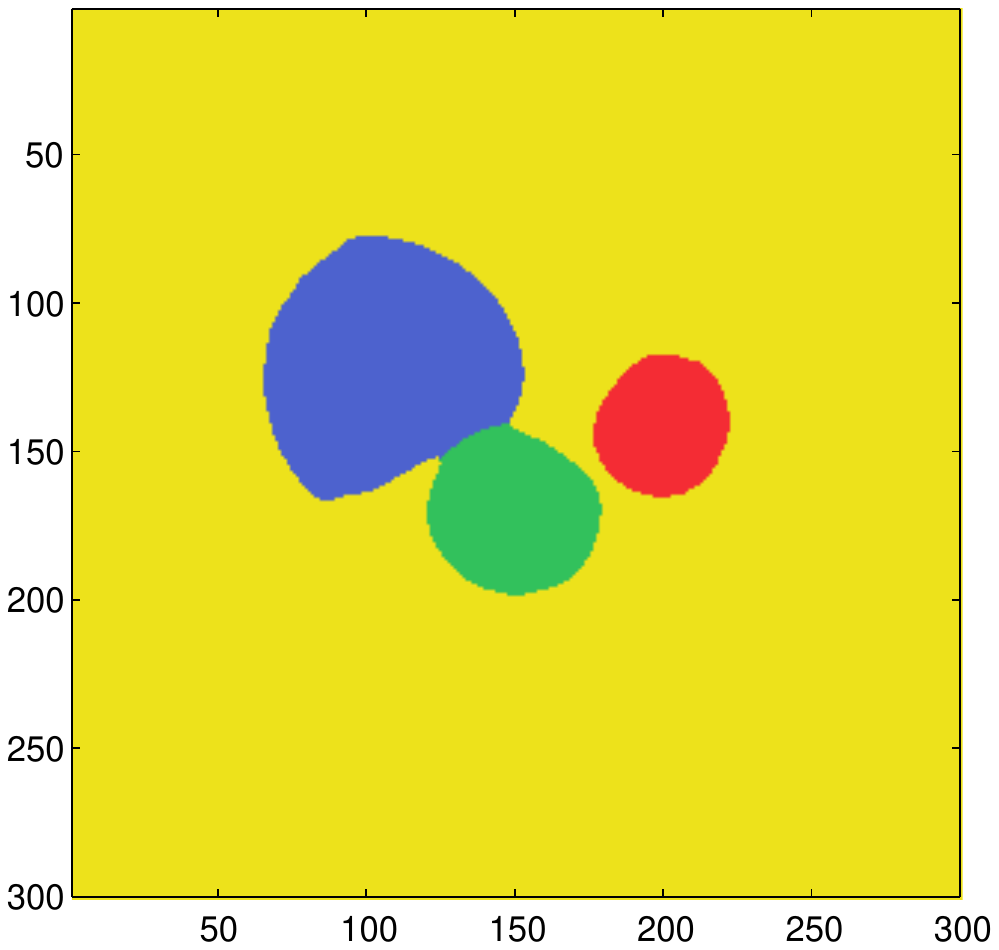}
\includegraphics[viewport = 150 280 450 560, width = 0.24\textwidth]{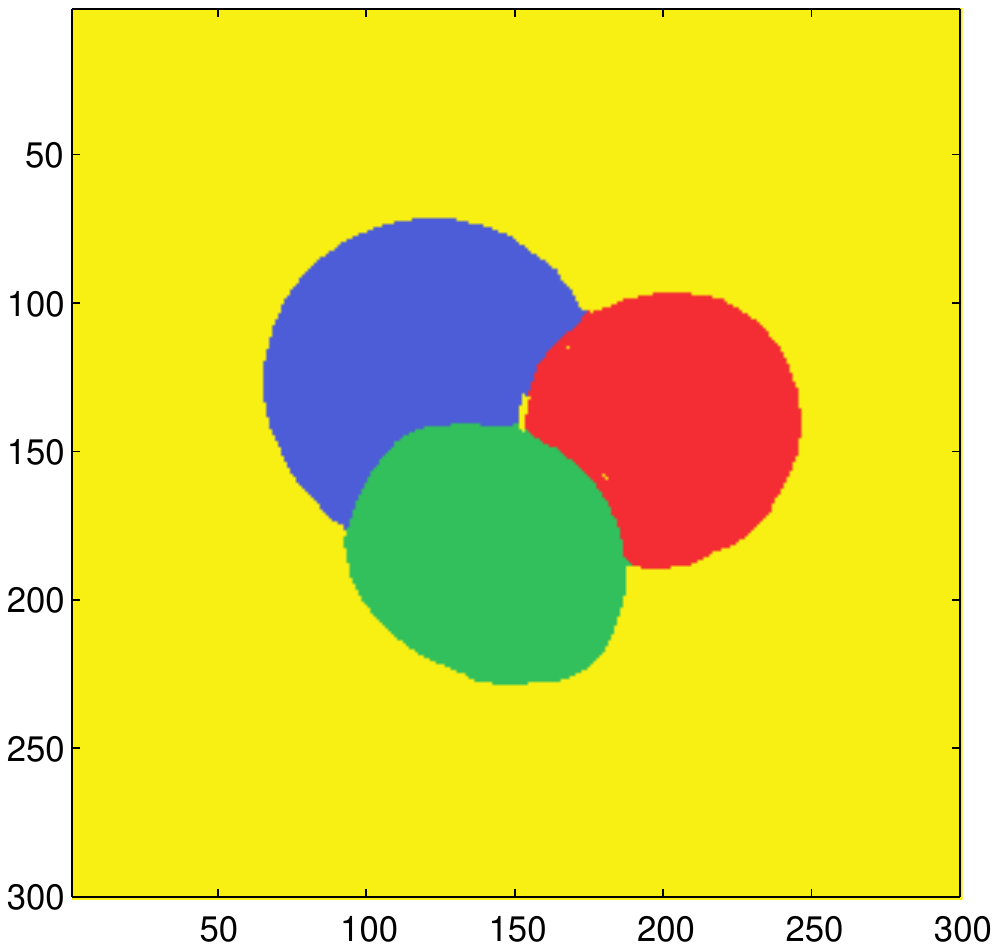}
\includegraphics[viewport = 150 280 450 560, width = 0.24\textwidth]{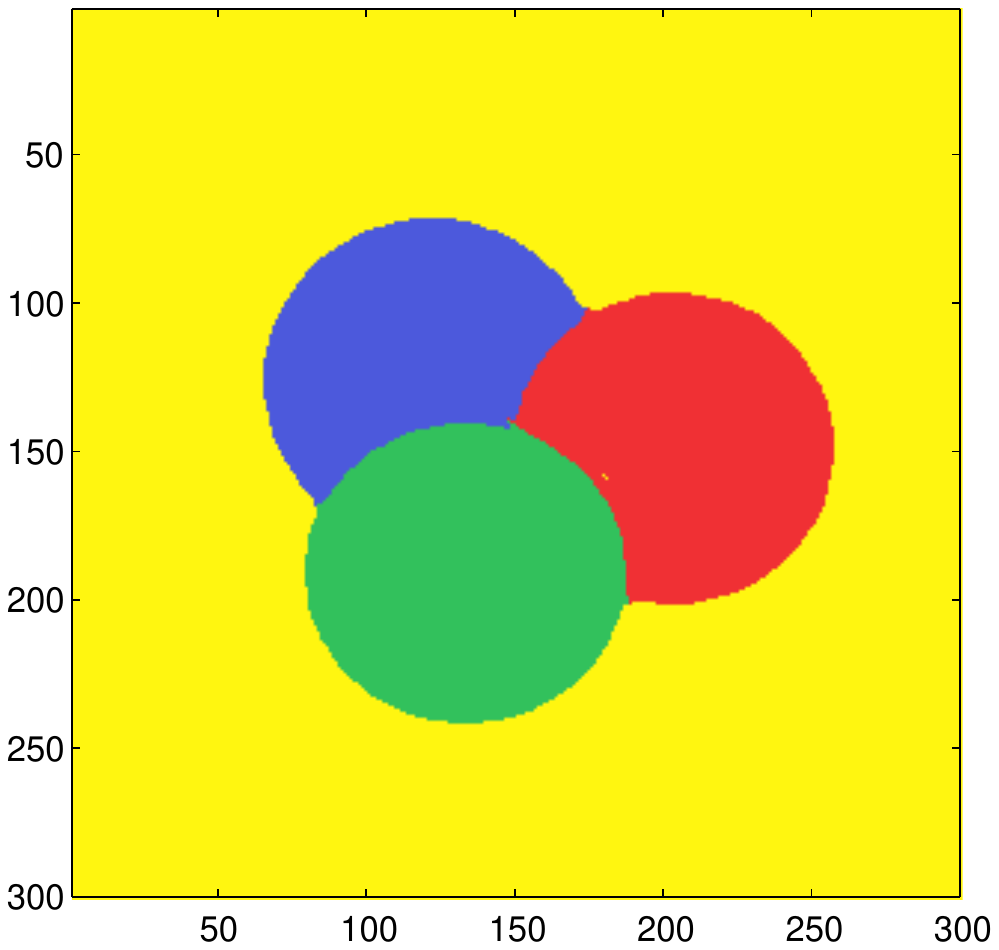}
\caption{Multi-phase image segmentation of a  color image using RGB space, $m=1,100,200,300$, $\Delta t=0.1$, $\lambda_1=\lambda_2=\lambda_3 = 5$, $\sigma$-factor $25\%$}
\label{fig:results_3balls}
\end{figure}

Figure \ref{fig:results_3balls} presents the segmentation results applying the algorithm on a color image. For this experiment, the RGB space is used and for all color channels $i\in \{1,2,3\}$ the parameter $\lambda_i=5$ is used. When two different curves touch, an intermediate quadruple junction occurs. A new small curve is created and the quadruple junction is replaced by two triple junctions connected by the new contour as illustrated in Section \ref{subsec:topchange}, Figure \ref{fig:quadruple}. 

\begin{figure}
\centering
\includegraphics[viewport = 105 265 475 560, width = 0.4\textwidth]{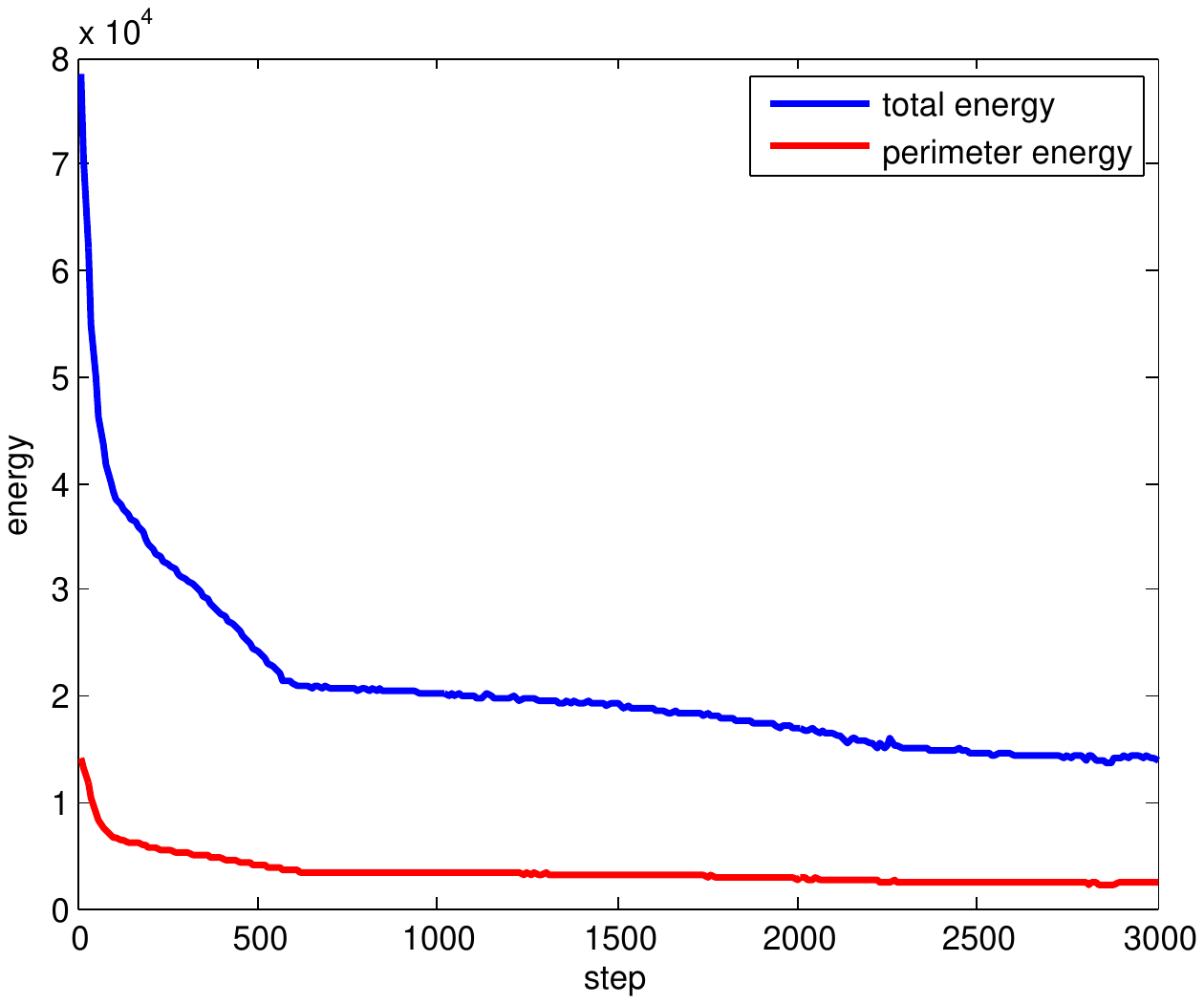}
\includegraphics[viewport = 105 265 475 560, width = 0.4\textwidth]{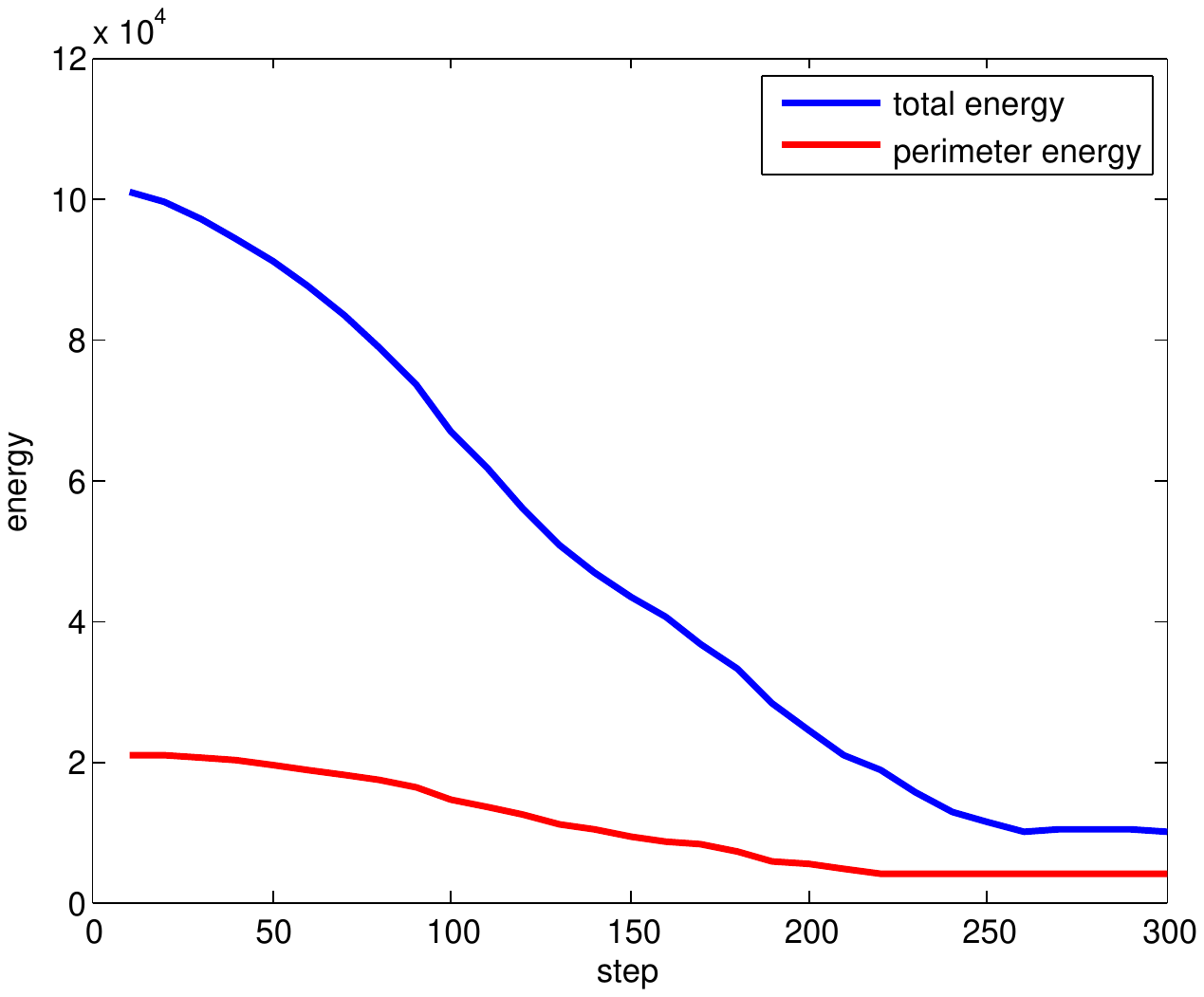}
\caption{Energy decrease for test images of Figure \ref{fig:results_3objects} (left) and Figure \ref{fig:results_3balls} (right)}
\label{fig:energies}
\end{figure}

The Mumford-Shah energy \eqref{eq:Mumford_Shah_const} for piecewise
constant approximations $u$ is shown in Figure
\ref{fig:energies}. Additional to the total energy, the perimeter
energy $\sigma |\Gamma|$ is plotted to visualize the proportion of the
length term which ensures smoothness of the curve.  The plots show the
energy decrease when segmenting the gray-scaled test image of Figure
\ref{fig:results_3objects} and the color image of Figure
\ref{fig:results_3balls}. The energy plotted on the left decreases
fast at the beginning. After approximately $600$ steps, the ring and
the triangle are both detected. The difference between the gray value
of the square and the gray value of the background is
small. Therefore, the segmentation of the square is much slower
compared to the segmentation of the black triangle and the white
ring. The energy decreases only slightly after the two objects have
already been detected. The segmentation of the color image of Figure
\ref{fig:results_3balls} is finished after $300$ time steps. In this
test image the color of the different objects significantly
differ. Therefore, the regions are easy to detect.

\begin{figure}
\centering
\includegraphics[viewport = 150 280 450 575, width =0.24\textwidth]{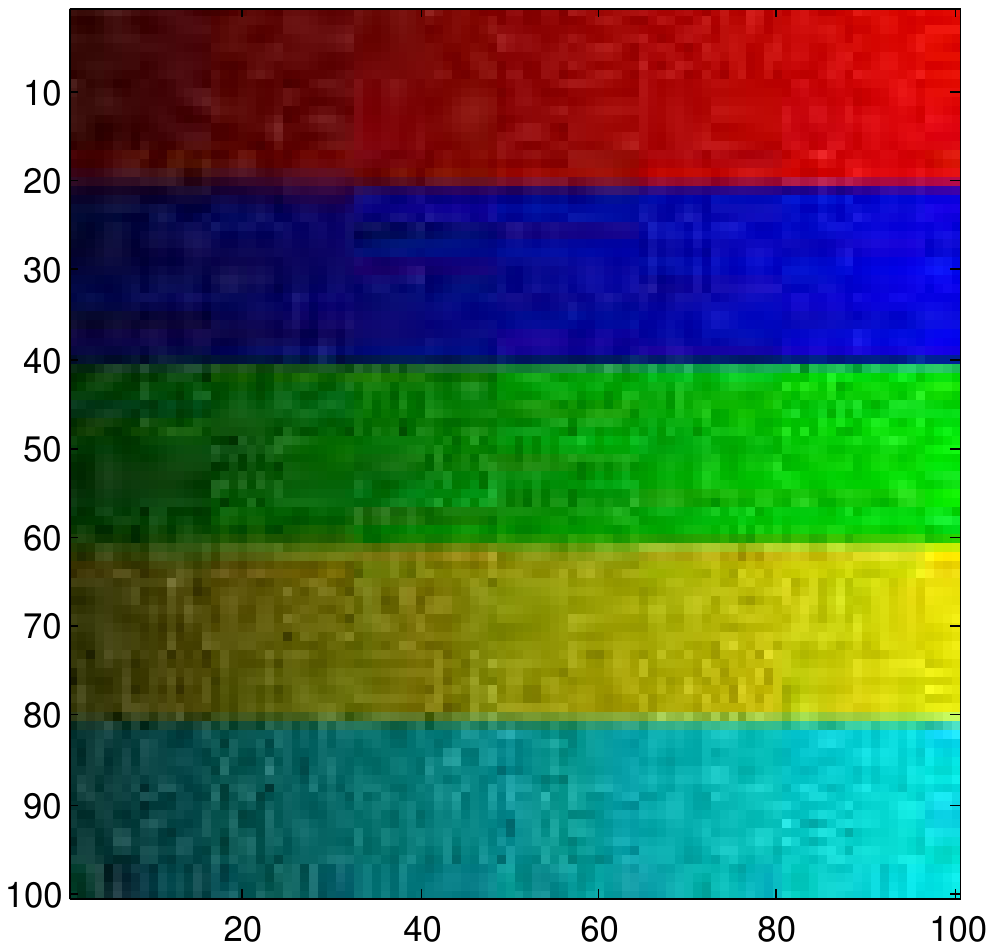}
\includegraphics[viewport = 150 280 450 575, width = 0.24\textwidth]{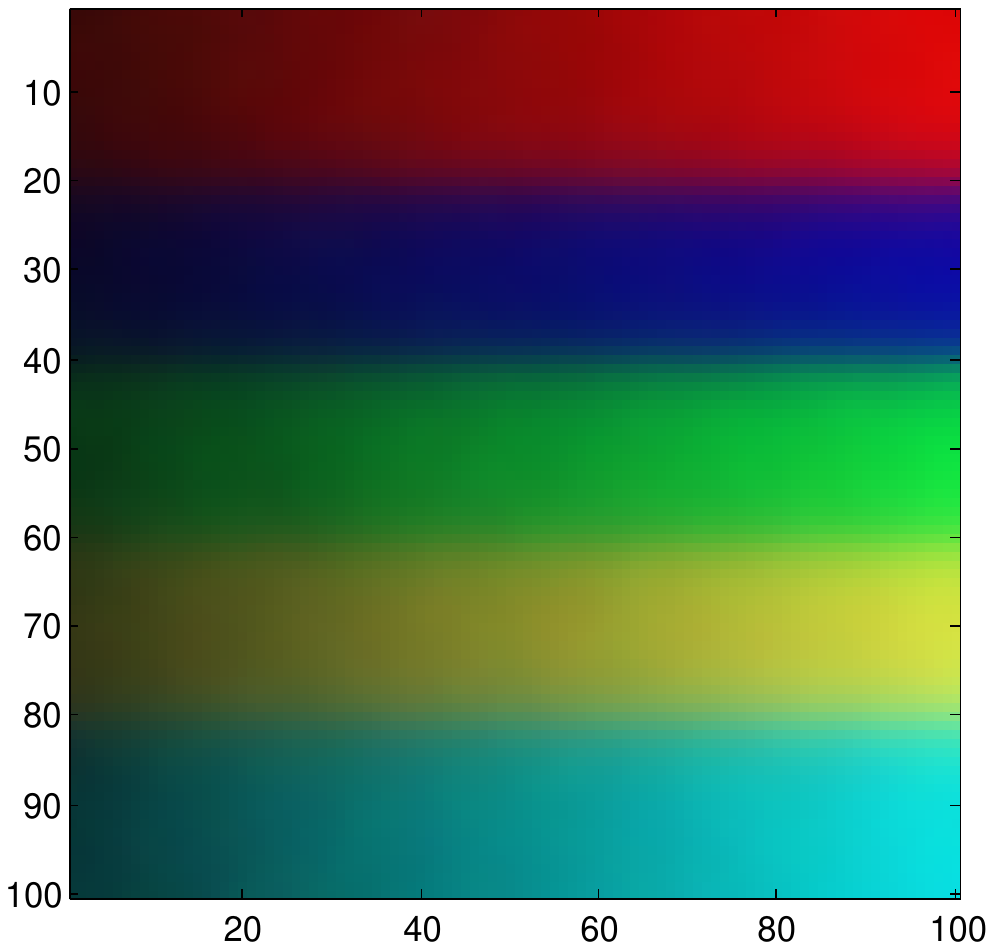}
\includegraphics[viewport = 150 280 450 575, width =0.24\textwidth]{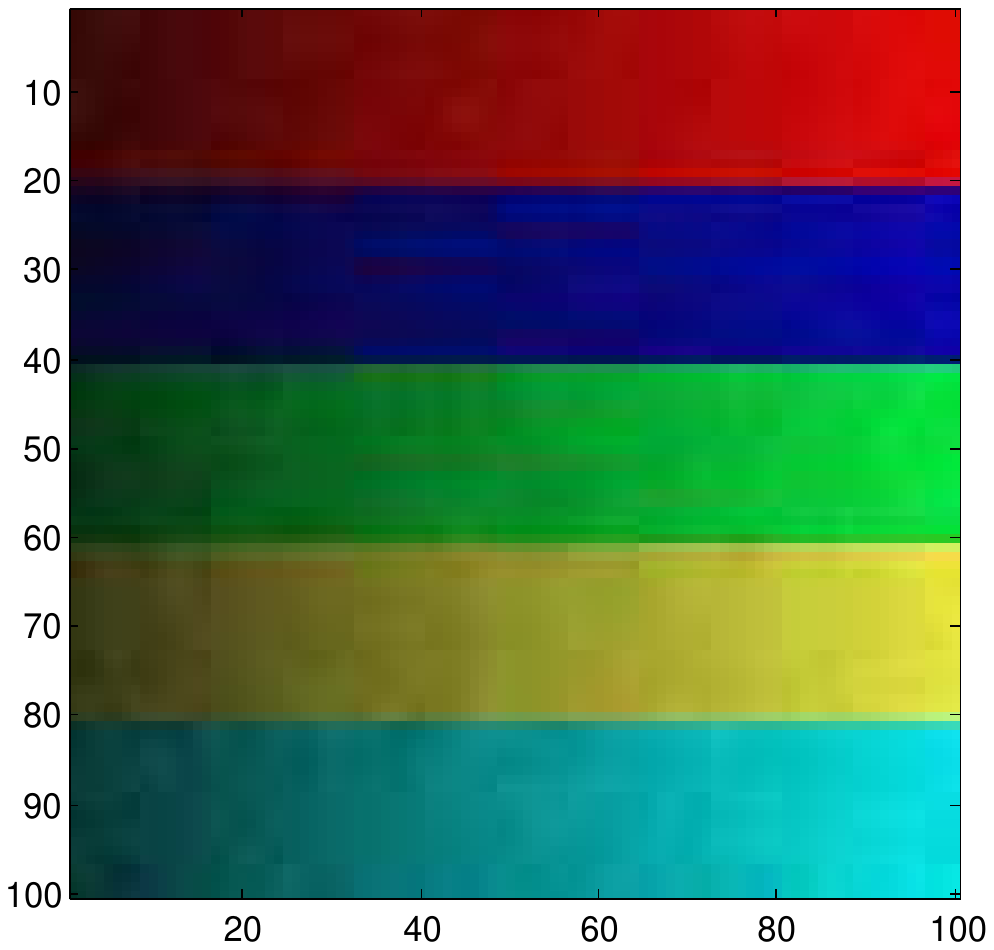}
\includegraphics[viewport = 150 280 450 575, width = 0.24\textwidth]{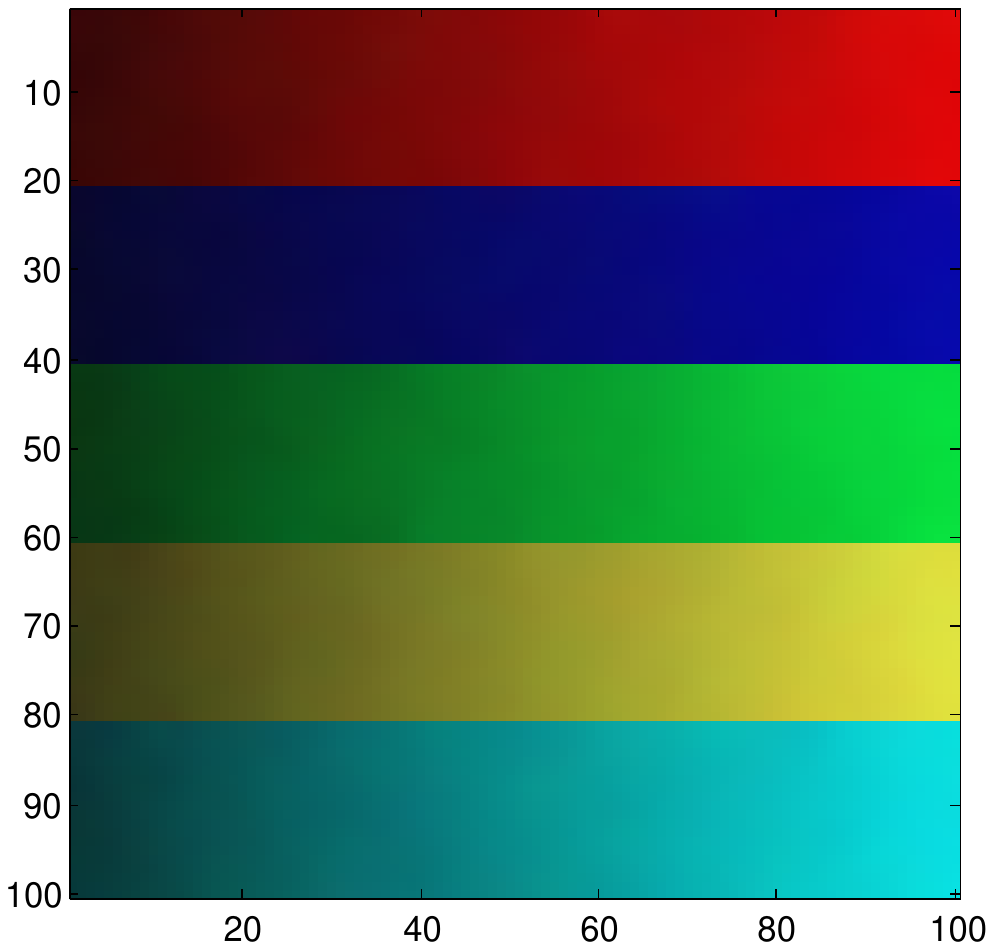}
\caption{Test image denoising: original image, image denoising without edge enhancement with $\lambda=0.1$, image denoising result with edge enhancement with $\lambda=1$ and $\lambda=0.1$}
\label{fig:results_denoising}
\end{figure} 

\begin{figure}
\centering
\includegraphics[viewport = 150 280 450 560, width = 0.24\textwidth]{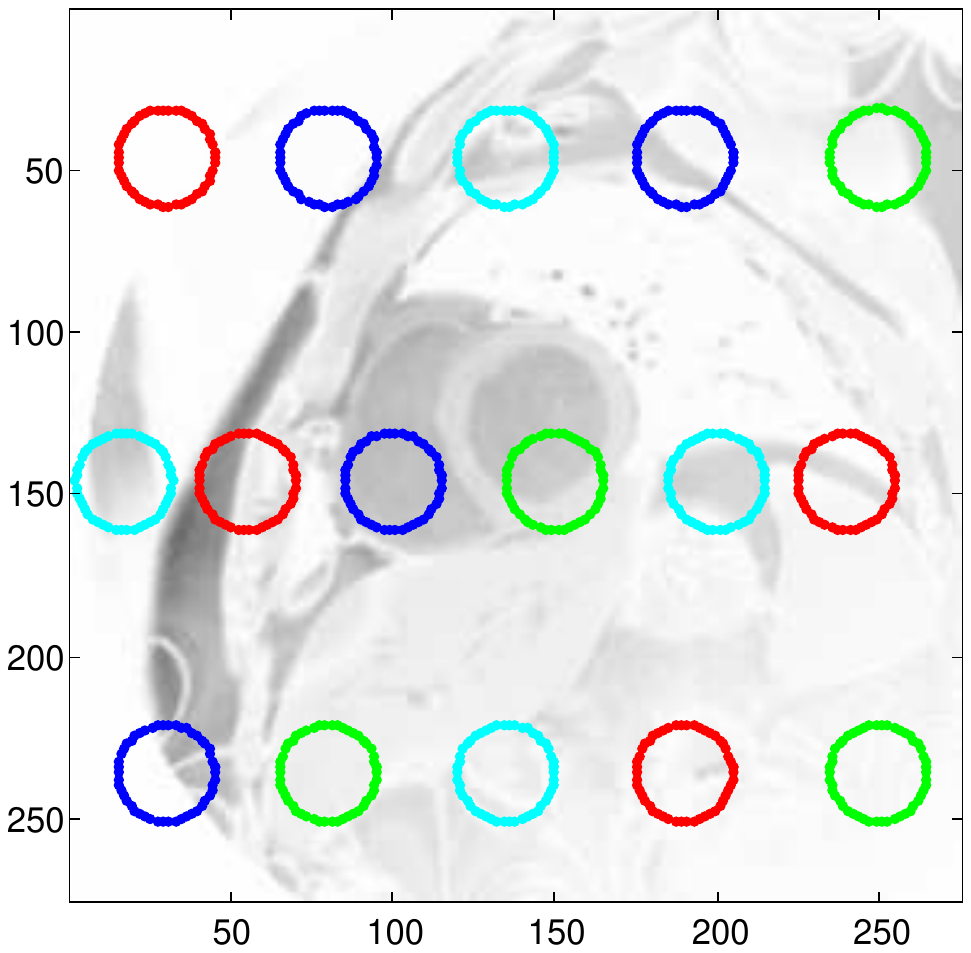}
\includegraphics[viewport = 150 280 450 560, width = 0.24\textwidth]{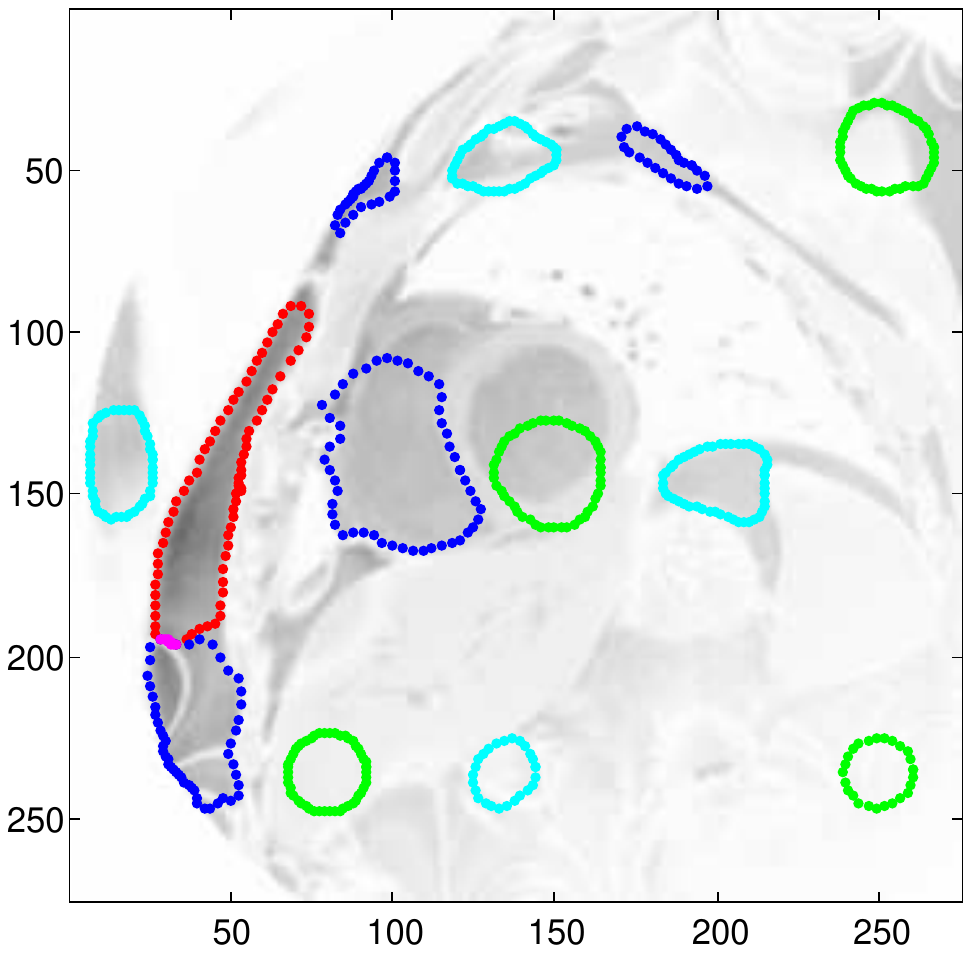}
\includegraphics[viewport = 150 280 450 560, width = 0.24\textwidth]{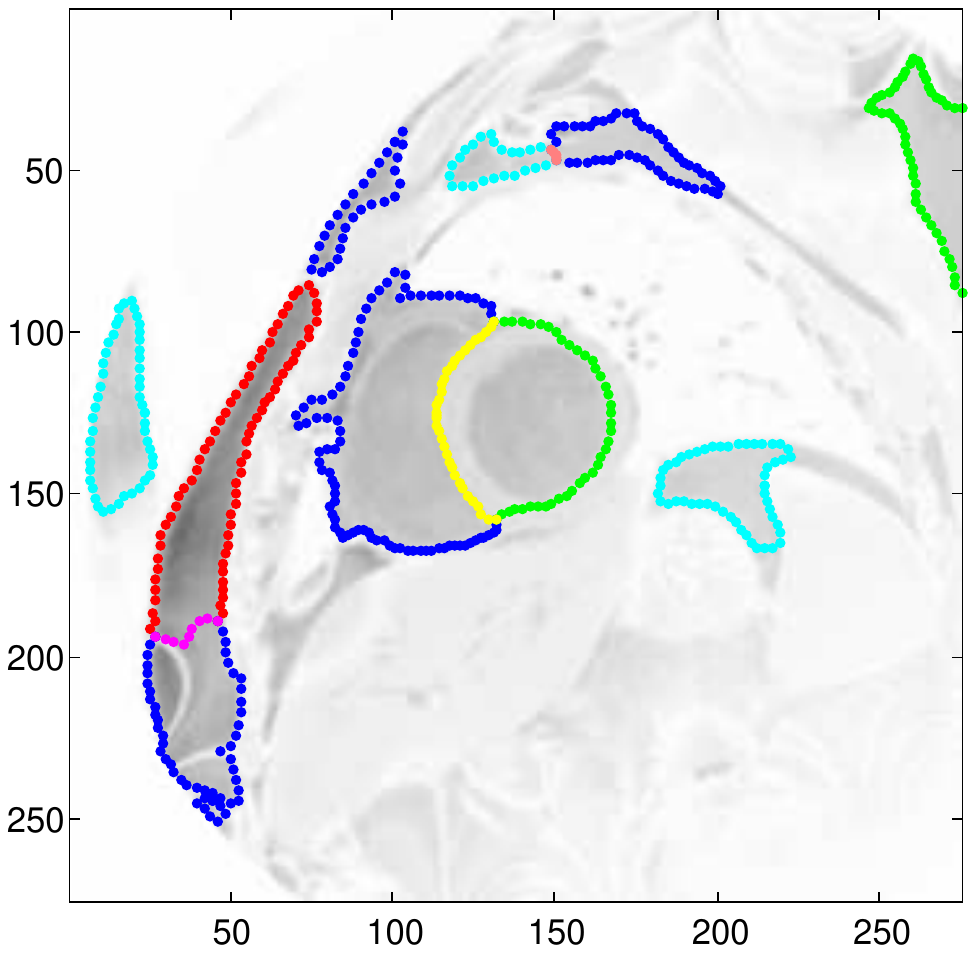}
\includegraphics[viewport = 150 280 450 560, width = 0.24\textwidth]{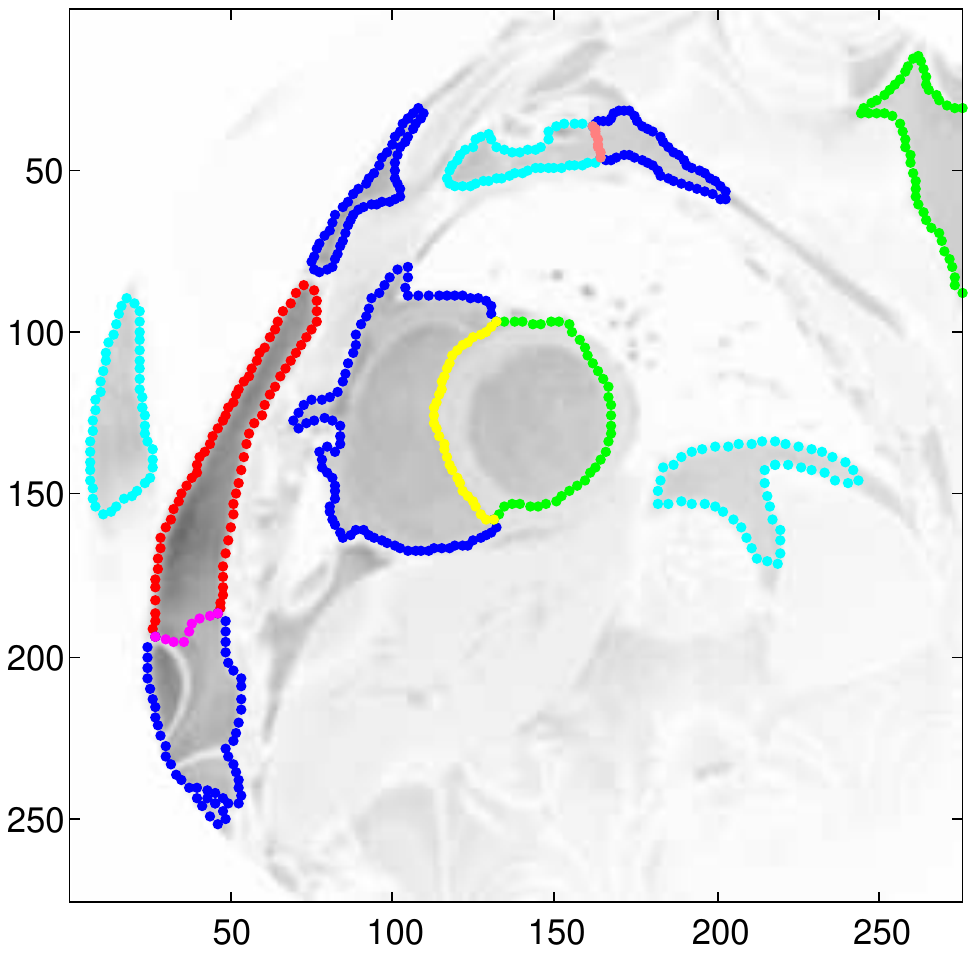}\\
\includegraphics[viewport = 150 280 450 560, width = 0.24\textwidth]{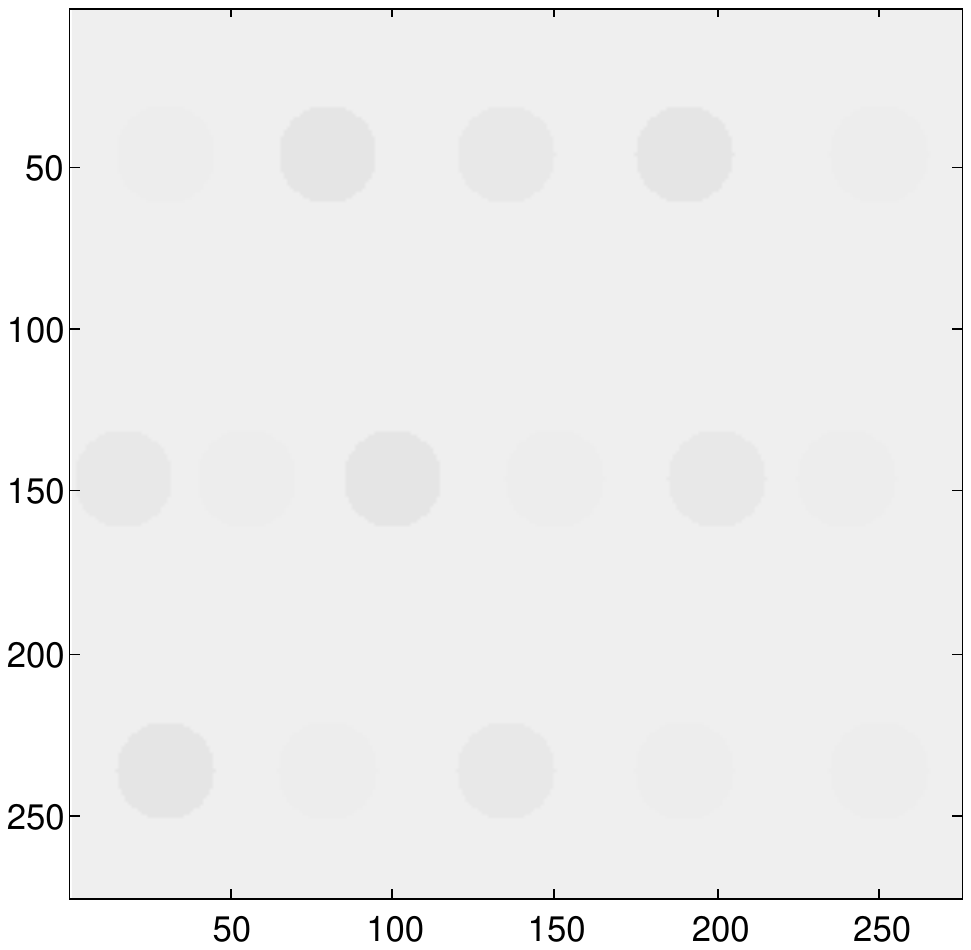}
\includegraphics[viewport = 150 280 450 560, width = 0.24\textwidth]{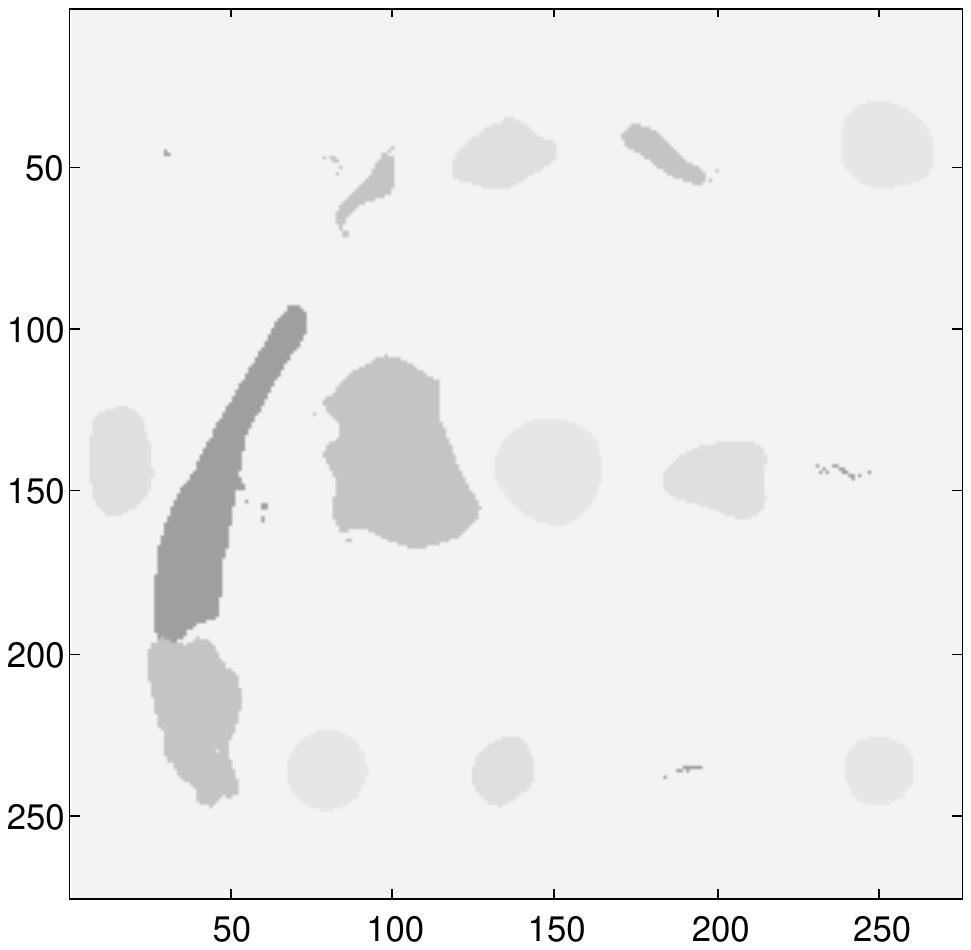}
\includegraphics[viewport = 150 280 450 560, width = 0.24\textwidth]{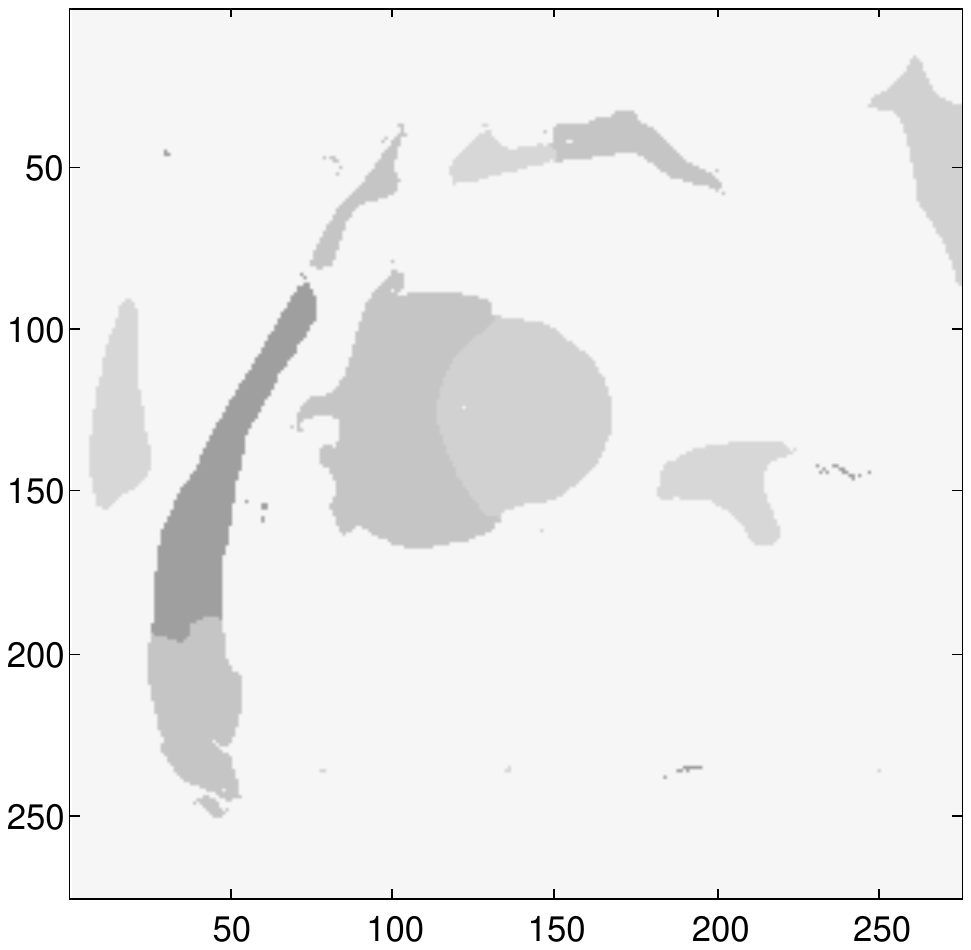}
\includegraphics[viewport = 150 280 450 560, width = 0.24\textwidth]{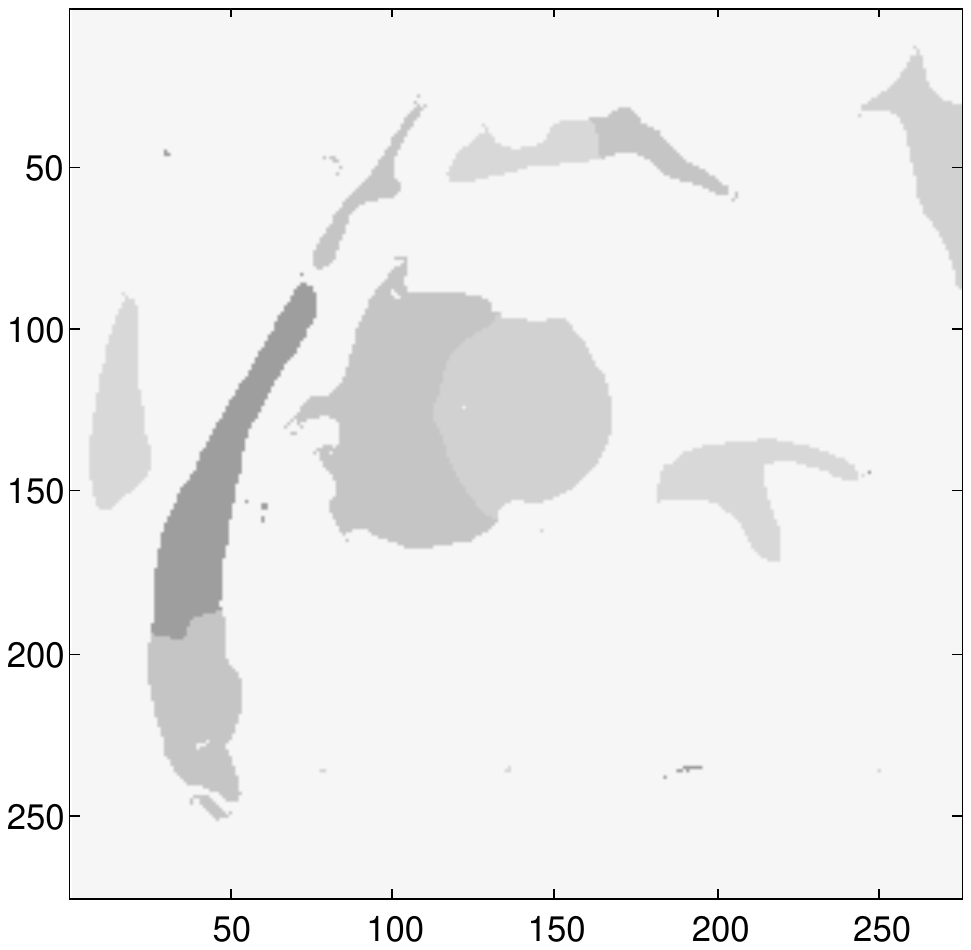}\\
\includegraphics[viewport = 170 300 430 540, width = 0.24\textwidth]{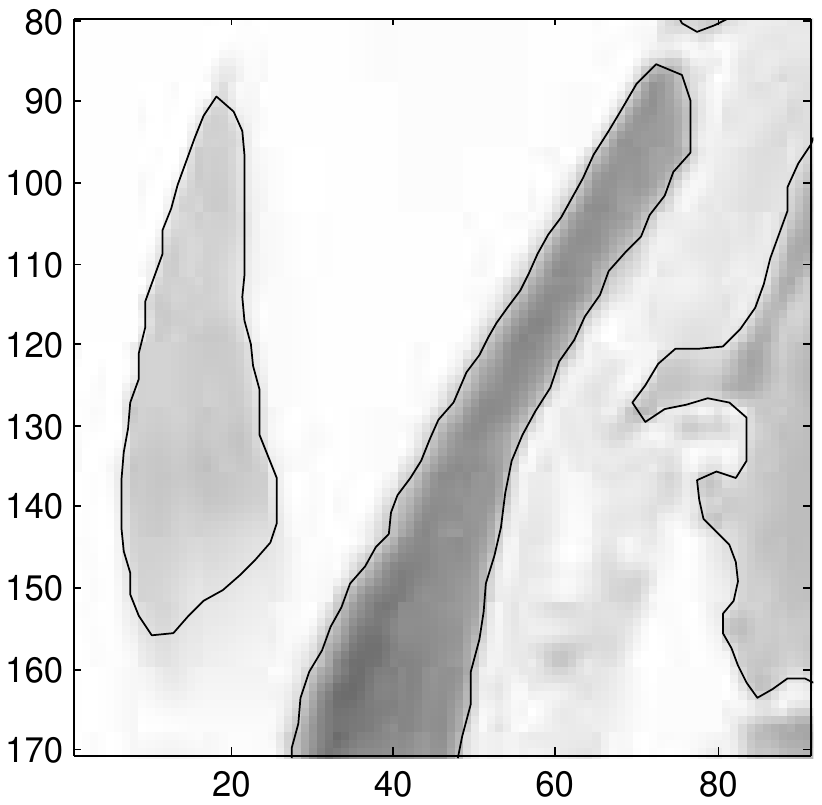}
\includegraphics[viewport = 150 280 450 560, width = 0.24\textwidth]{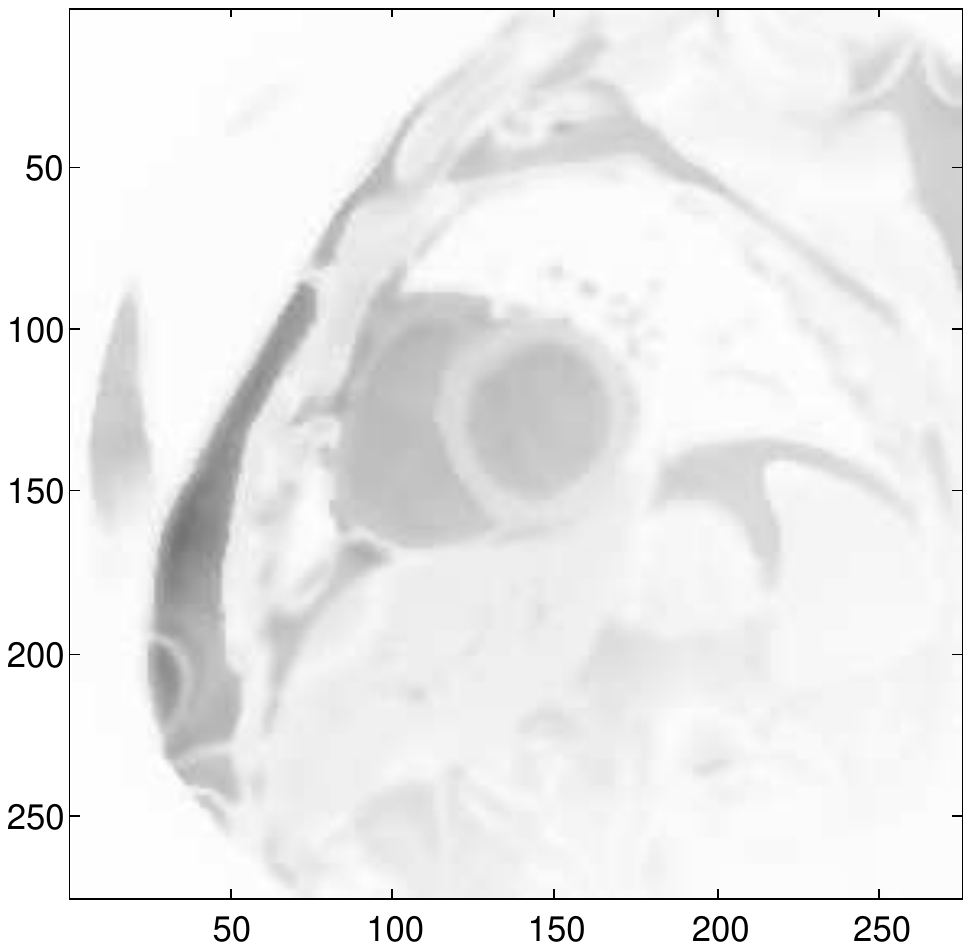}
\includegraphics[viewport = 150 280 450 560, width = 0.24\textwidth]{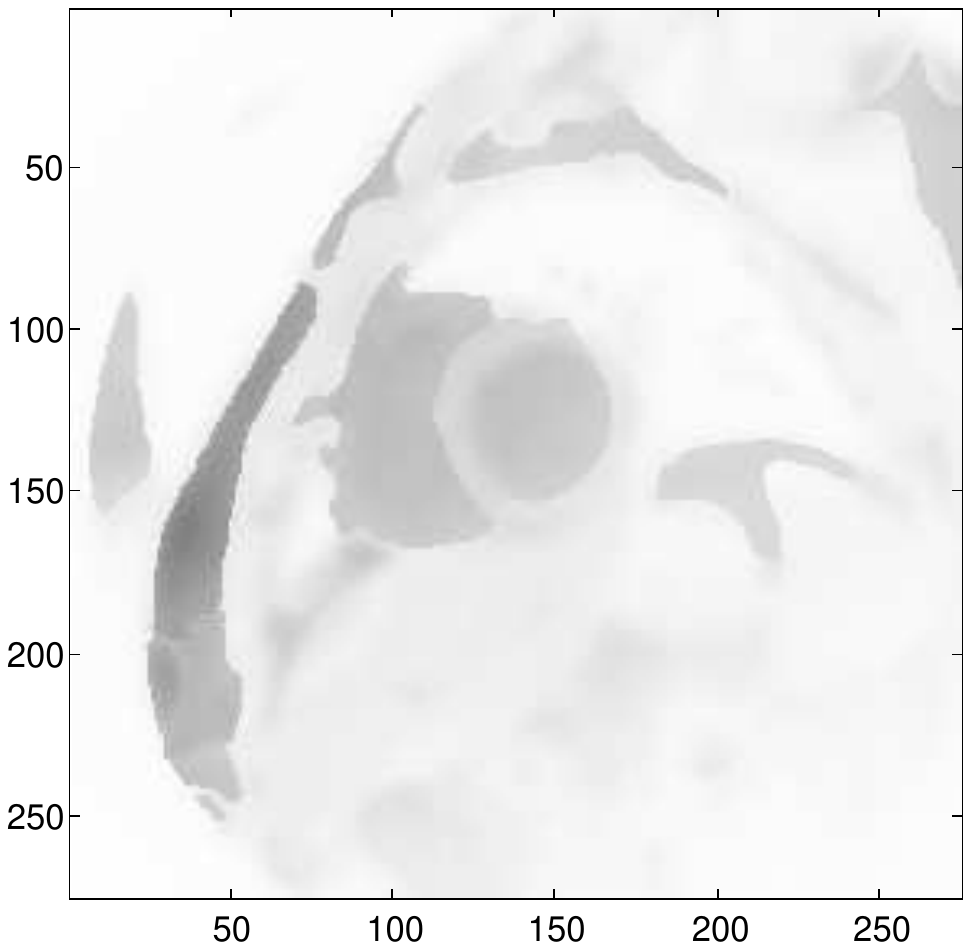}
\caption{Multi-phase image segmentation of a medical image, $m=1,100,500,1500$, $\Delta t=0.02$, $\lambda = 400$, $\sigma$-factor $5\%$, last row: magnification of final segmentation to demonstrate a weak edge, approximation $u$ computed in the post-processing step with $\lambda_k=1$ and $\lambda_k=0.1$, $k=1,\ldots,N_R$}
\label{fig:medical}
\end{figure}


In Section \ref{subsec:image_smoothing} we proposed a method for image
smoothing. The partial differential equation $-\frac1\lambda \Delta u
+ u = u_0$ with Neumann boundary conditions is solved on each phase
separately. Having previously detected the regions, the image
smoothing is performed as a post-processing step. Figure
\ref{fig:results_denoising} shows a noisy image (1st sub-figure) with
five differently colored stripes. The brightness increases from left
to right. The second sub-figure to the left shows the result solving the diffusion
equation on $\Omega$ neglecting the previously detected regions. The
edges are strongly smoothed out.

This motivates to search for an approximation of the image which
enhances the edges. It is necessary to use a piecewise smooth
approximation since a piecewise constant approximation would be a too
strong simplification and the brightness change of the original image
would get lost. The smoothing effect severely depends on the parameter
$\lambda$. If $\lambda$ is large, $\frac1\lambda$ is small and the
approximation $u$ is close to the original image $u_0$. The smaller
$\lambda$ on the contrary, the bigger is the smoothing
effect. Therefore, choosing $\lambda=1$ the noise is not completely
smoothed out, cf. third sub-figure in Figure
\ref{fig:results_denoising}. Choosing $\lambda=0.1$ results in a
smooth approximation $u$ of $u_0$, cf. forth sub-figure. The change in
the brightness from left to right is still conserved. By solving the
diffusion equation with Neumann boundary conditions separately in each
phase, the edges remain sharp.

Additional to artificial test images, the segmentation technique is
applied on real images. Figure \ref{fig:medical} presents the
segmentation of a medical image. It shows the original image and the
segmentation for $m=1,100,500$ and $1500$ (1st row) and the piecewise
constant approximation (2nd row). The parameter
$\lambda$ which weights the external forcing term is chosen high
($\lambda=400$), as the brightness differences of some objects and the
background is small. The $\sigma$-factor is set to $5\%$. Additional,
a lower limit of $\sigma_\mathrm{min}=15$ for $m<400$ and
$\sigma_\mathrm{min}=5$ for $m\geq 400$ is applied. A magnification of the final
segmentation demonstrates the ability of region based methods to
handle weak edges (3rd row, left). The Figure additionally presents
the final piecewise smooth approximation for two different parameters
$\lambda$ (3rd row, sub-figure 2-3). This example also demonstrates
the creation of new interfaces and triple junctions. 

\begin{figure}
\centering
\includegraphics[viewport = 150 280 450 560, width = 0.24\textwidth]{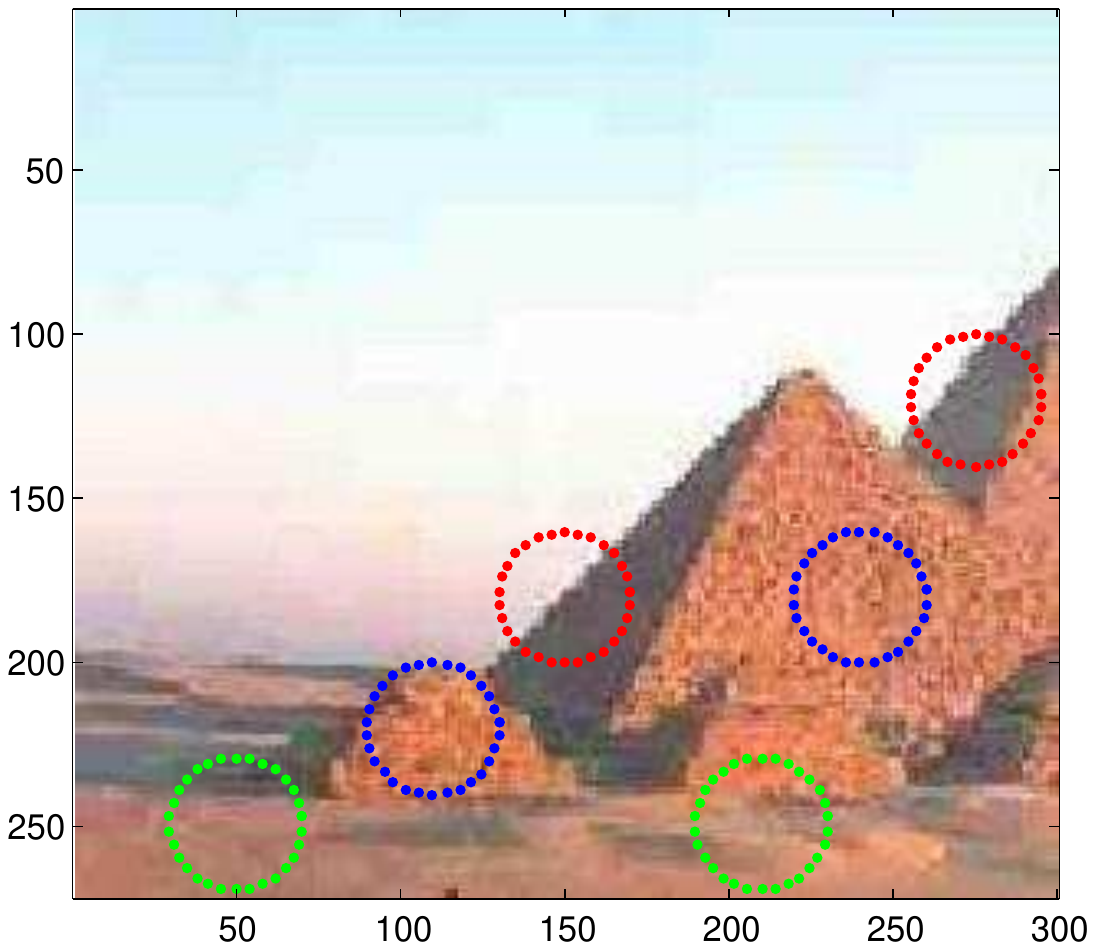}
\includegraphics[viewport = 150 280 450 560, width = 0.24\textwidth]{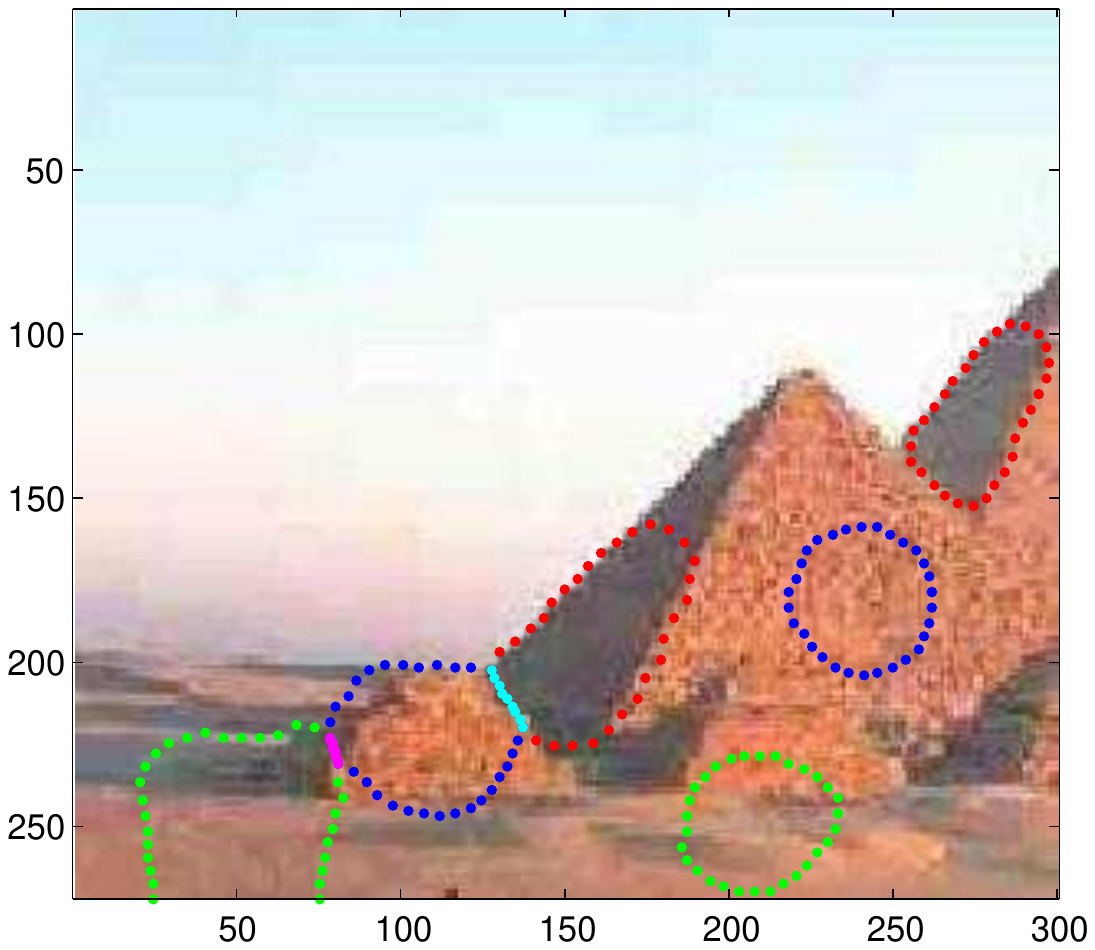}
\includegraphics[viewport = 150 280 450 560, width = 0.24\textwidth]{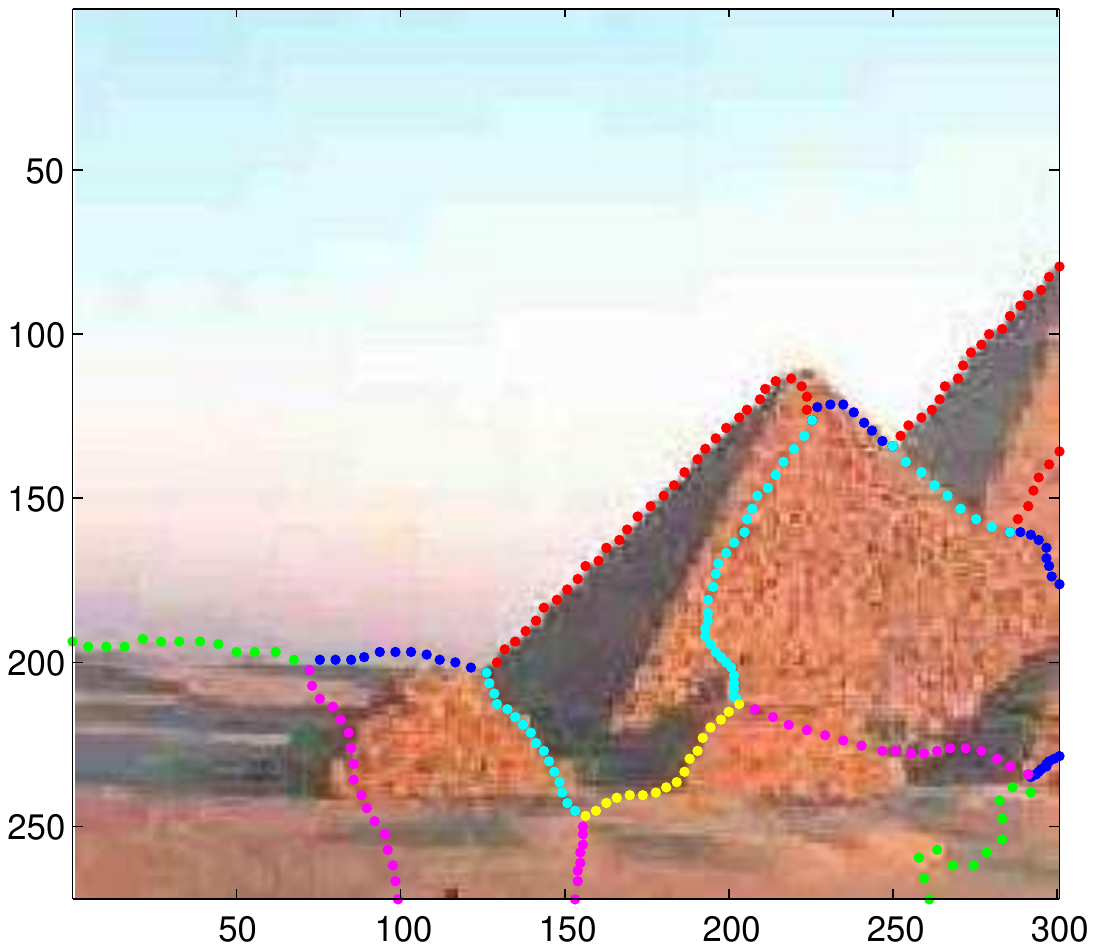}
\includegraphics[viewport = 150 280 450 560, width = 0.24\textwidth]{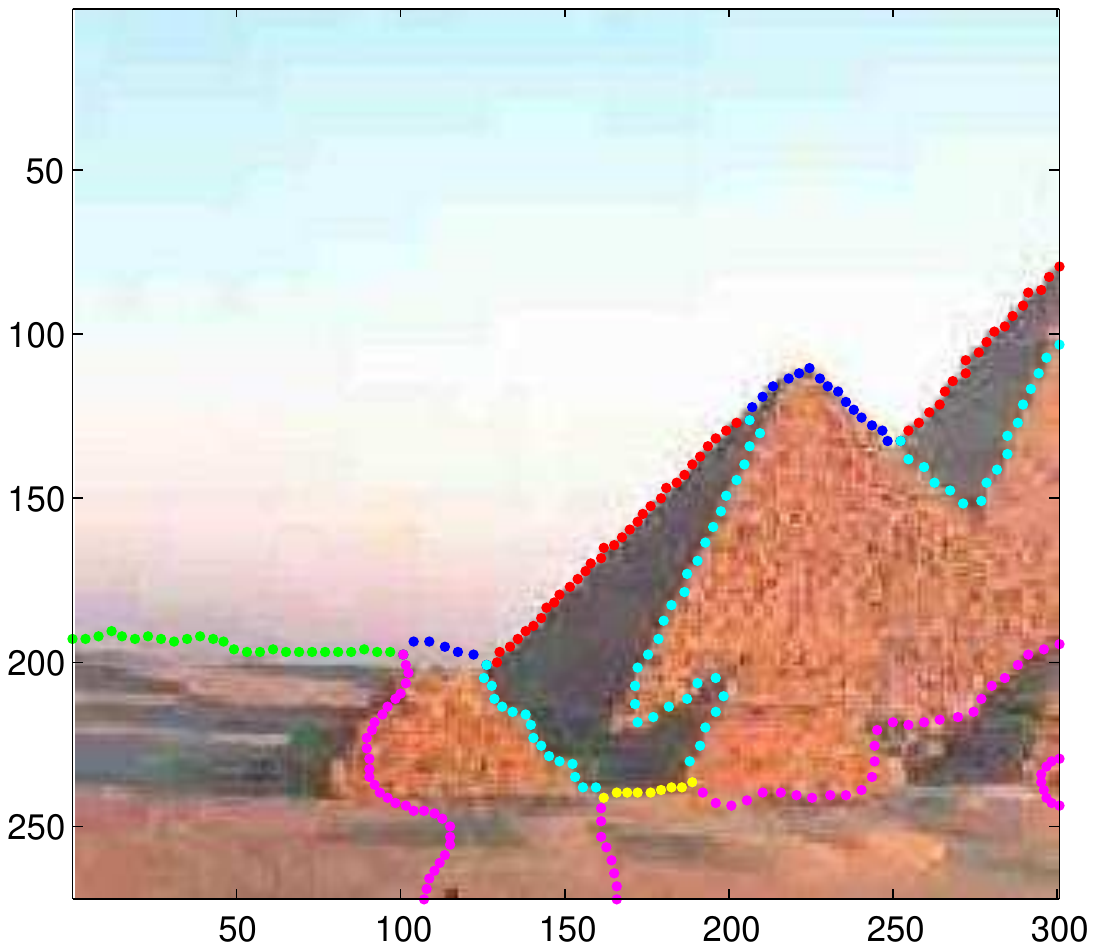}\\
\includegraphics[viewport = 150 280 450 560, width = 0.24\textwidth]{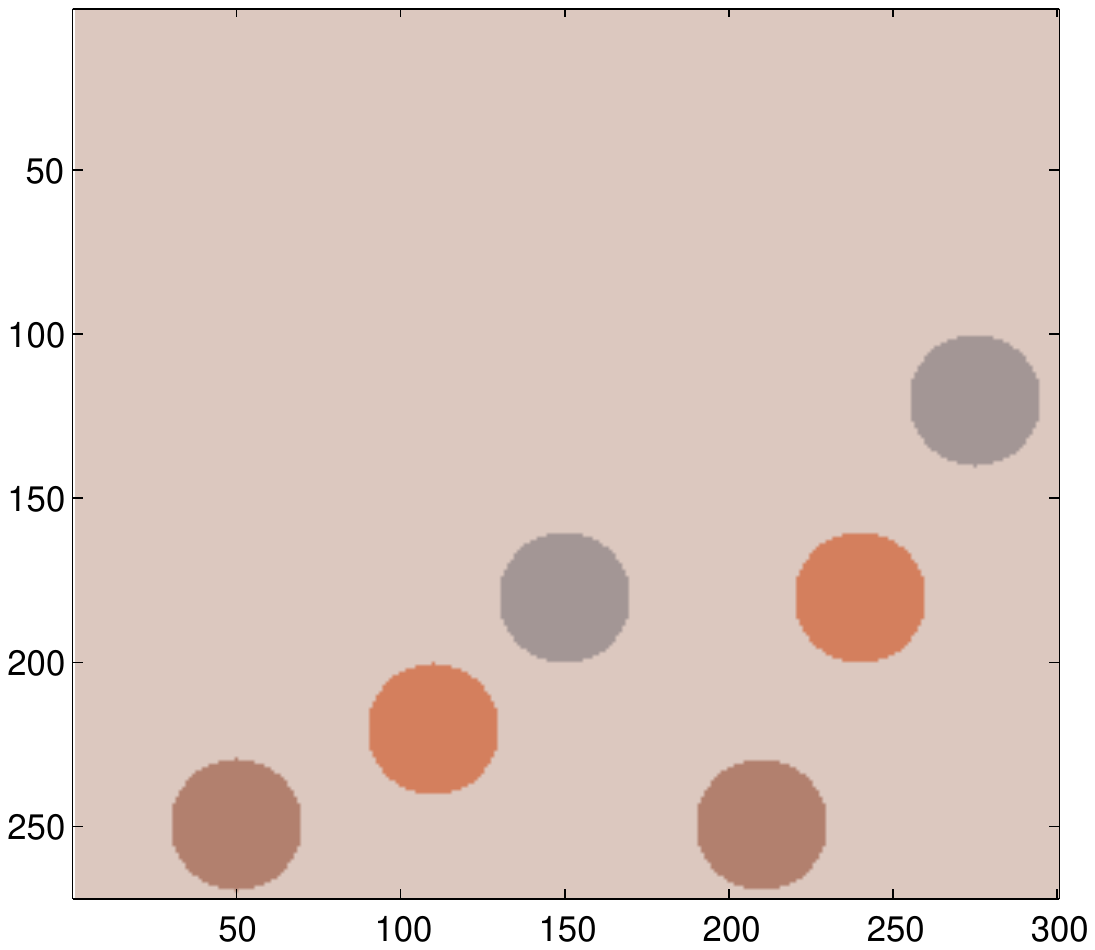}
\includegraphics[viewport = 150 280 450 560, width = 0.24\textwidth]{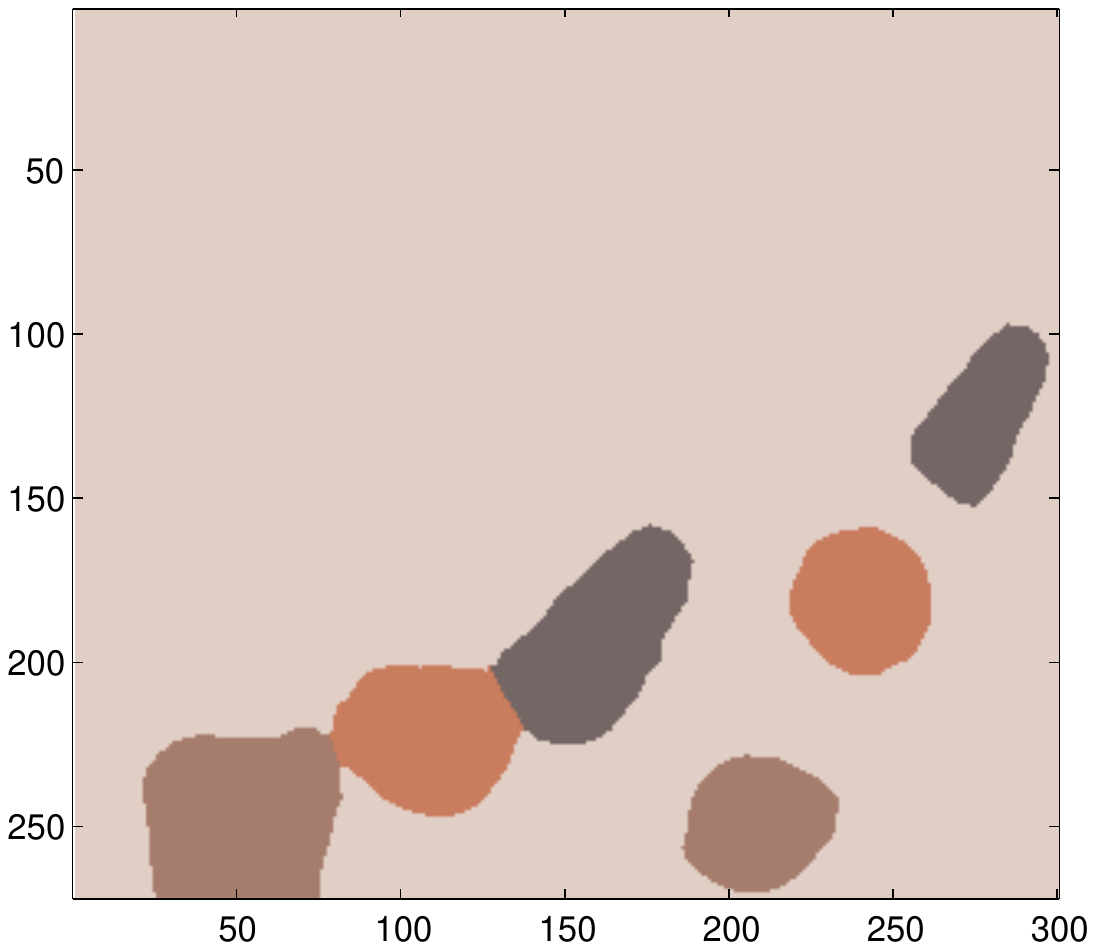}
\includegraphics[viewport = 150 280 450 560, width = 0.24\textwidth]{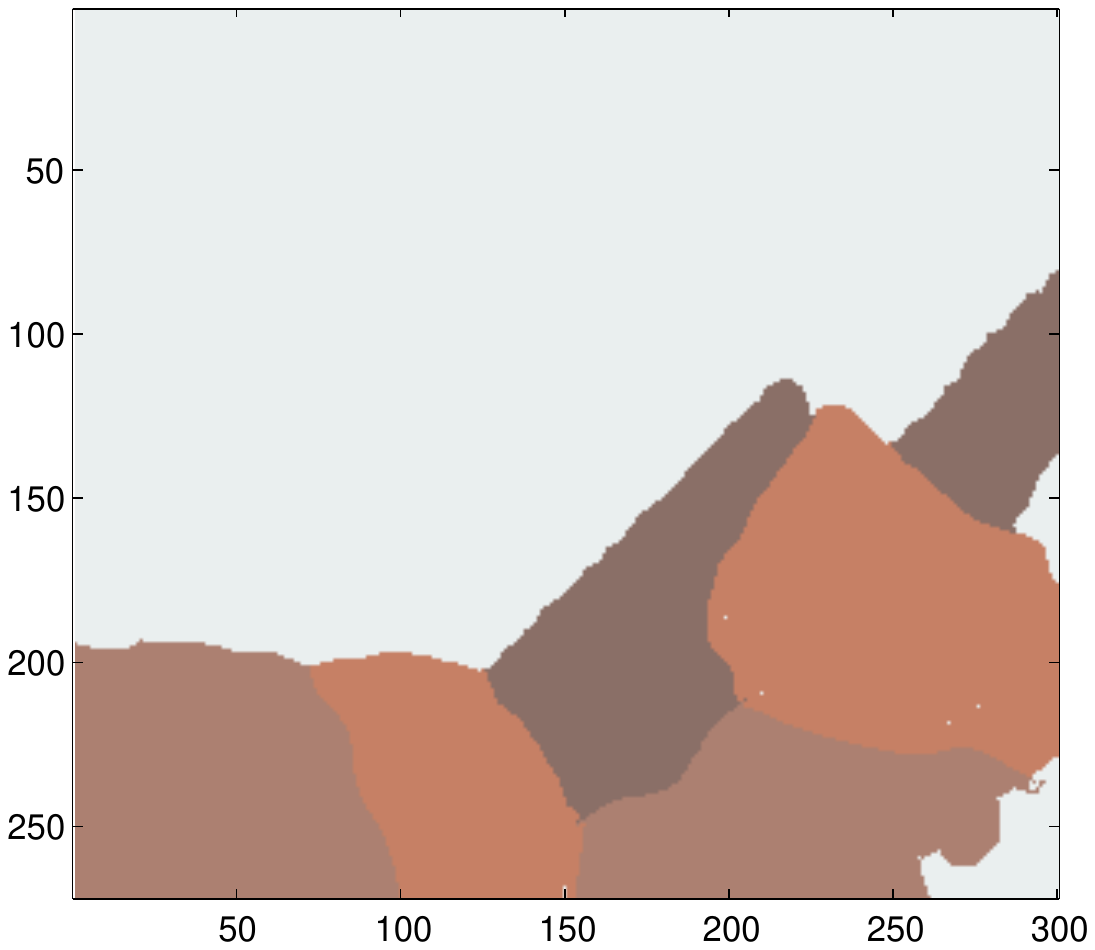}
\includegraphics[viewport = 150 280 450 560, width = 0.24\textwidth]{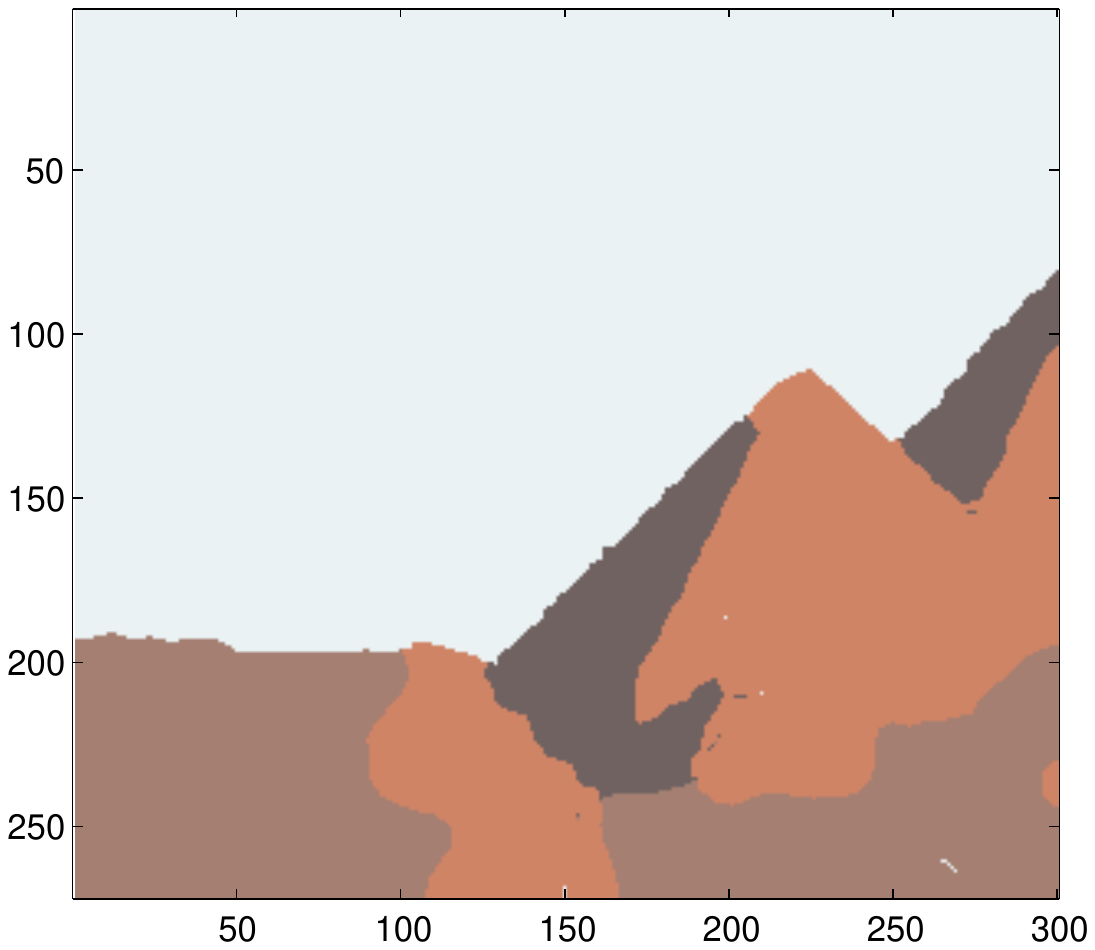}\\
\includegraphics[viewport = 150 280 460 560, width = 0.24\textwidth]{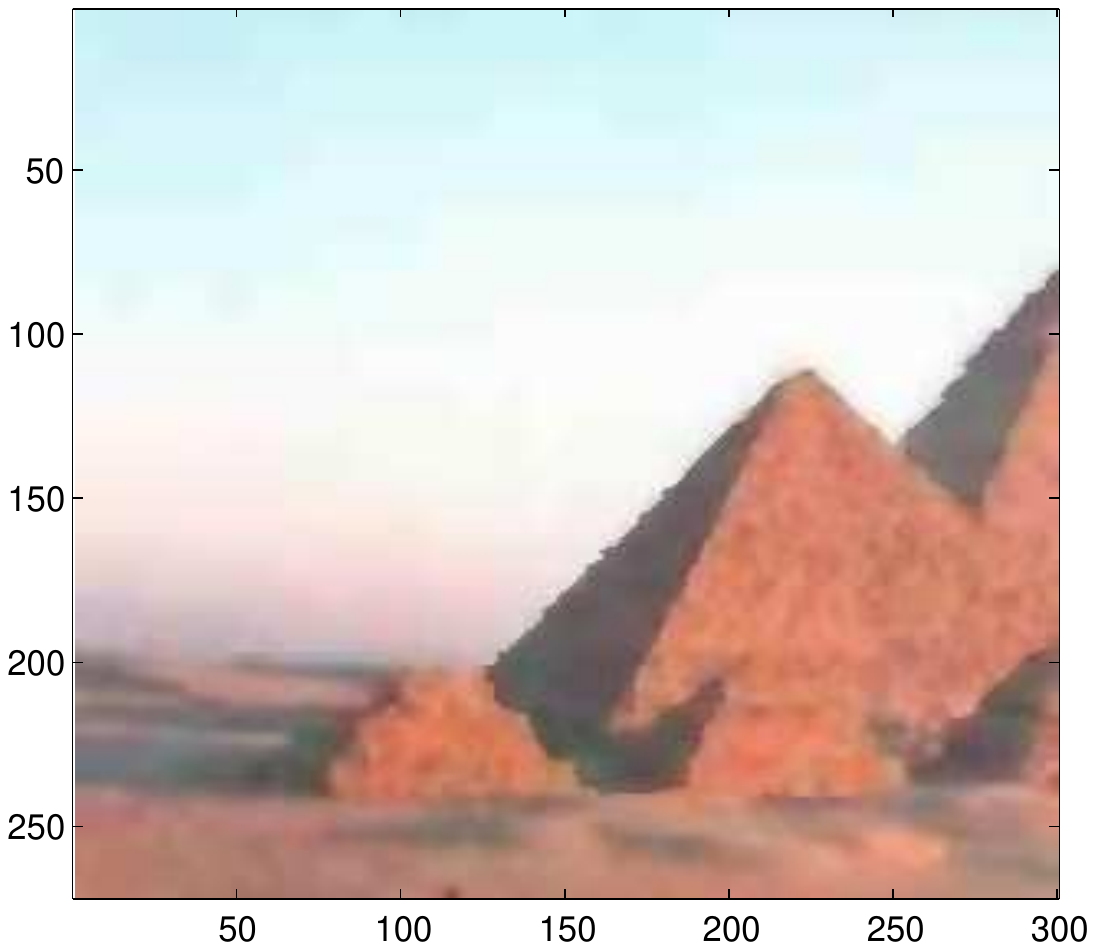}
\includegraphics[viewport = 150 280 460 560, width = 0.24\textwidth]{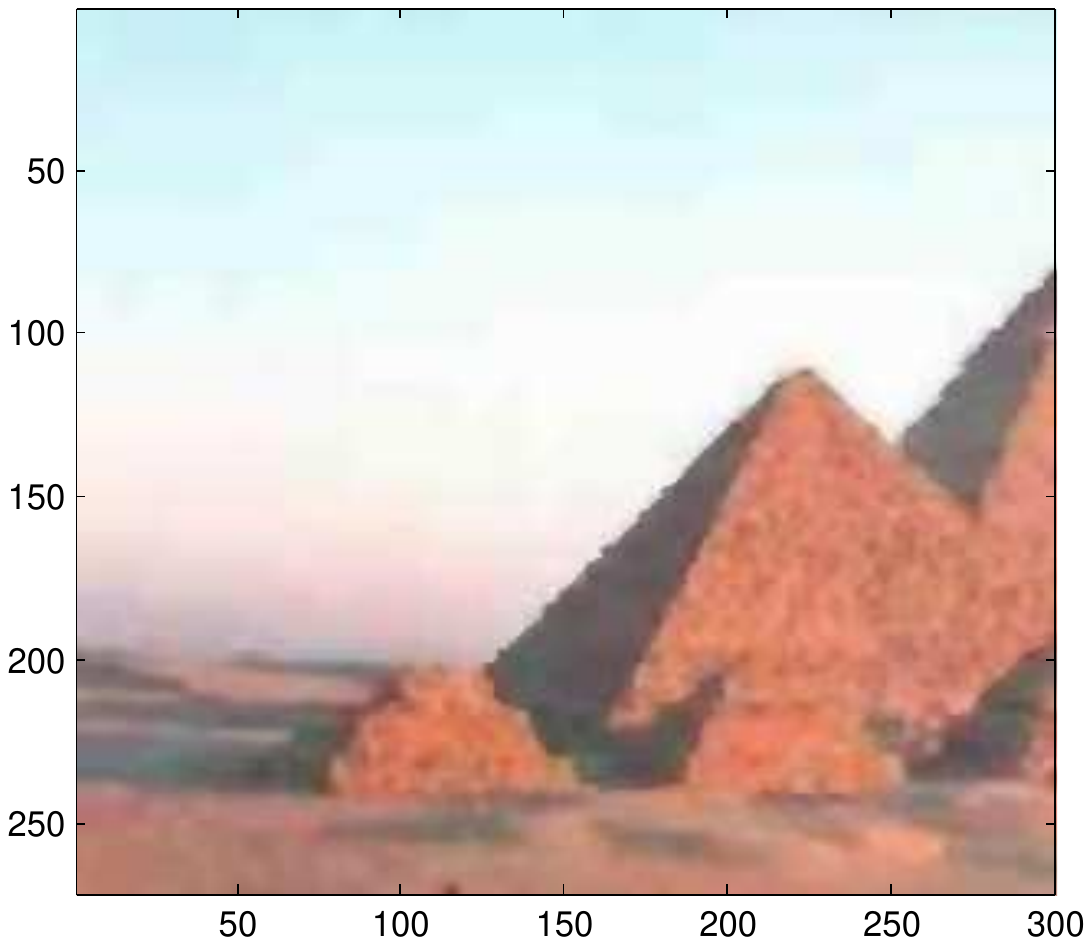}
\includegraphics[viewport = 105 280 475 570, width = 0.3\textwidth]{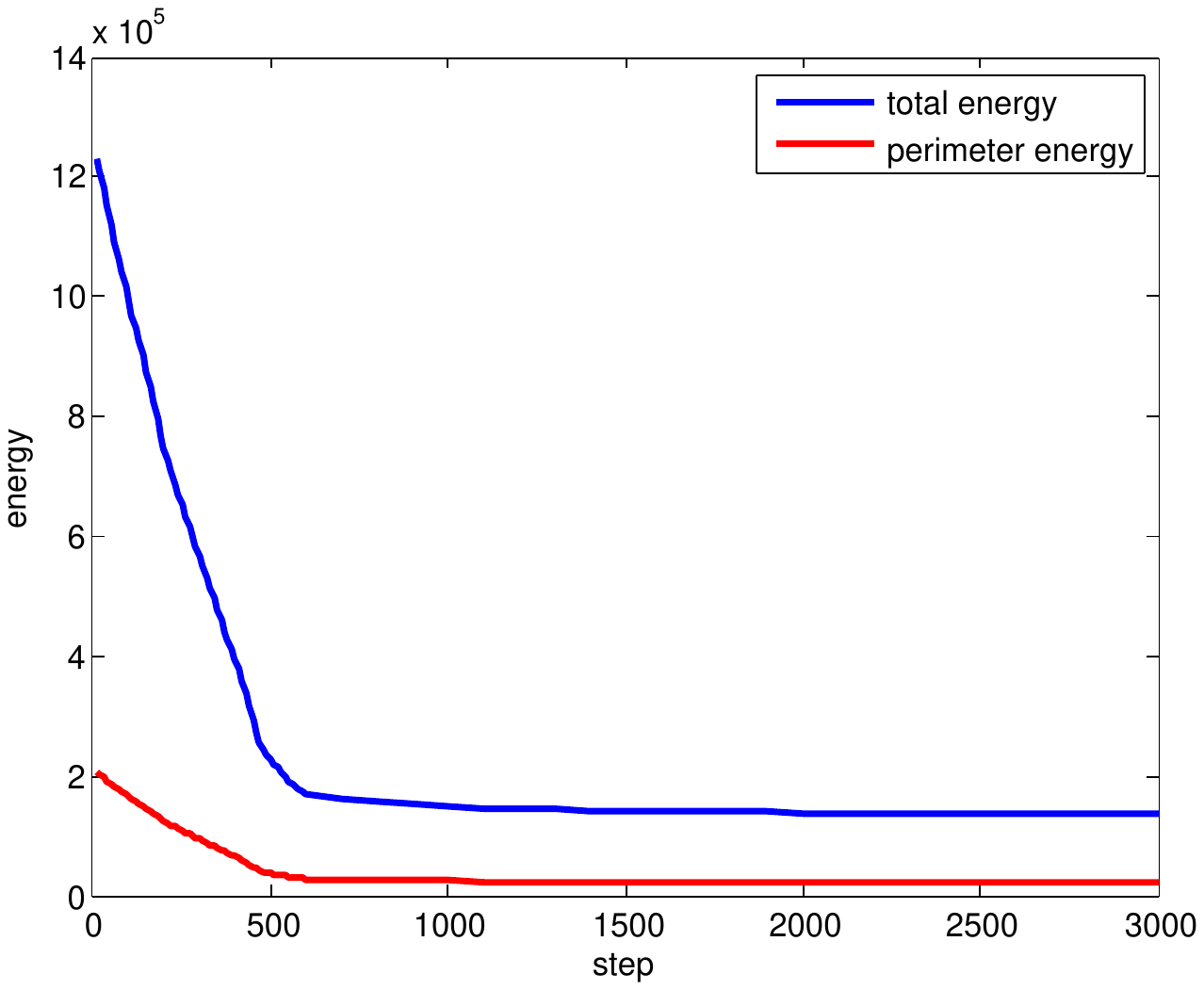}
\caption{Multi-phase image segmentation of a color image using CB space, image from Caltech database \citep{FeiFei04}, $m=1,100,500,3000$, $\Delta t=0.005$, $\lambda_C = 80$, $\lambda_B = 20$, $\sigma$-factor $20\%$, last row: approximation $u$ computed in the post-processing step with $\lambda_k=0.1$ and $\lambda_k=1$, $k=1,\ldots,N_R$, and plot of the Mumford Shah energy}
\label{fig:pyramide}
\end{figure}

Figure \ref{fig:pyramide} shows the result of segmenting a colored image from the Caltech image database \citep{FeiFei04}. 
For this experiment, the chromaticity-brightness color space is used. The weighting parameters for the external forcing terms are set to $\lambda_C = 80$ and $\lambda_B=20$ such that the chromaticity has a higher influence on the region based segmentation. The ability to handle multiple phases, topological changes (i.e. boundary intersection and creation of triple junctions) and vector-valued image data is demonstrated in this example.

\begin{figure}
\centering
\includegraphics[viewport = 120 300 470 540, width = 0.24\textwidth]{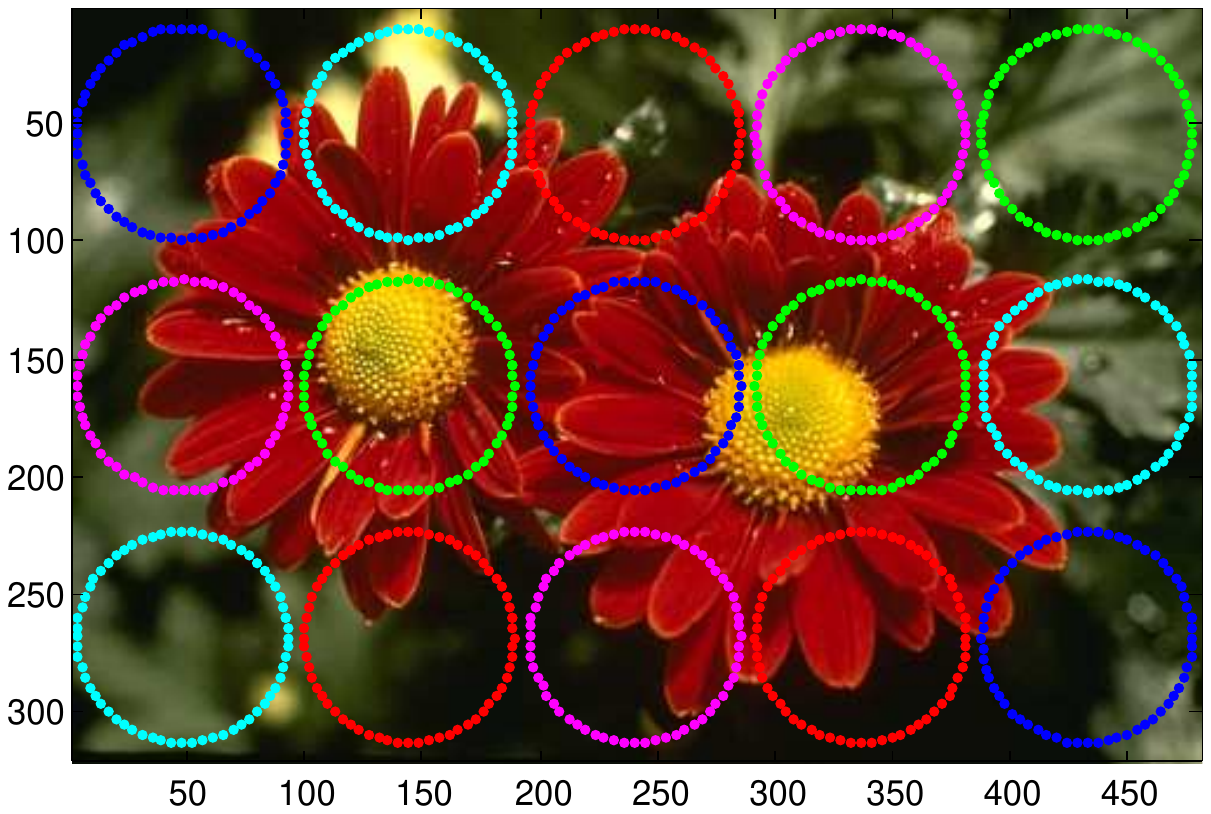}
\includegraphics[viewport = 120 300 470 540, width = 0.24\textwidth]{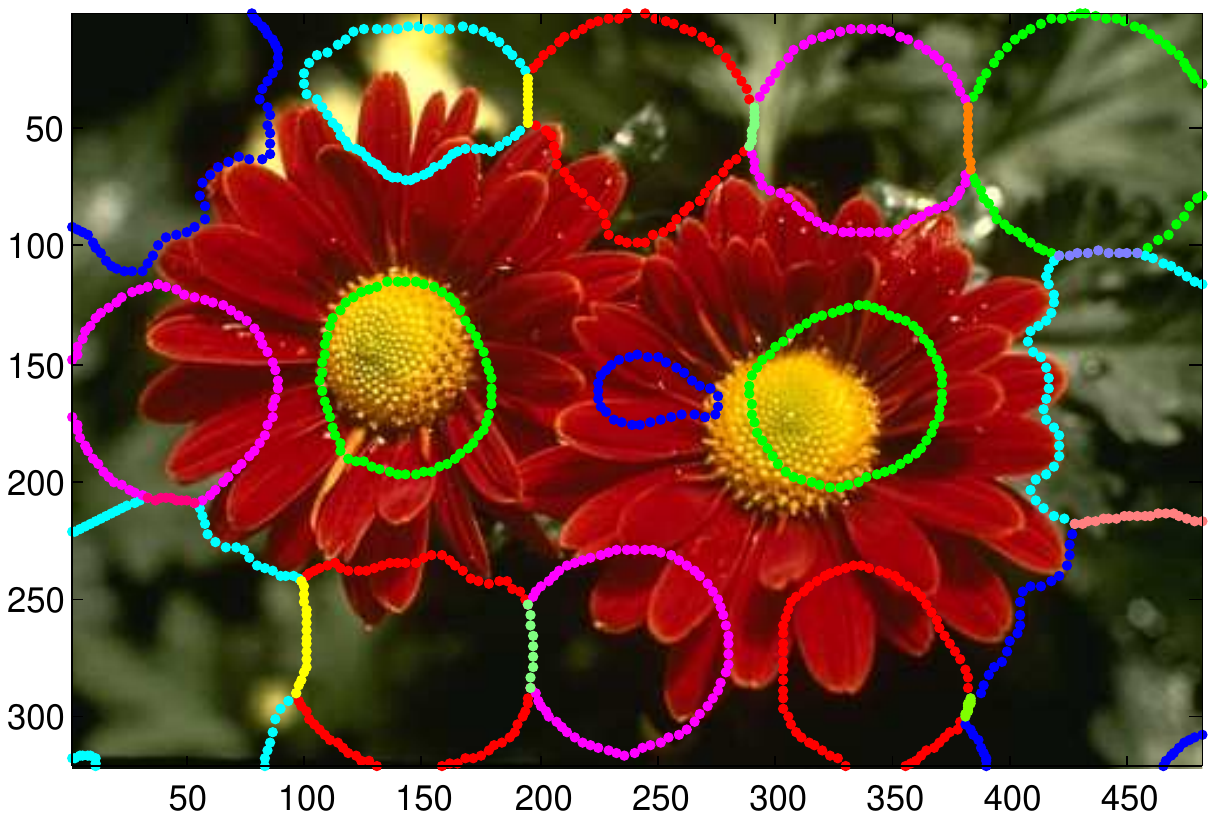}
\includegraphics[viewport = 120 300 470 540, width = 0.24\textwidth]{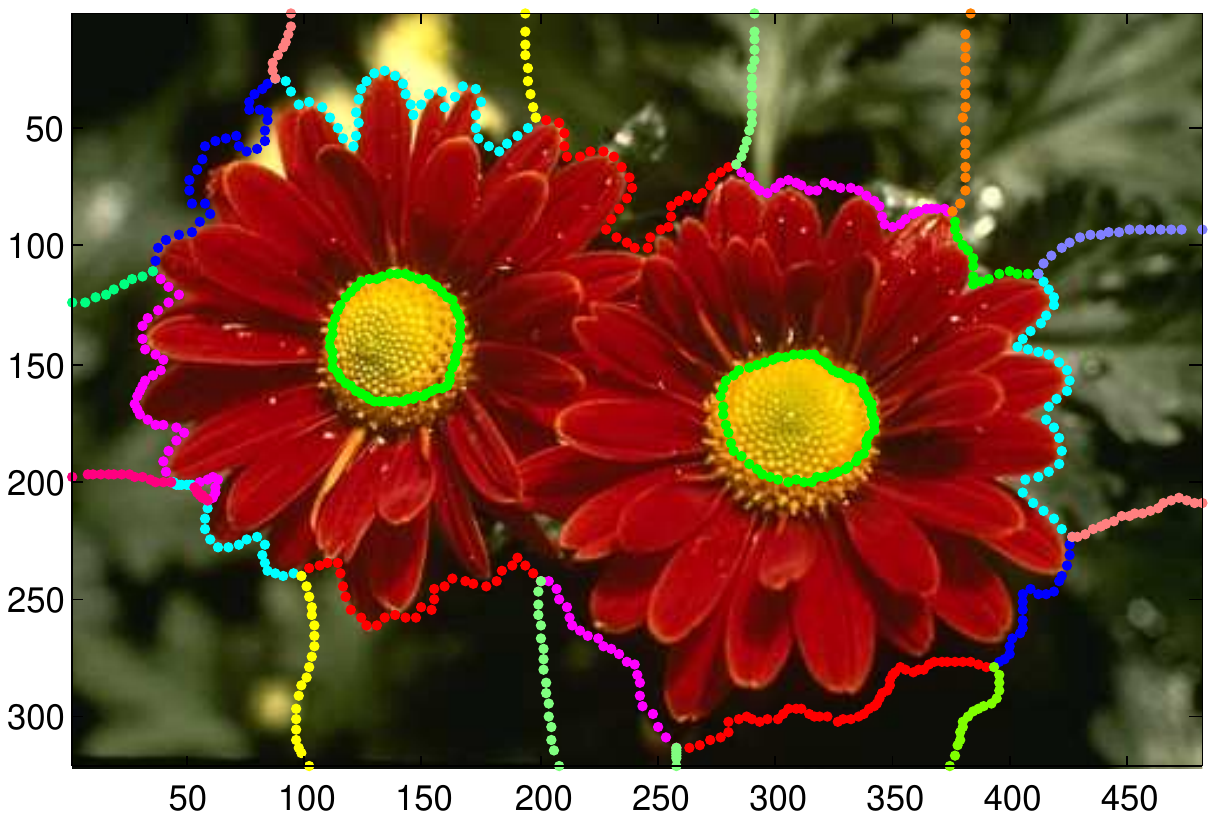}
\includegraphics[viewport = 120 300 470 540, width = 0.24\textwidth]{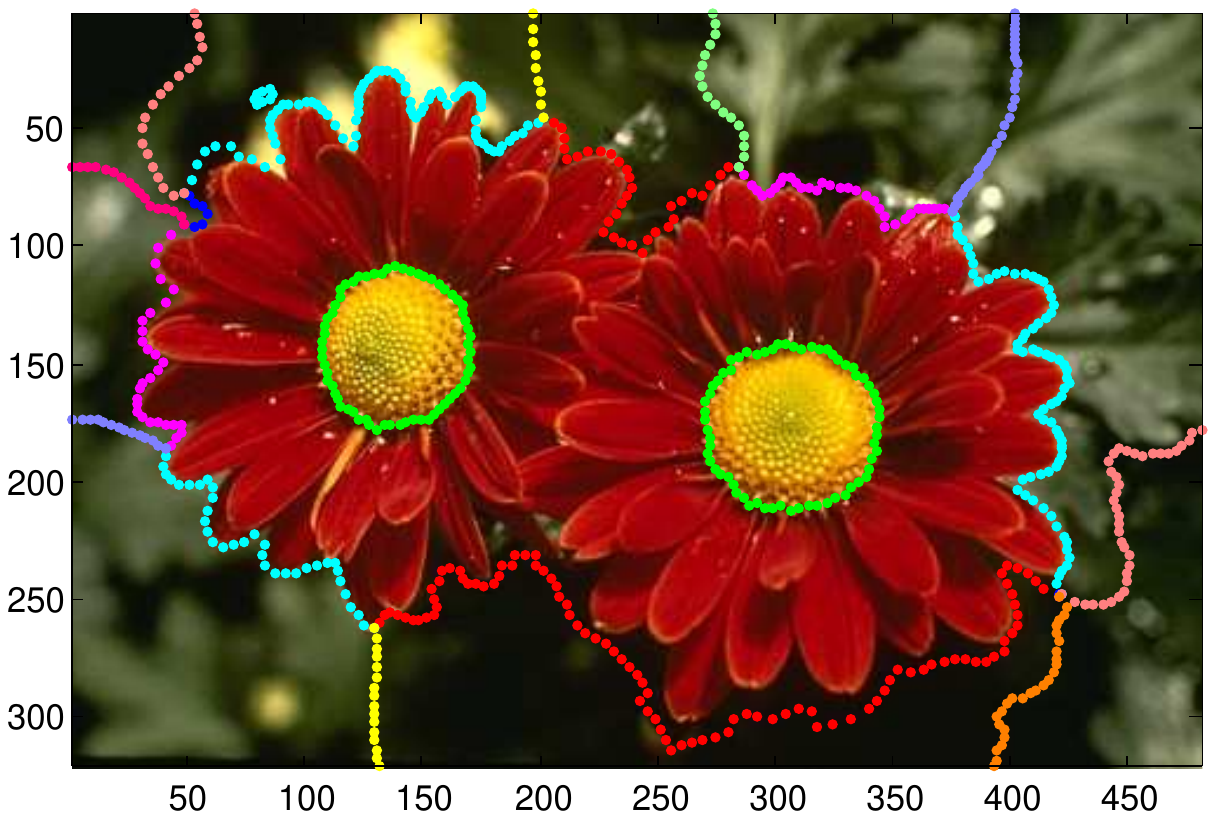}\\
\includegraphics[viewport = 120 300 470 540, width = 0.24\textwidth]{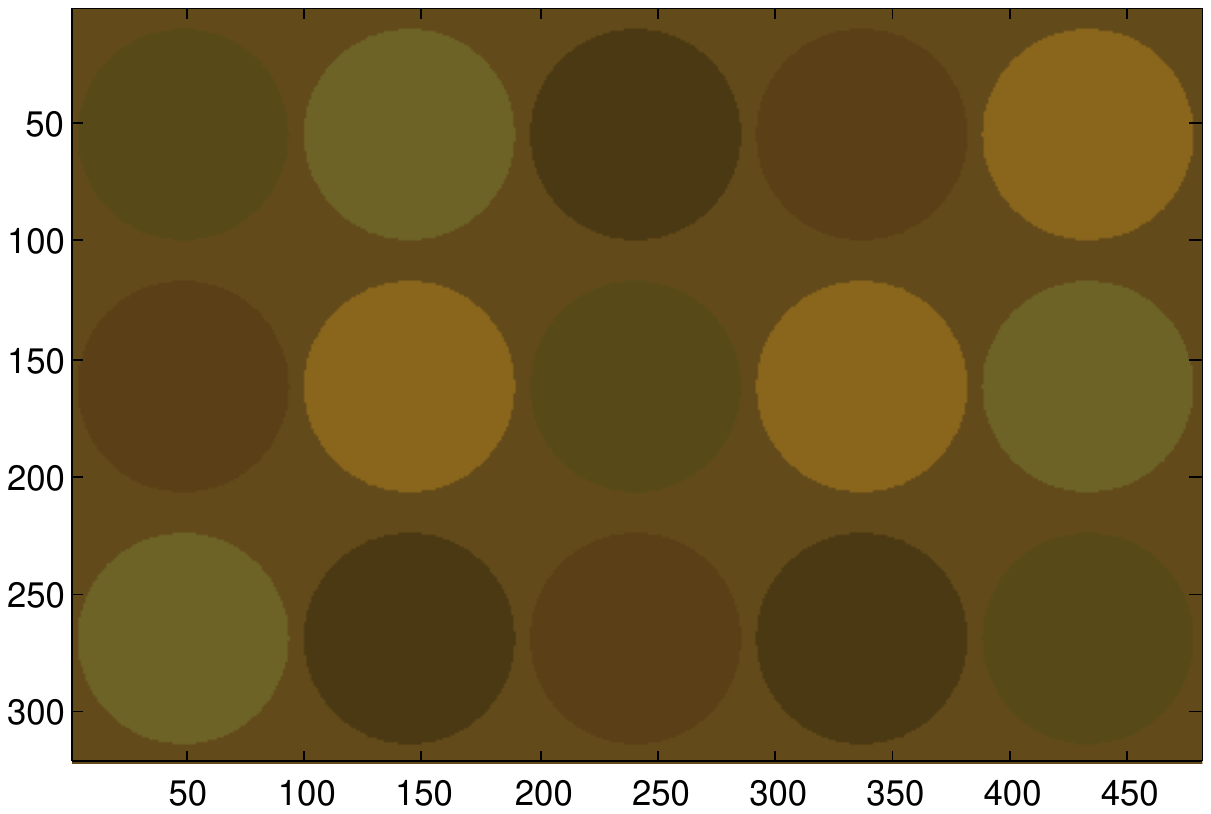}
\includegraphics[viewport = 120 300 470 540, width = 0.24\textwidth]{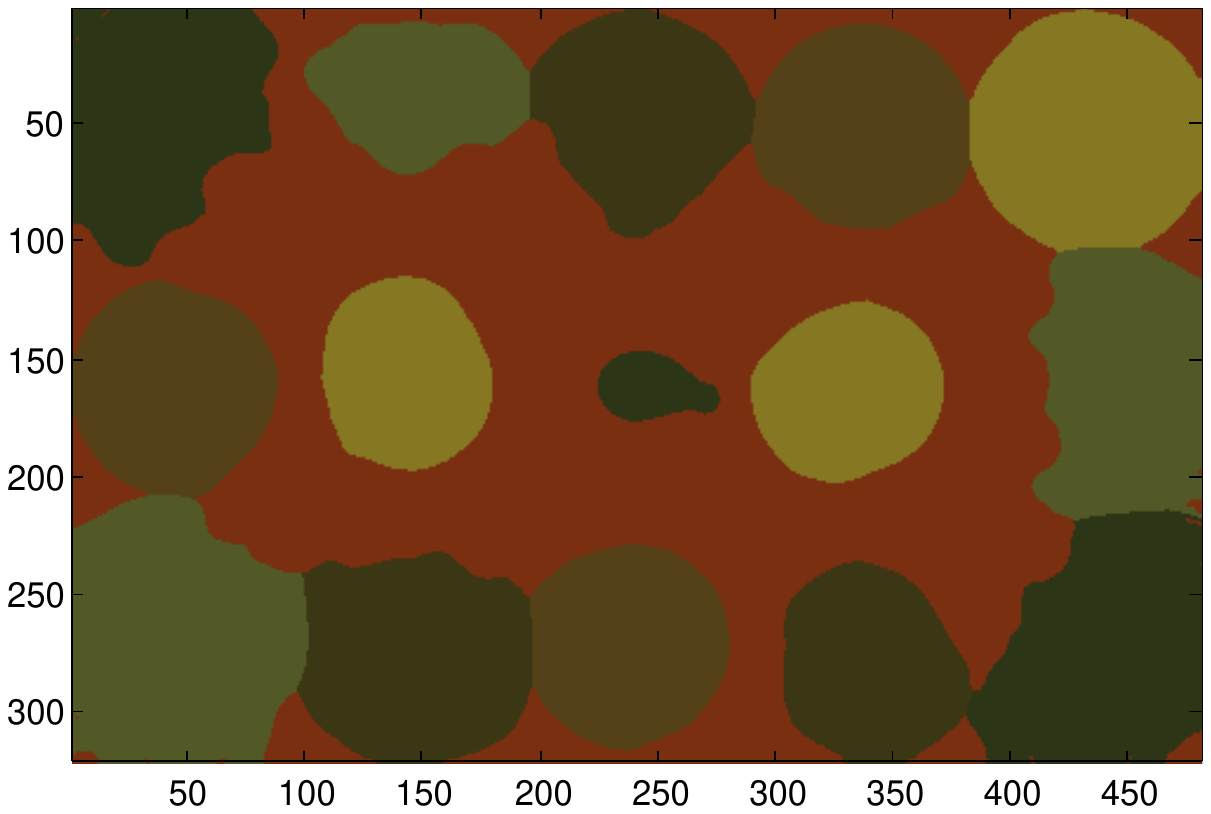}
\includegraphics[viewport = 120 300 470 540, width = 0.24\textwidth]{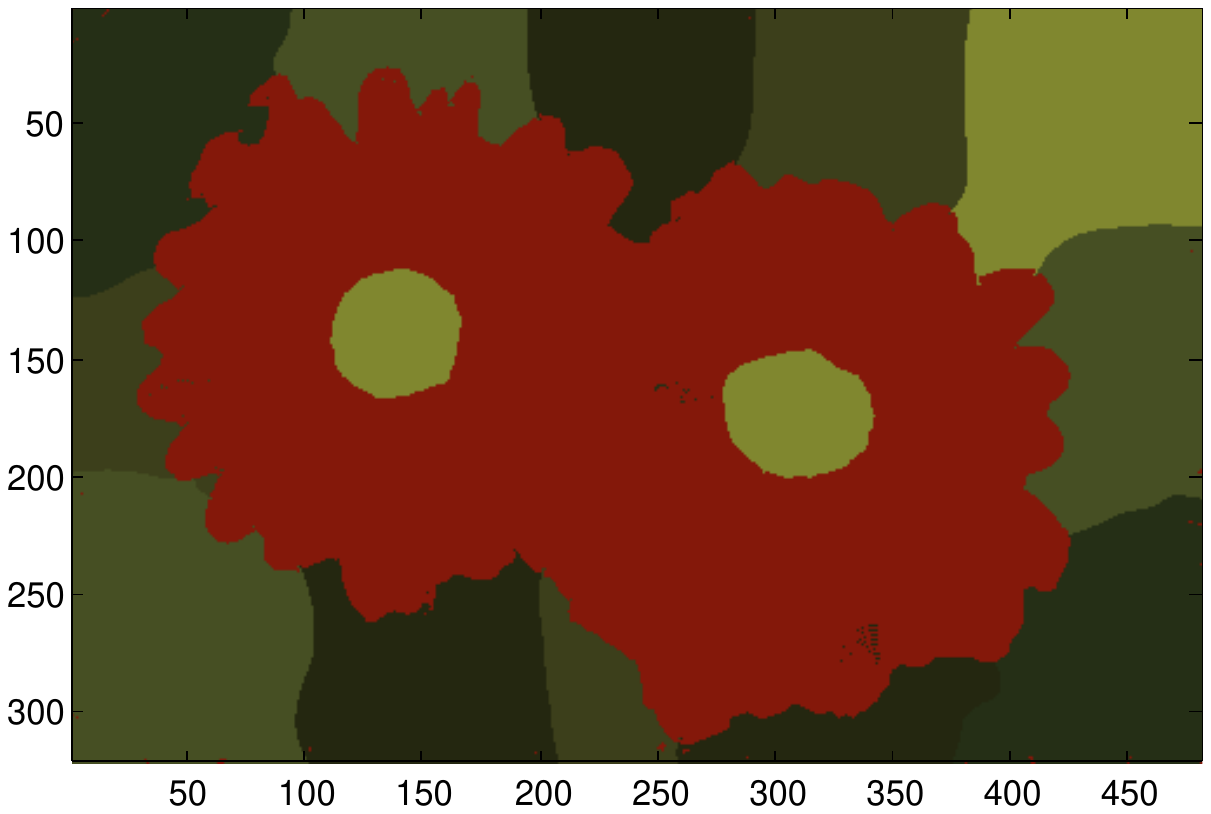}
\includegraphics[viewport = 120 300 470 540, width = 0.24\textwidth]{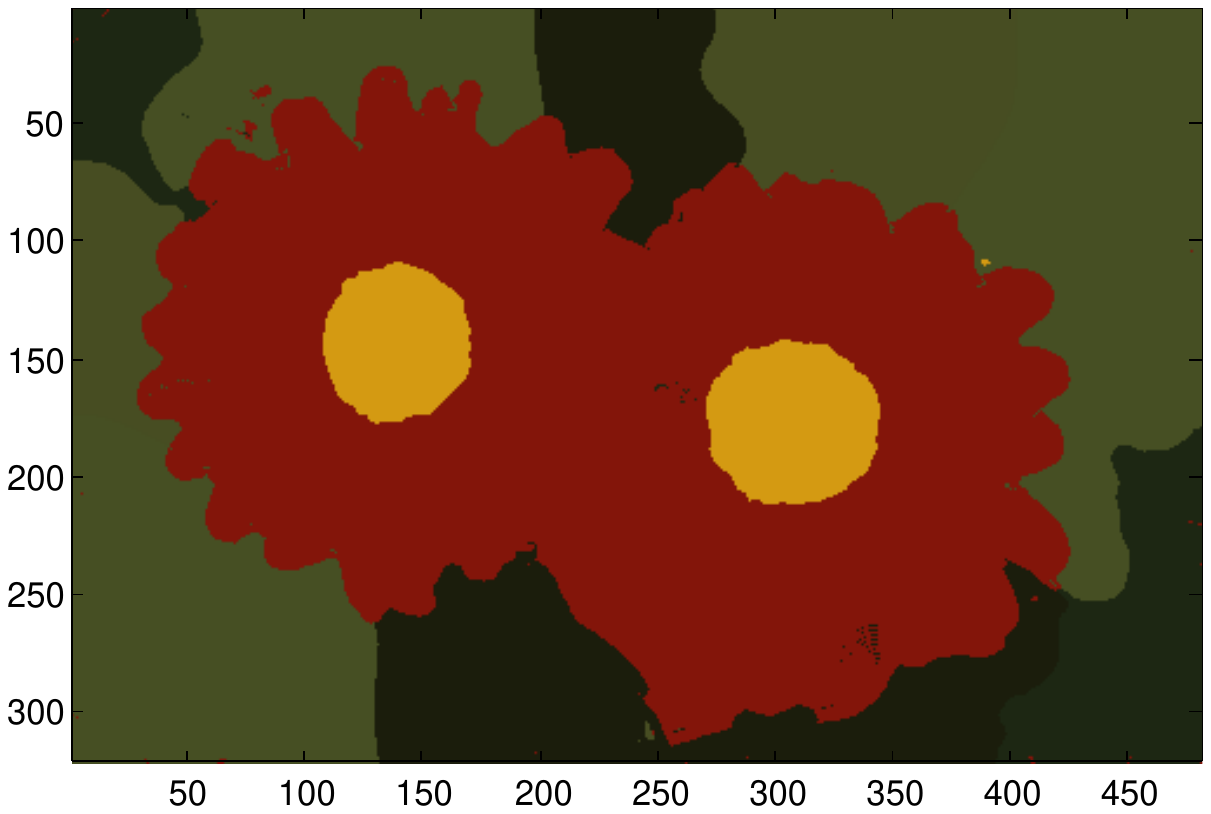}\\
\caption{Multi-phase image segmentation of a color image using HSV space, image from Berkeley database \citep{Arbelaez11}, $m=1,25,100,1900$, $\Delta t=0.01$, $\lambda_H = 150$, $\lambda_S = 50$, $\lambda_V = 50$, $\sigma$-factor $15\%$}
\label{fig:Flowers_HSV}
\end{figure}

\begin{figure}
\centering
\includegraphics[viewport = 120 300 470 540, width = 0.24\textwidth]{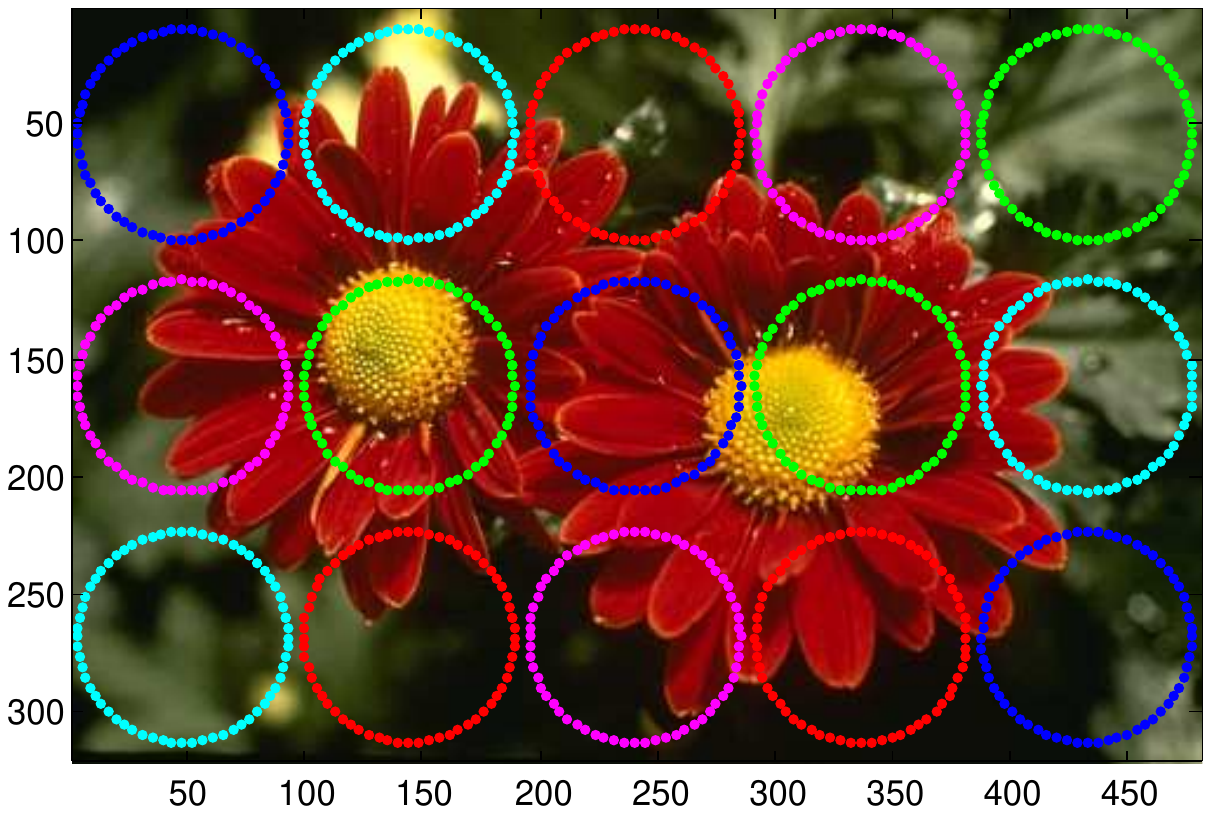}
\includegraphics[viewport = 120 300 470 540, width = 0.24\textwidth]{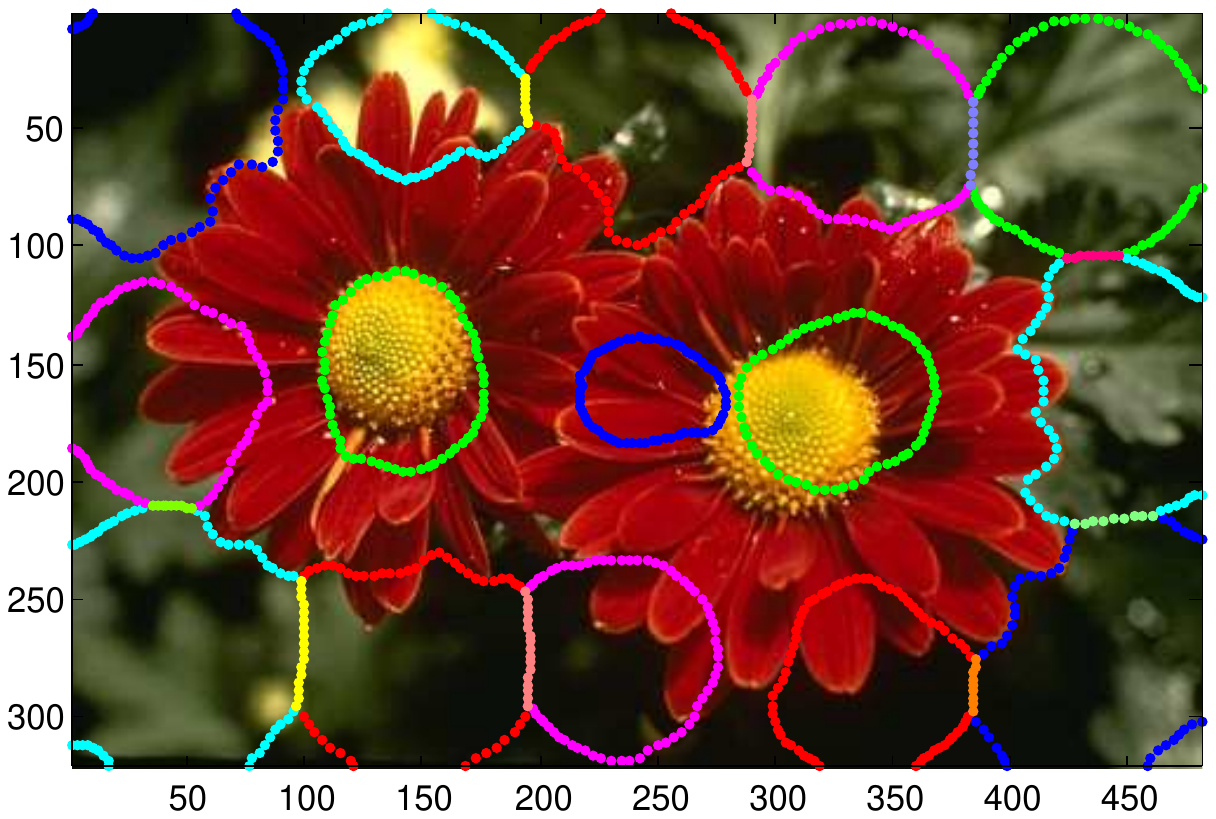}
\includegraphics[viewport = 120 300 470 540, width = 0.24\textwidth]{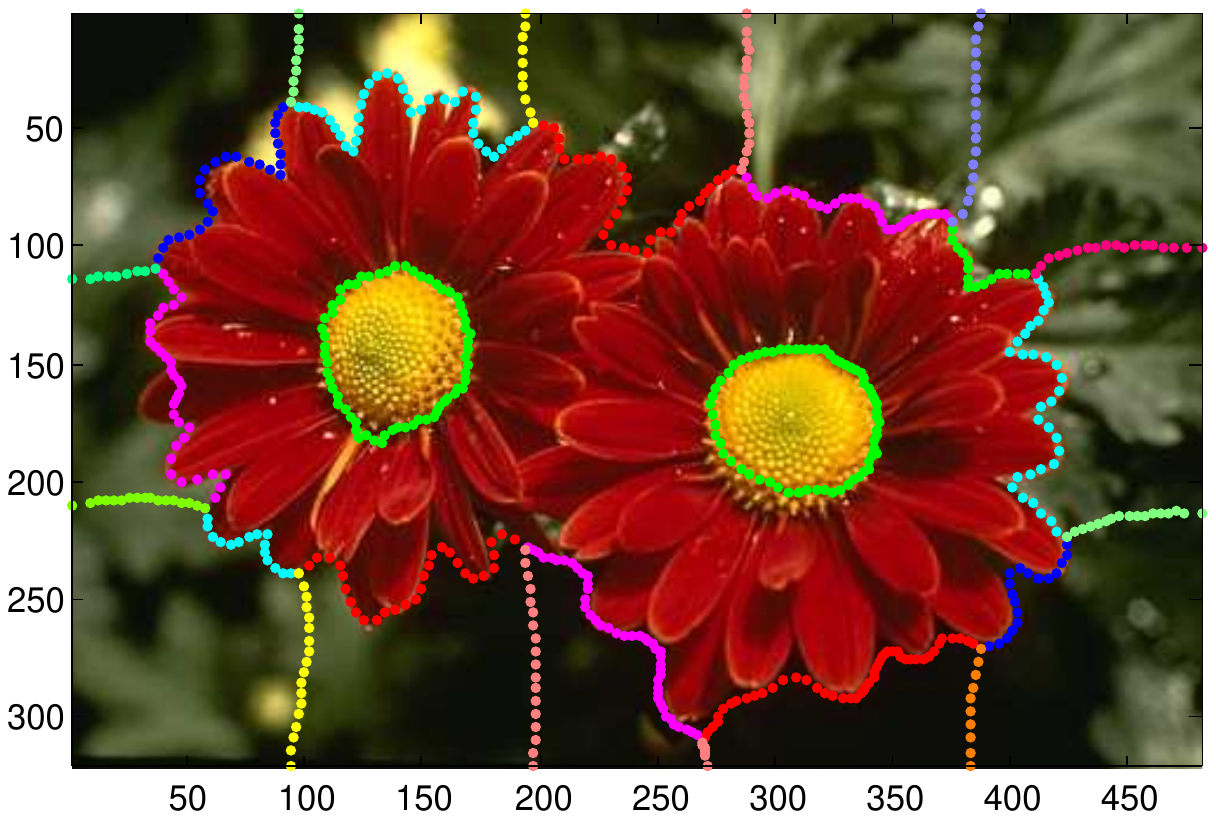}
\includegraphics[viewport = 120 300 470 540, width = 0.24\textwidth]{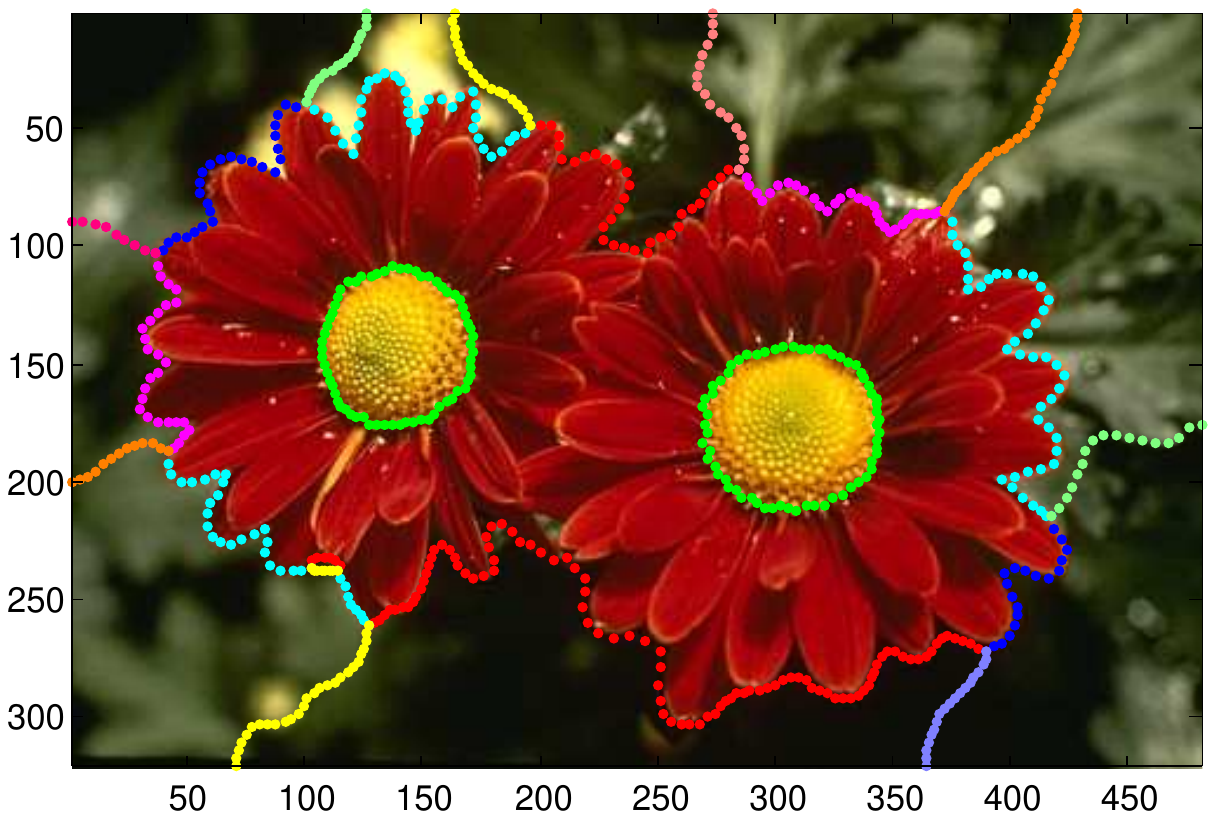}\\
\includegraphics[viewport = 120 300 470 540, width = 0.24\textwidth]{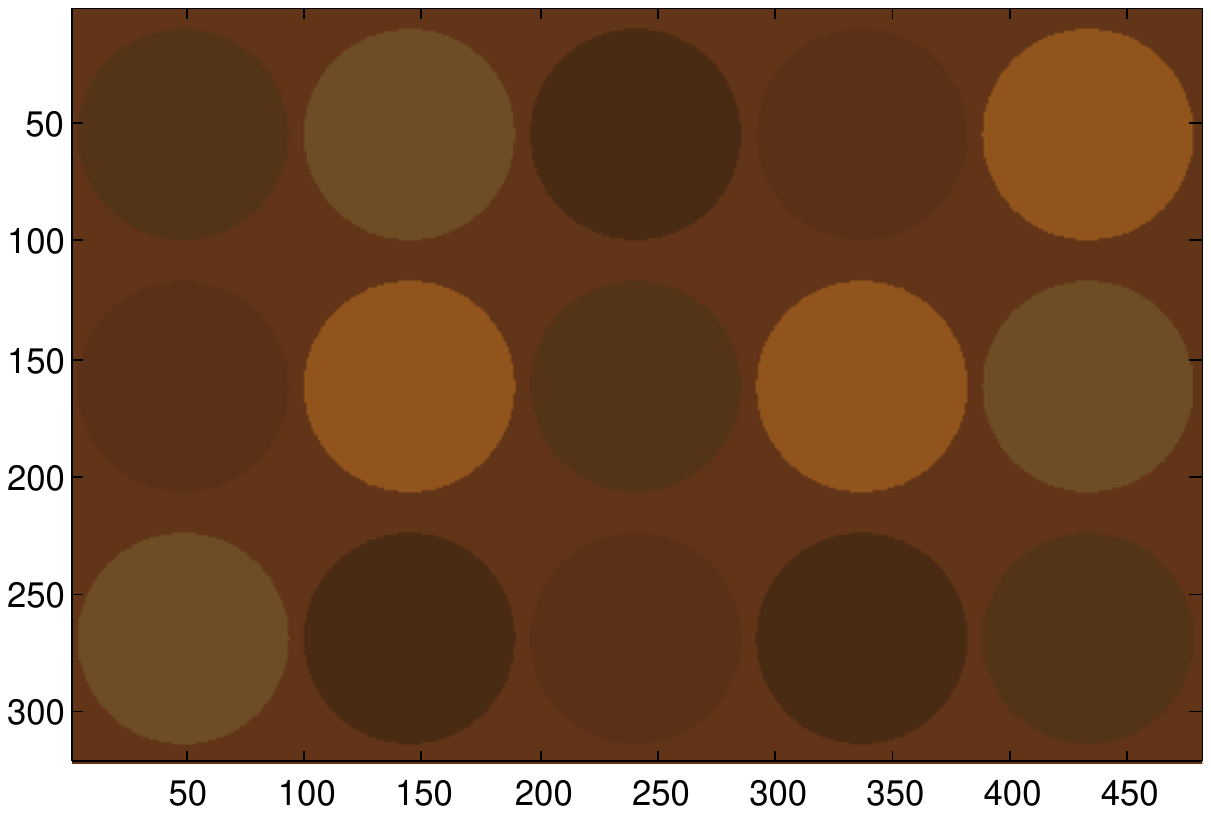}
\includegraphics[viewport = 120 300 470 540, width = 0.24\textwidth]{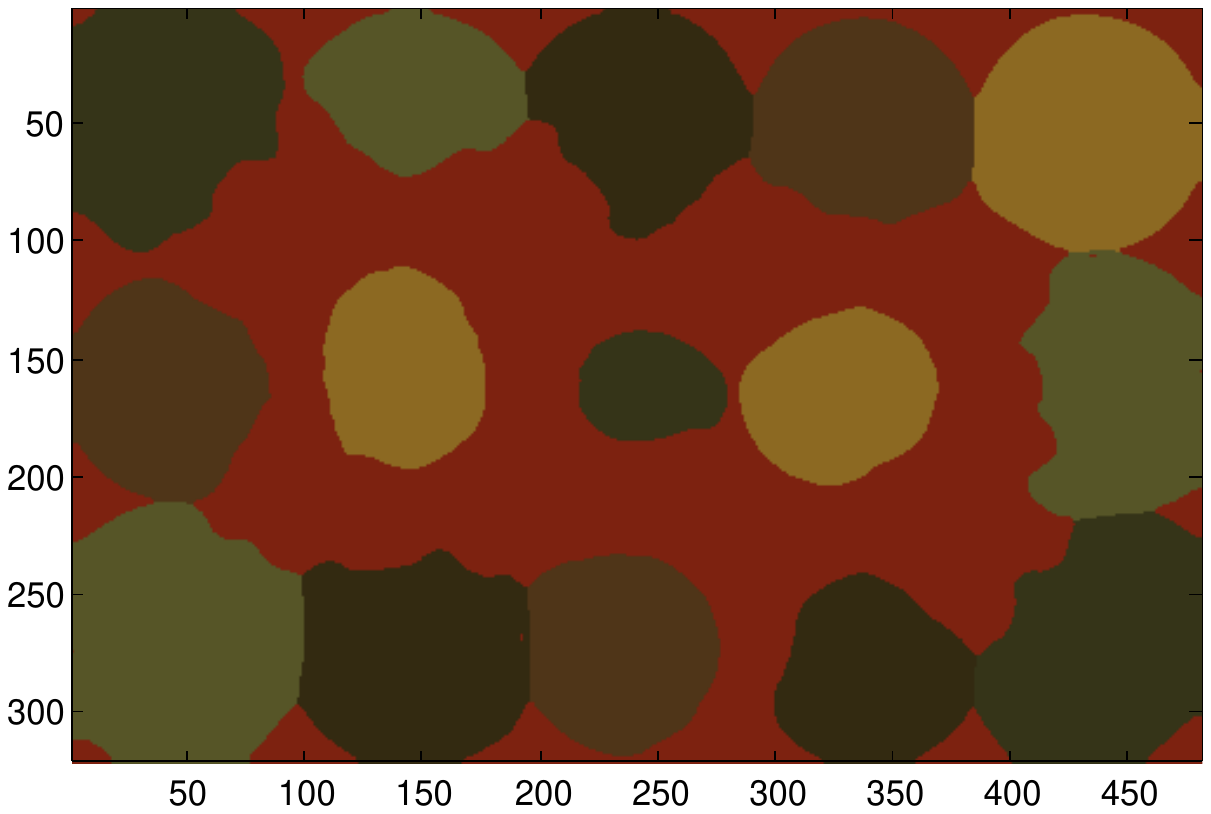}
\includegraphics[viewport = 120 300 470 540, width = 0.24\textwidth]{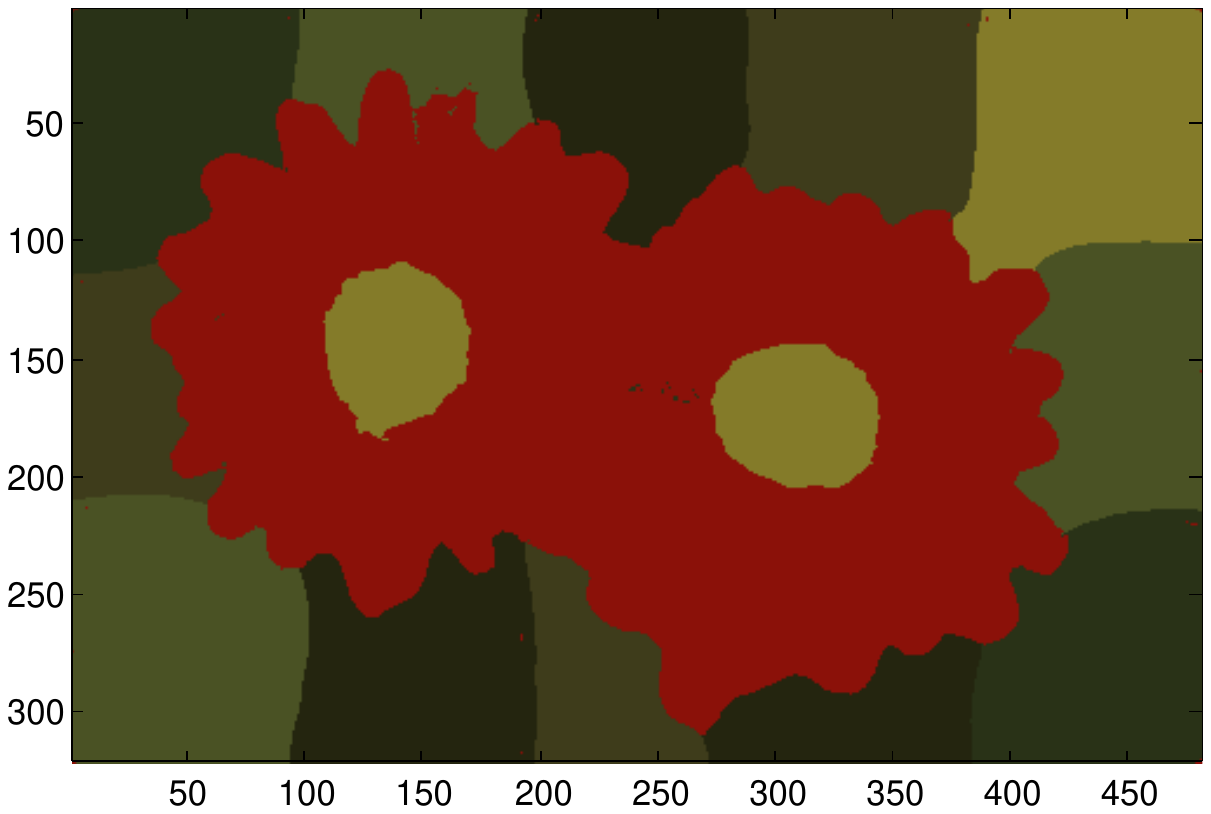}
\includegraphics[viewport = 120 300 470 540, width = 0.24\textwidth]{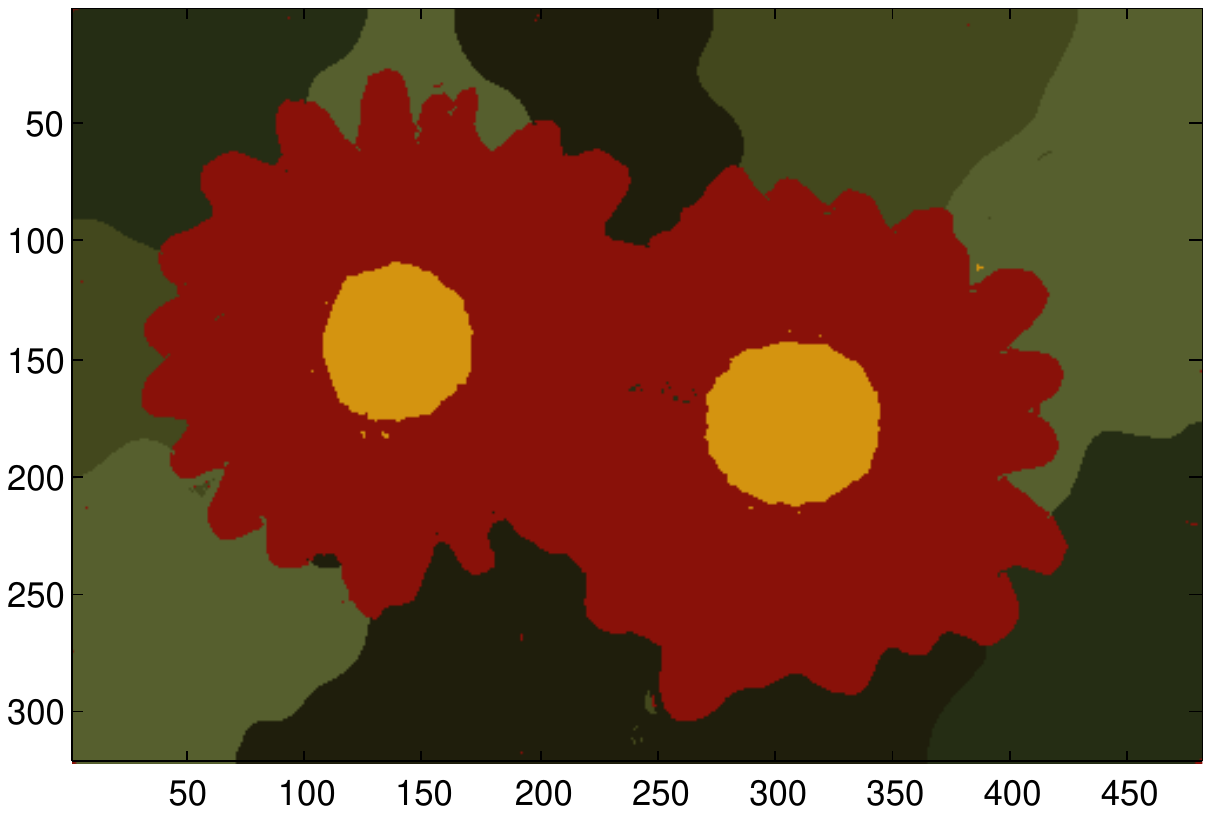}\\
\caption{Multi-phase image segmentation of a color image using CB space, image from Berkeley database \citep{Arbelaez11}, $m=1,50,100,1900$, $\Delta t=0.01$, $\lambda_C = 180$, $\lambda_B = 40$, $\sigma$-factor $15\%$}
\label{fig:Flowers_CB}
\end{figure}

\begin{figure}
\centering
\includegraphics[viewport = 105 280 475 570, width = 0.3\textwidth]{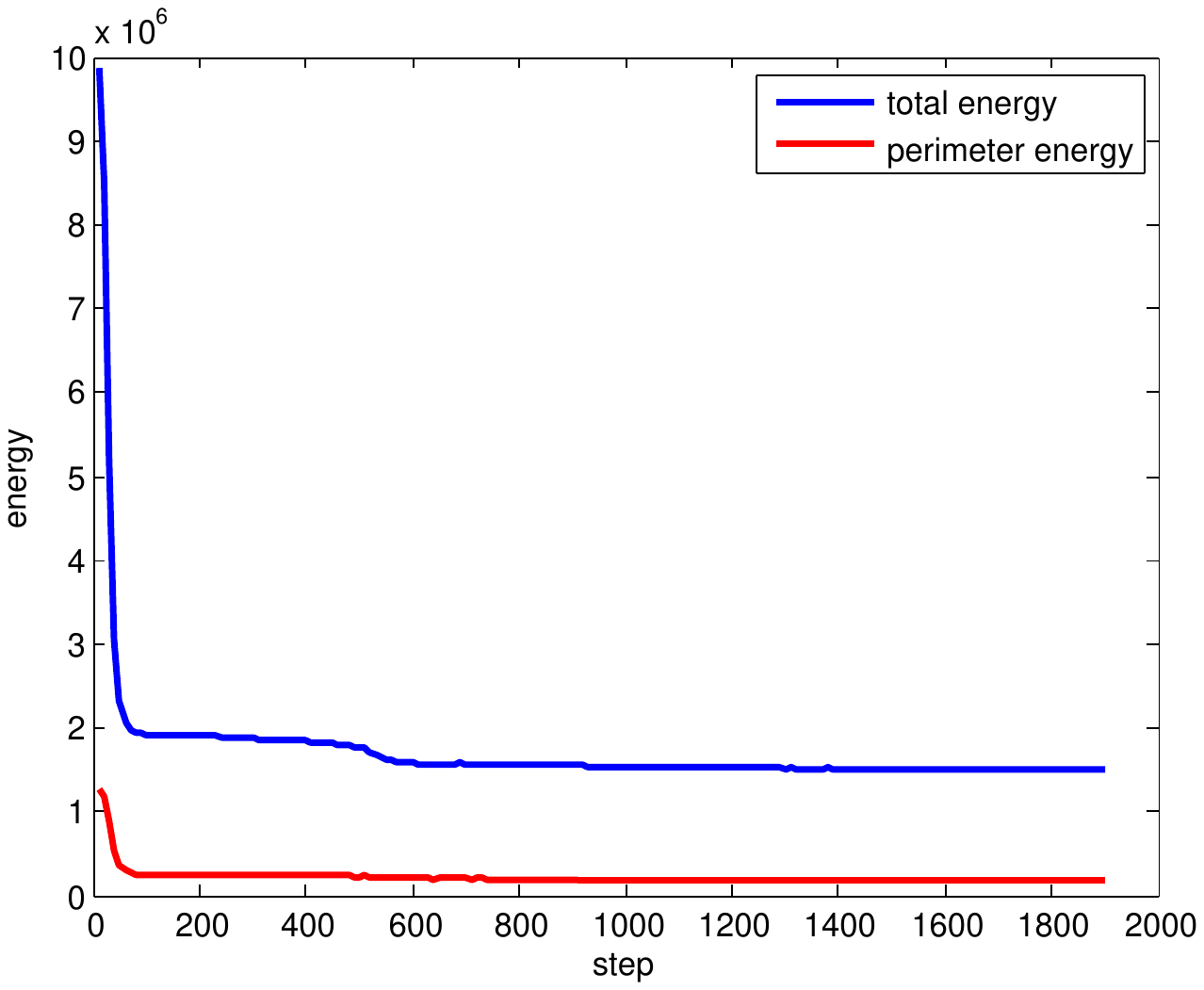}
\includegraphics[viewport = 105 280 475 570, width = 0.3\textwidth]{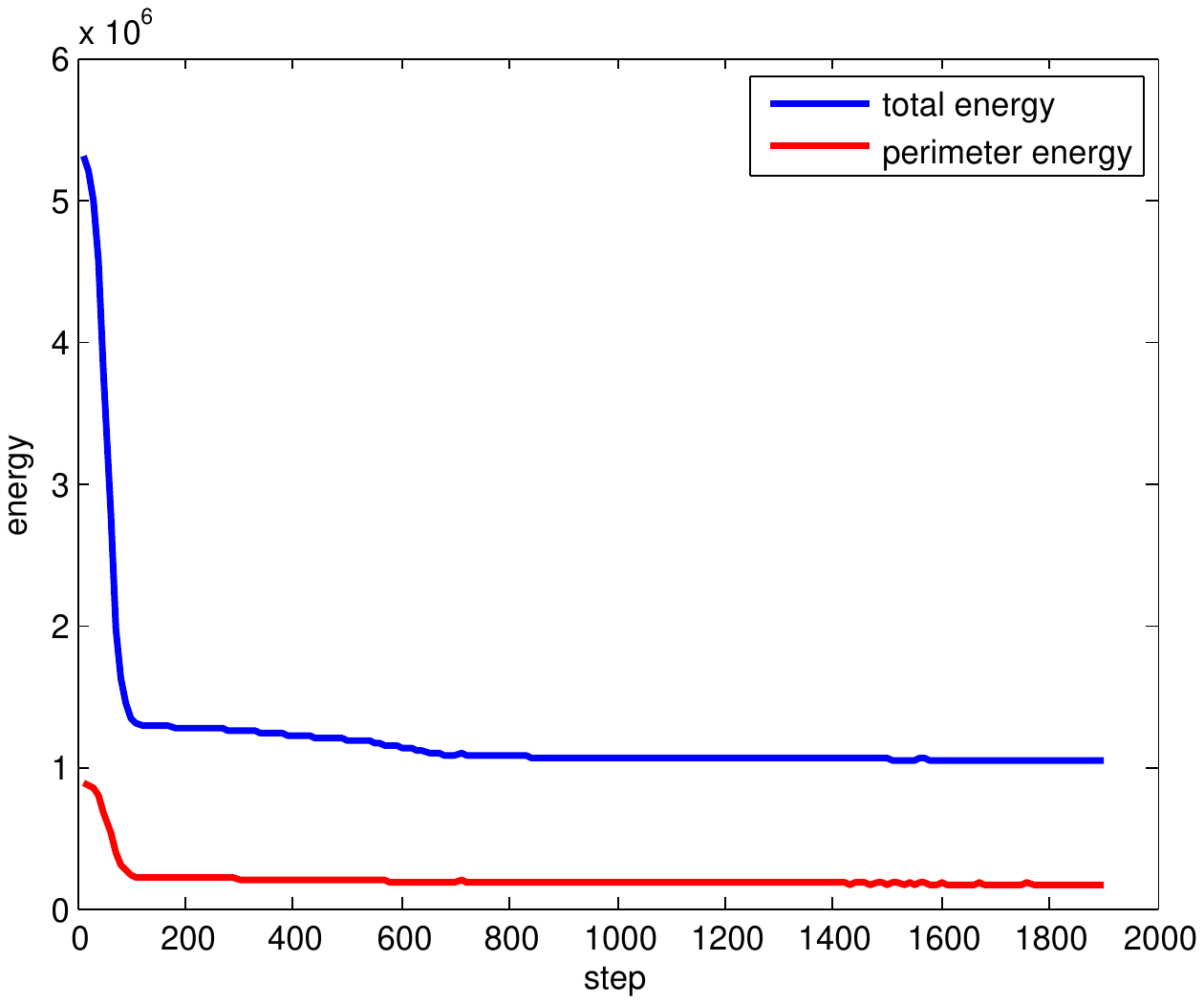}
\caption{Energy decrease segmenting the flower image from Berkeley database, Left: HSV color space, Right: CB color space }
\label{fig:energies_flowers}
\end{figure}

\begin{figure}
\centering
\includegraphics[viewport = 120 300 470 540, width = 0.33\textwidth]{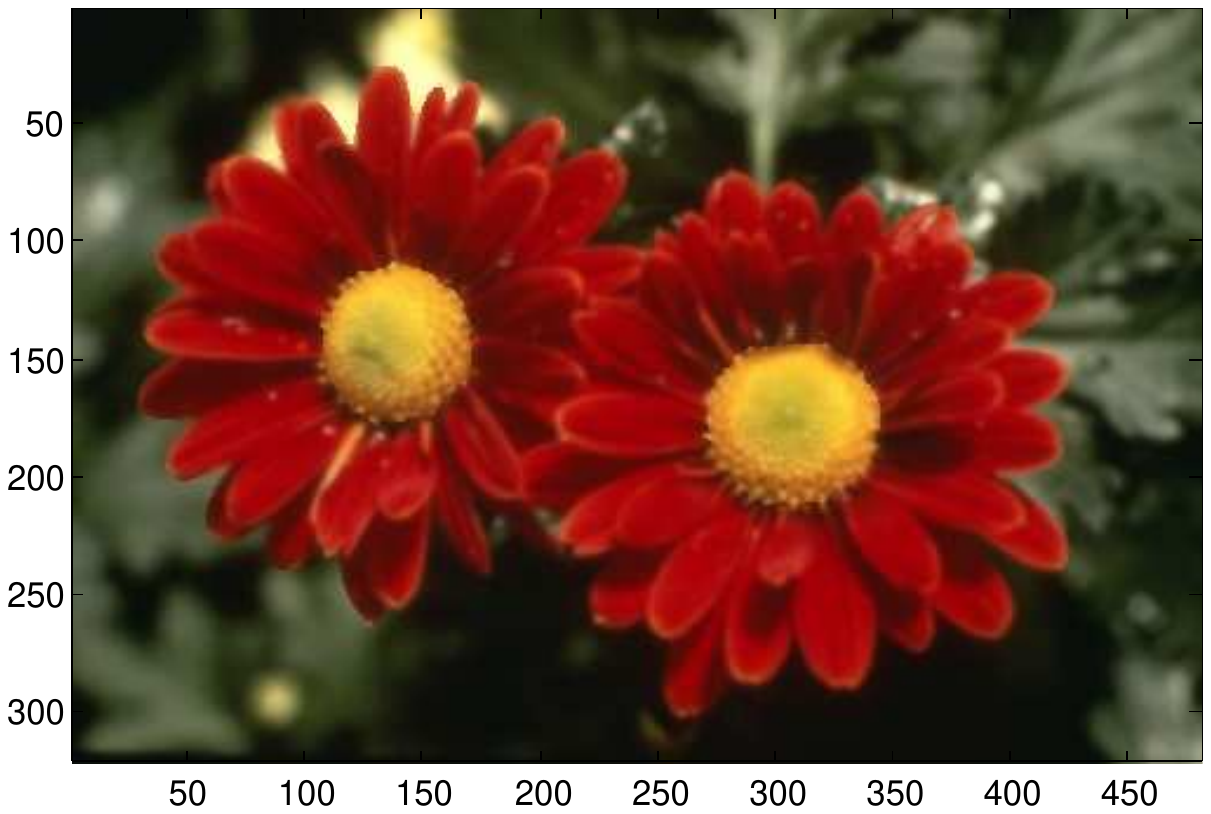}
\includegraphics[viewport = 120 300 470 540, width = 0.33\textwidth]{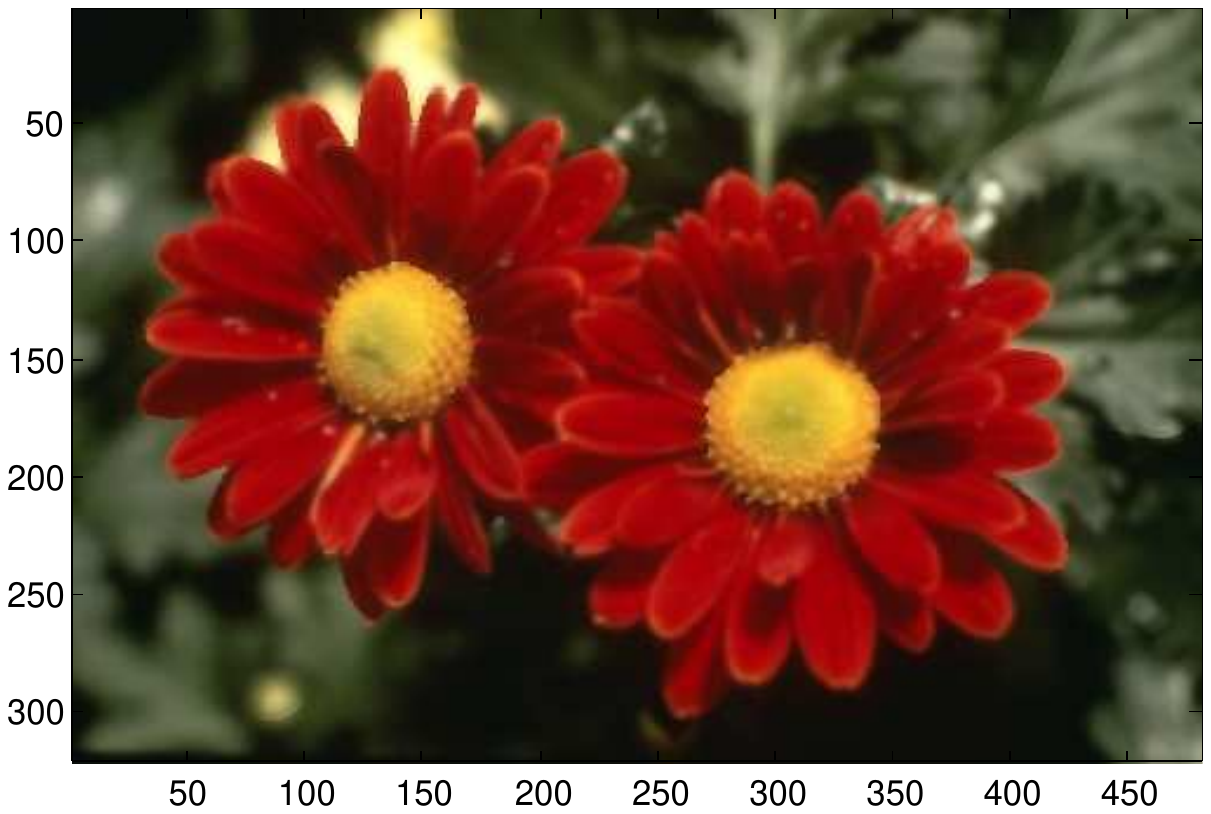}\\
\includegraphics[viewport = 120 300 470 540, width = 0.33\textwidth]{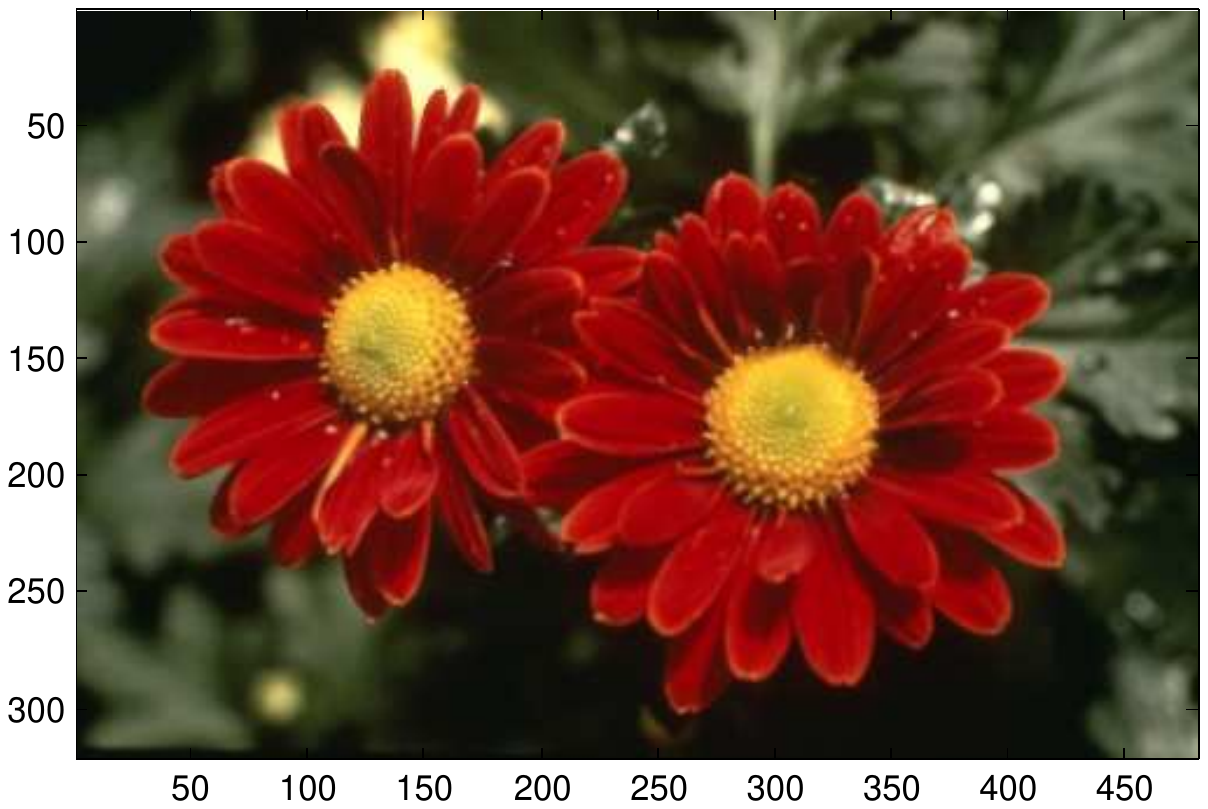}
\includegraphics[viewport = 120 300 470 540, width = 0.33\textwidth]{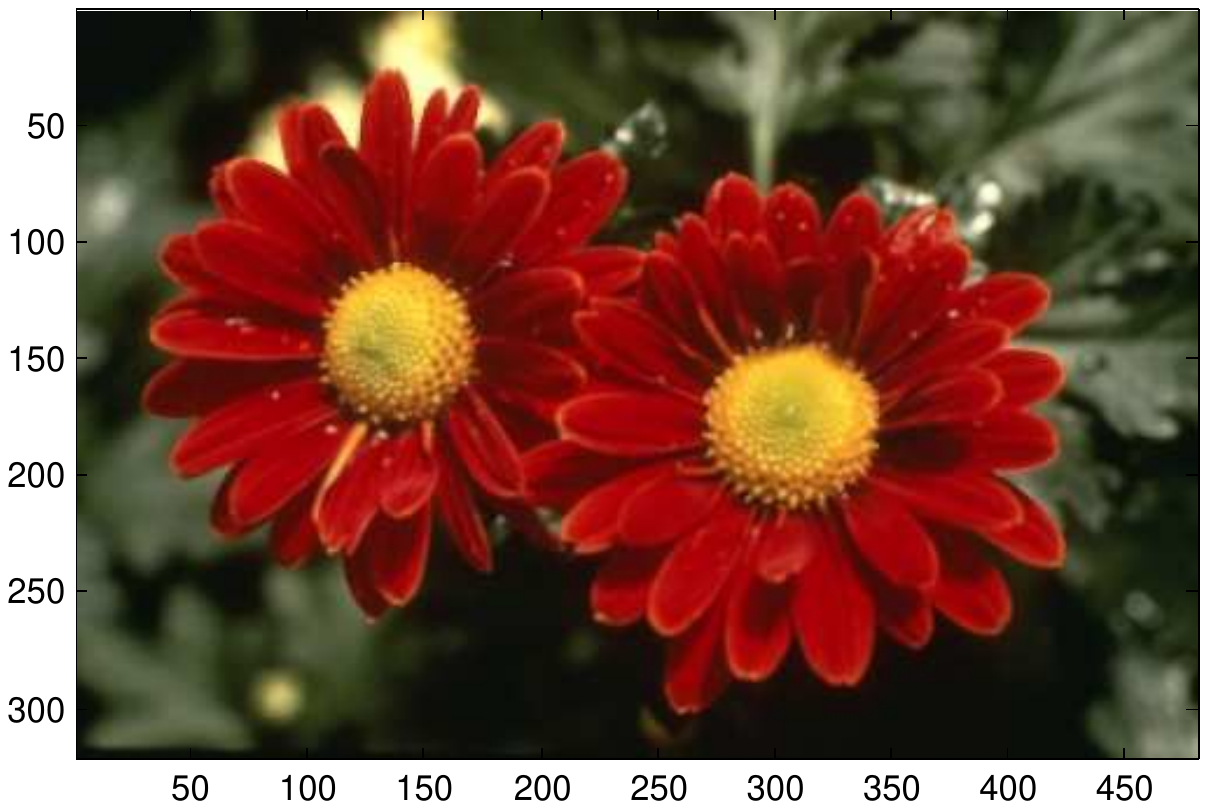}
\caption{Result of post-processing smoothing of the flower image from Berkeley database using $\lambda_k=1$ (first row) and $\lambda_k=10$ (second row), $k=1,\ldots,N_R$, Left: HSV color space, Right: CB color space }
\label{fig:final_approx_flowers}
\end{figure}

A second real colored image from the Berkeley image database \citep{Arbelaez11}) showing two flowers is used to demonstrate the algorithm and to compare the color spaces HSV and CB. Figure \ref{fig:Flowers_HSV} and \ref{fig:Flowers_CB} present the results, i.e. the contours and the piecewise constant segmentation  at time step $m=1,25,100,1900$. Apart of the weights for external forcing terms, the same parameters are used for the two runs. The segmentation using the CB color space results in a better final segmentation. The interface between red phase and dark green phase near the right flower matches better with the edges of the flower using the CB space. Furthermore, the different green phases of the background with different brightness are segmented more accurately. Figure \ref{fig:energies_flowers} shows the energy decrease. In each case, the fastest energy decrease can be noted at the beginning, i.e. in the first $200$ steps. Then, only small improvements of the segmentation are conducted where the energy decreases only slightly. The result of the post-processing image denoising step $u$, i.e. the solution of $-\frac1\lambda \Delta u + u = u_0$ for $\lambda=1$ and $\lambda=10$ with Neumann boundary conditions, is presented in Figure \ref{fig:final_approx_flowers}. 


\section{Conclusion}
We proposed a parametric approach for image segmentation using region
based active contours. The evolution equations for image segmentation
are based on the functional of \cite{Mumford89}. The evolution
equations consist of a curvature term and an external forcing term
designed for image segmentation purposes. Similar as \cite{Chan01}, we
consider piecewise smooth approximations of the original image
function. In contrast to Chan and Vese who make use of level-set
techniques, the contours are represented by smooth parameterizations
in this work. Following the parametric approach of \cite{BGN07a}, we
introduced a framework for multiple curves which can also meet at
triple junctions and can intersect with the outer boundary. Further,
an elliptic partial differential equation with Neumann boundary
conditions is derived from the Mumford-Shah functional for piecewise
smooth approximating image functions. Having detected the regions, the
image can be denoised by solving the PDE on each phase separately. Due
to the Neumann boundary conditions, the region boundaries (edges) are
not smoothed out.

A numerical approximation based on finite differences has been
presented where the smooth curves are approximated by polygonal
lines. The numerical scheme automatically provides good mesh
quality. Indeed, a semi-discrete variant of the scheme leads to meshes
with an equidistribution property. As parametric methods cannot detect
topological changes automatically, we used an efficient scheme to
detect topological changes such as splitting and merging of curves and
creation of triple junctions and boundary intersection points.  The
segmentation technique has been successfully applied on several
artificial and real images with scalar and vector-valued image data.

\section*{Acknowledgement}
The authors thank Dr. Declan O'Regan and the Robert Steiner MR Unit, MRC Clinical Sciences Centre, Imperial College London, for providing the heart slice image used in Figure \ref{fig:medical}. 

\bibliography{literatur}
\bibliographystyle{plainnat}
\end{document}